\documentclass[runningheads]{llncs}

\usepackage{eccv}

\usepackage{eccvabbrv}

\usepackage{graphicx}
\usepackage{booktabs}
 \usepackage{multirow}
\usepackage{algorithm}
\usepackage{algorithmic}
\usepackage{amsmath}
\usepackage{xcolor}
\usepackage{pifont}
\usepackage{dashrule}

\usepackage{listings}

\usepackage[accsupp]{axessibility}  

\usepackage{lstautogobble}

\usepackage[pagebackref,breaklinks,colorlinks,citecolor=eccvblue]{hyperref}

\usepackage{orcidlink}
\usepackage{wrapfig}

\begin{document}

\title{RGBT-GroundBench: Visual Grounding Beyond RGB in Complex Real-World Scenarios}

\titlerunning{RGBT-GroundBench for Visual Grounding}

\renewcommand{\thefootnote}{\fnsymbol{footnote}}

\author{
Tianyi Zhao\inst{1}$^*$\orcidlink{0000-0002-2487-3944}
\and
Jiawen Xi\inst{1}$^*$\orcidlink{0009-0006-7472-9154}
\and
Linhui Xiao\inst{2}\orcidlink{0000-0003-2592-5264}
\and
Junnan Li\inst{1}\orcidlink{0009-0005-1948-0651}
\and
Xue Yang\inst{3}\orcidlink{0000-0002-7084-9101}
\and
Maoxun Yuan\inst{1}$^\dagger$\orcidlink{0000-0001-7463-7328} \and
Xingxing Wei\inst{1}$^\dagger$\orcidlink{0000-0002-0778-8377}
}
\authorrunning{T.~Zhao et al.}


\institute{Institute of Artificial Intelligence, State Key Laboratory of Virtual Reality Technology and Systems, Beihang University, Beijing, China  \and Pengcheng Laboratory, Shenzhen, China \and Shanghai Jiao Tong University, Shanghai, China \\
\email{\{ty\_zhao, xijiawen, li\_jun\_nan, mxyuan, xxwei\}@buaa.edu.cn;} \\ \email{xiaolinhui16@mails.ucas.ac.cn;}~~\email{yangxue-2019-sjtu@sjtu.edu.cn}\\ 
Codes: \url{https://github.com/crazyxiaoxi/RGBT-GroundBench} \\
Dataset: \url{https://huggingface.co/datasets/JiawenXi/RGBT-Ground-Dataset} \\
$^*$ Equal contribution. \quad $^\dagger$ Corresponding author.\\ 
}

\maketitle
\makeatletter
\renewcommand{\theHsection}{appendix.\thesection}
\renewcommand{\theHsubsection}{appendix.\thesubsection}
\makeatother

\begin{abstract}

Visual grounding (VG) localizes target objects in an image from natural-language expressions. In real-world perception, RGB cues often degrade under low illumination and adverse weather, making visual grounding substantially more challenging. However, existing VG benchmarks are largely RGB-only and provide limited, structured coverage of such conditions, hindering systematic robustness evaluation and cross-spectral comparison. We present \textbf{RGBT-GroundBench}, the first large-scale benchmark for RGB-Thermal (TIR) visual grounding in complex environments. It contains over 40K images (21,535 RGB-TIR pairs) and 38,760 object instances with referring expressions, bounding boxes, and fine-grained annotations at three levels: scene types, environmental conditions (illumination and weather), and object properties (size and occlusion). 
As a benchmark suite, RGBT-GroundBench provides not only curated RGB-TIR grounding annotations but also a unified evaluation protocol supporting RGB-only, TIR-only, and RGB+TIR inputs. Under this protocol, we benchmark 11 representative VG models across diverse scenes and environmental conditions. Our results show that grounding accuracy is strongly correlated with scene complexity, LoRA-based models are more robust in complex scenes, and low-illumination conditions cause significant performance degradation that has been rarely explored. Guided by these observations, we introduce \textbf{RGBT-VGNet}, a simple and reproducible reference baseline under the unified protocol, featuring Asymmetric Modality Adaptation, Language-Aware Visual Synergy, and Tri-Prior Fusion for reliability-aware RGB-TIR integration. Resources, annotations, code, checkpoints, and evaluation scripts have been publicly released.

\end{abstract}


\section{Introduction}

Visual grounding (VG) aims to localize objects in an image from natural language expressions and is a fundamental capability for vision–language systems. It plays a key role in many applications, including human–robot interaction, embodied agents, autonomous driving, and assistive technologies. Over the past decade, visual grounding has advanced significantly with large-scale benchmarks such as RefCOCO/+/g~\cite{yu2016modeling,nagaraja2016modeling}, ReferIt~\cite{kazemzadeh2014referitgame}, and Flickr30K~\cite{plummer2015flickr30k}, and vision-language pretraining paradigms (\textit{e.g.}, CLIP~\cite{radford2021learning}, BEIT3~\cite{wang2023image}, LLaVA~\cite{liu2023visual}, LLaVA-SP~\cite{lou2025llava}, etc.), enabling rapid progress in vision–language models and multimodal reasoning.

\begin{table*}[t]
    \centering
    \caption{Comparison of RGBT-GroundBench with existing benchmarks. This table highlights key characteristics such as modality and the proportion of weak-light scenes and small objects.}
     \vspace{-0.3cm}
\resizebox{\textwidth}{!}{
    \begin{tabular}{c|c|c|c|c|c|c|c}
        \toprule
        \textbf{Benchmark} & \textbf{Modality} & \textbf{Typical res.} & \textbf{\#Instance} & \textbf{\#Weak-light} & \textbf{\#Small Object} & \textbf{Query Length} & \textbf{Data Source} \\
        \midrule
        Flickr30K~\cite{plummer2015flickr30k} & RGB & 500$\times$375 & 276,000 & 2,588~(0.9\%) & 0~(0.0\%) & 1.59 & Flickr~\cite{kazemzadeh2014referitgame}, IAPR-TC~\cite{grubinger2006iapr} \\
        ReferIt~\cite{kazemzadeh2014referitgame} & RGB & 480$\times$360 & 96,654 & 2,588~(2.77\%) & 0~(0.0\%) & 3.45 & IAPR-TC~\cite{grubinger2006iapr} \\
        RefCOCO~\cite{yu2016modeling} & RGB & 640$\times$480 & 50,000 & 3,068~(6.1\%) & 0~(0.0\%) & 3.49 & MSCOCO~\cite{lin2014microsoft} \\
        RefCOCO+~\cite{yu2016modeling} & RGB & 640$\times$480 & 49,856 & 1,415~(2.8\%) & 5,998~(12.0\%) & 3.58 & MSCOCO~\cite{lin2014microsoft} \\
        RefCOCOg~\cite{nagaraja2016modeling} & RGB & 640$\times$480 & 49,822 & 5,342~(10.7\%) & 22,609~(45.4\%) & 8.47 & MSCOCO~\cite{lin2014microsoft} \\
        RGBT-GroundBench & RGB+TIR & 640$\times$512 & 38,760 & 16,726~(\textbf{43.2\%}) & 22,009~(\textbf{56.8\%}) & 14.24 & FLIR~\cite{zhang2020multispectral}, M$^{3}$FD~\cite{liu2022target}, MFAD~\cite{hu2025ei} \\
        \bottomrule
    \end{tabular}
}
     \vspace{-0.4cm}
    \label{tab:benchmarks_comparison}
\end{table*}

\begin{figure}[t]
    \centering
    \includegraphics[width=\linewidth]{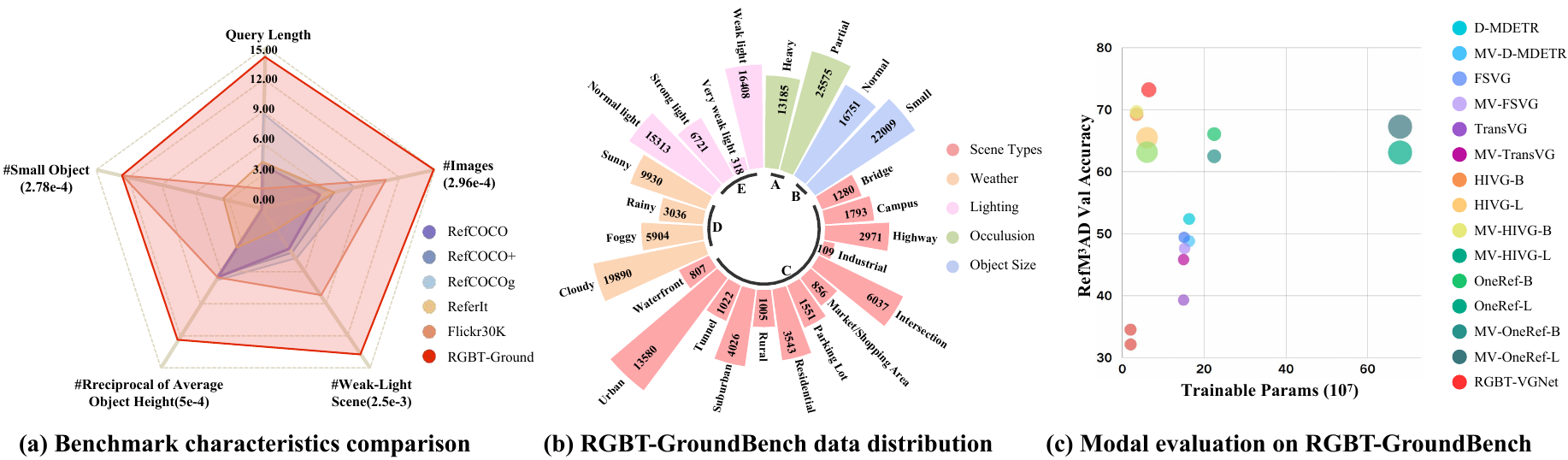}
    \caption{Comparative overview of RGBT-GroundBench and evaluations of representative grounding methods. (a) Comparison with classical visual grounding benchmarks. (b) Distribution of lighting, occlusion, object, weather, and scene types. (c) Evaluation of representative methods and our RGBT-VGNet baseline on RGBT-GroundBench.}
    \label{fig:1}
     \vspace{-0.6cm}
\end{figure}

Despite remarkable progress in visual grounding, existing VG benchmarks still suffer from several critical limitations that hinder their applicability in real-world scenarios:
1) Limited Scene Complexity.
Most benchmarks are built from datasets collected in clean environments, typically featuring centered and salient objects. Such settings fail to reflect the cluttered and dynamic nature of real-world scenes.
2) Insufficient Object Diversity.
Small, distant, and occluded objects are underrepresented, making it difficult to evaluate model robustness in safety-critical domains such as autonomous driving and embodied perception.
3) RGB-Centric Modality Design.
Current benchmarks rely almost exclusively on RGB imagery, which limits the evaluation of grounding models under conditions where visual cues are degraded.

Meanwhile, real-world perception often occurs under challenging conditions such as low illumination, nighttime scenes, adverse weather, or severe occlusion, where RGB sensors alone may provide unreliable visual cues and degrade grounding accuracy. Thermal infrared (TIR) sensors can offer a complementary modality by capturing heat signatures and remaining robust under poor lighting conditions, suggesting potential benefits for grounding in such environments. Although RGB-Thermal fusion has been extensively studied in object detection~\cite{yuan2024improving, zhao2025removal,zhao2025rethinking} and semantic segmentation~\cite{yuan2025unirgb, zhao2023mitigating}, its integration with language-guided perception remains largely unexplored. This motivates the study of visual grounding under RGB-thermal sensing.

To support research in this direction, we present \textbf{RGBT-GroundBench}, the first large-scale benchmark for RGB-Thermal visual grounding in complex real-world environments. As shown in Figure~\ref{fig:1} and Table~\ref{tab:benchmarks_comparison}, RGBT-GroundBench includes a curated data resource of over 40K images~(21,535 RGB-TIR pairs) and 38,760 object instances with fine-grained annotations at three levels: scene (13 scene types), environment (4 illumination and 4 weather conditions), and object (size and occlusion). Compared with previous benchmarks, it includes a substantially higher proportion of small objects (56.8\%) and low-light instances (43.2\%), as well as richer language descriptions with an average length of 14.24 words per query. 
These annotations enable systematic evaluation of grounding robustness across diverse real-world conditions. We hope that RGBT-GroundBench will serve as a comprehensive benchmark for RGB-Thermal visual grounding, providing not only curated RGB-TIR grounding resources but also standardized splits, fine-grained condition labels, and unified RGB/TIR/RGB+TIR evaluation settings for future research.

As part of RGBT-GroundBench, we construct a unified evaluation framework for the RGB-Thermal visual grounding task and conduct large-scale experiments to compare existing VG models under the proposed RGBT-GroundBench protocol. Specifically, we evaluate 11 visual grounding models under the 13 scenes, 4 weather conditions and 4 illumination levels. These models are evaluated with different input modalities including RGB-only, TIR-only and RGB+TIR. From the evaluation results, we find that: \textit{\textbf{1)}} the localization accuracy of VG models is highly correlated with the scene complexity. \textit{\textbf{2)}} LoRA-based VG models such as HiVG are more resistant in complex scenes. \textit{\textbf{3)}} low-illumination scenes have a significant impact on the VG model performance, rarely explored before. \textbf{\textit{More discussions are provided in Section~\ref{sec:discussion}}}.



Guided by the above benchmark observations, we introduce RGBT-VGNet as a simple and competitive baseline under our unified RGBT grounding protocol. It includes three practical components:
\textbf{AMA} (Asymmetric Modality Adaptation) follows our observation that LoRA-based grounding models are more robust in complex scenes. It can improve domain adaptation, extending RGB-pretrained VLM to cross-spectral inputs. Thus, we adopt an asymmetric design assigning higher adaptation capacity to the TIR branch to better align RGB-TIR representations.
\textbf{TPF} (Tri-Prior Fusion) addresses the strong performance drop under low illumination. It performs illumination-aware fusion by adjusting the contributions of RGB and TIR cues with simple reliability priors.
\textbf{LAVS} (Language-Aware Visual Synergy) is introduced to provide more informative cross-modal features for the TPF module. It uses a text-guided cross-modal interaction mechanism, where textual features act as queries to select and align informative regions from each modality before fusion.
Overall, our main contributions are as follows:
\begin{itemize}
\item \textbf{Comprehensive Benchmark.}
We introduce \textbf{RGBT-GroundBench}, the first large-scale benchmark for RGB-Thermal visual grounding in complex real-world scenarios, with fine-grained annotations and a unified protocol for fair RGB/TIR/RGB+TIR evaluation.
\item \textbf{Insightful Findings.}
Through systematic benchmark-wide evaluation,  we derive insightful findings into real-world cross-spectral grounding, highlighting:
(i) scene-complexity sensitivity,
(ii) LoRA robustness gains,
(iii) low-light failure modes,
and (iv) MLLM cross-spectral limitations.
\item \textbf{Competitive Baseline.}
We present RGBT-VGNet as a competitive baseline for benchmark validation, with practical components for modality adaptation and illumination-aware cross-modal interaction.
\item \textbf{Release and Availability.} We have released RGBT-GroundBench resources, annotations, splits, code, checkpoints, and evaluation scripts. There are no hidden test sets or private servers and source datasets follow original licenses.

\end{itemize}

\section{Related Works}
\label{sec:relatedworks}


\subsection{Visual Grounding in the RGB Domain}


With the rise of deep learning~\cite{he2016deep}, visual grounding has rapidly advanced~\cite{chen2020uniter, gan2020large, yang2022unitab, wang2022ofa, li2022mplug}, driven by increasingly powerful visual and language representation models~\cite{devlin2019bert, dosovitskiy2020image, vaswani2017attention}. Early approaches such as ReSC~\cite{yang2020improving} adopted CNN-based region-ranking pipelines. Later, transformer-based architectures, including TransVG~\cite{deng2021transvg}, QRNet~\cite{ye2022shifting}, D-MDETR~\cite{shi2023dynamic}, and FSVG~\cite{wang2025simple}, enabled end-to-end grounding with stronger global and contextual reasoning. The development of vision-language pre-training further boosted performance, as demonstrated by methods like CLIP-VG~\cite{xiao2023clip} and HiVG~\cite{xiao2024hivg}, which leverage large-scale image-text alignment to enhance cross-modal representations. Other lines of works explore attention-centric grounding. AttBalance~\cite{kang2025visual} imposes and balances attention-driven training constraints to refine language-relevant regions, while F-LMM~\cite{wu2025f} and Localization Head~\cite{kang2025your} methods instead exploit intrinsic text-to-image attention in frozen MLLMs as grounding cues, either by translating attention to mask logits or by selecting a few localization heads for training-free localization.
However, existing VG methods are mainly designed and evaluated on RGB-only visual input datasets, making them sensitive to illumination changes and adverse weather, and limiting robustness in challenging real-world conditions.

\subsection{Visual Grounding Beyond the RGB Domain}

Existing 2D visual grounding methods rely almost exclusively on RGB imagery, making them sensitive to illumination changes and adverse weather conditions. While some works explore extending VG beyond RGB, for example, by incorporating depth information in RGB-D settings~\cite{liu2021refer,chen2023unit3d,miyanishi2024cross3dvg}, their formulations target 3D environments and are not directly applicable to 2D grounding. Moreover, depth sensing often degrades under low-light or adverse weather conditions, limiting its effectiveness as a complementary modality for robust 2D grounding.
In contrast, thermal imaging is inherently illumination-invariant~\cite{yuan2026seeing} and provides stable cues in nighttime, low-light, and adverse weather scenarios, motivating the exploration of RGB-Thermal (RGBT) visual grounding.
Although RGBT perception has made notable progress in detection~\cite{zhao2025removal,zhao2025rethinking, yuan2024c2former,do2024d3t,song2025refinefuse} and segmentation~\cite{yuan2025unirgb,li2022rgb,zhang2021abmdrnet}, RGBT visual grounding remains largely unexplored. This gap motivates the construction of our RGBT-GroundBench, which enables systematic evaluation of multi-sensor grounding models under diverse and challenging real-world conditions.

\begin{table*}[!t]
    \centering
    \begin{minipage}[t]{0.52\textwidth}
        \centering
        \captionof{table}{Dataset composition of RGBT-GroundBench, detailing the number of instances from each source and the split across Train, Val, Test, and specialized test subsets.}
        \label{tab:RGBT-GroundBench_benchmark_data_split}
        \vspace{-0.3cm}
        \resizebox{\textwidth}{!}{
            \begin{tabular}{cc|c|ccc|ccc|c}
                \toprule
                \textbf{Benchmark}&\textbf{Sub-dataset} & \textbf{\#Ins} & \textbf{Train} & \textbf{Val} & \textbf{Test} & \textbf{TestA} &\textbf{TestB} & \textbf{TestC}&\textbf{Data Source} \\ \midrule
                & RefFLIR   & 9,712  & 7,000  & 608  & 2,104&  837& 640  & 968 &FLIR~\cite{zhang2020multispectral} \\
                {\textbf{RGBT-}}&RefM$^{3}$FD   & 7,548  & 3,604  & 168  & 3,776 &1,232 & 1,094 & 1,848&M$^{3}$FD~\cite{liu2022target} \\ 
                {\textbf{GroundBench}}& RefMFAD   & 21,500 & 16,000 & 1,256 & 4,244  &789& 2,452 & 2,544&MFAD~\cite{hu2025ei}  \\ 
                &Total  & 38,760 & 26,604 & 2,032 & 10,124&2,858 & 4,186 & 5,360& ALL\\ \bottomrule
            \end{tabular}
        }
    \end{minipage}
    \hfill
    \begin{minipage}[t]{0.46\textwidth}
        \centering
        \captionof{table}{Lighting-weather cross statistics in RGBT-GroundBench for double-checking annotations; distributions follow illumination-weather relations.}
        \label{tab:RGBT-GroundBench_benchmark_distrbution_of_weather_light}
        \vspace{-0.3cm}
        \resizebox{\textwidth}{!}{
            \begin{tabular}{c|c|c|c|c}
                \toprule
                \textbf{Light-Weather}&\textbf{Foggy (FY)} &\textbf{Rainy (RY)} & \textbf{Sunny (SY)} & \textbf{Cloudy (CY)} \\ \midrule
                \textbf{Very Weak (VWL)}& 71(0.33\%) & 29(0.13\%) & 0(0.00\%) & 101(0.47\%) \\ 
                \textbf{Weak (WL)} & 2,754(12.79\%) & 1,298(6.03\%) & 14(0.07\%) & 4,704(21.84\%) \\ 
                \textbf{Normal (NL)} & 521(2.42\%) & 150(0.70\%) & 2,191(10.19\%) & 6,157(28.59\%) \\ 
                \textbf{Strong (SL)} & 4(0.02\%) & 0(0.00\%) & 3,152(14.64\%) & 389(1.81\%) \\ \bottomrule
            \end{tabular}
        }
    \end{minipage}
    \vspace{-0.35cm}
\end{table*}

\section{Benchmark: {RGBT-GroundBench}}
\subsection{Overview}

\begin{figure*}[!t]
    \centering
    \includegraphics[width=\linewidth]{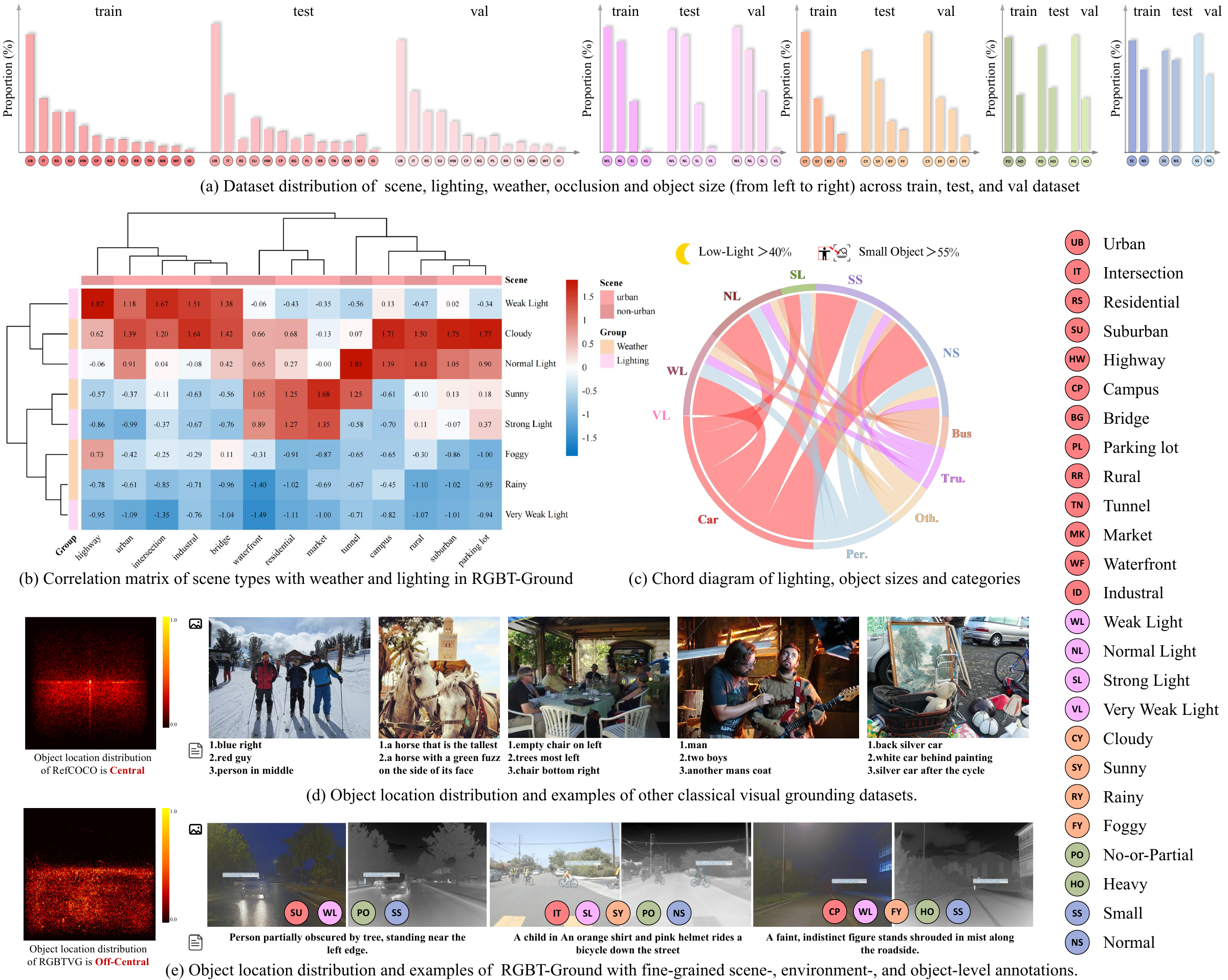}
    \vspace{-0.4cm}
    \caption{Overview of RGBT-GroundBench. (a-c) show distributions and correlations of scene-, environment-, and object-level annotations; (d-e) show object-location distributions and example samples.}
    \vspace{-0.4cm}
\label{fig:benchmark_comparison}
\end{figure*}

\textbf{RGBT-GroundBench} is, to the best of our knowledge, the first large-scale RGB-Thermal visual grounding benchmark for complex real-world scenarios. It contains over 40K images (21,535 RGBT image pairs) and 38,760 object instances from diverse environments, with spatially aligned or weakly aligned RGB-TIR pairs. Each pair is annotated with high-quality natural language referring expressions, corresponding object bounding boxes, and fine-grained annotations.  
As illustrated in Figure~\ref{fig:benchmark_comparison}, RGBT-GroundBench provides comprehensive annotations covering scene types, illumination conditions, weather variations, object sizes, and occlusion levels, factors that frequently occur in practical applications but remain largely underrepresented in existing benchmarks.  
These characteristics substantially enhance both the scenario realism and task difficulty of the dataset, thereby encouraging the development of more robust and generalizable visual grounding models.

\label{sec:RGBT-Groundbenchmark}

 Overall, RGBT-GroundBench introduces several key advances over existing visual grounding benchmarks:
\begin{itemize}
\item \textbf{Fine-grained multi-level annotations.} Covering scene-, environment-, and object-level attributes for comprehensive contextual understanding.
\item \textbf{Off-central object distribution.} Objects appear in more diverse and realistic spatial positions, better reflecting real-world perception challenges.
\item \textbf{Increased diversity of challenging samples.} Including a higher proportion of nighttime, low-light, and long-range instances for robust evaluation~(as shown in Table~\ref{tab:benchmarks_comparison}).
\item \textbf{Paired multi-modal visual RGBT images.} Enabling reliable grounding across complex real-world scenarios.
\end{itemize}



\subsection{Data Collection, Filtering and Captioning}

\noindent\textbf{Collection.} 
We begin by collecting multi-modal object detection datasets~\cite{liu2022target,hu2025ei,zhang2020multispectral} covering diverse real-world scenarios. As summarized in Tables~\ref{tab:benchmarks_comparison} and \ref{tab:RGBT-GroundBench_benchmark_data_split}, these data sources provide spatially aligned RGB and TIR image pairs, encompassing a wide spectrum of weather, lighting, and scene complexities. Each image pair is accompanied by object-level bounding box annotations, which ensure accurate multi-modal analysis and visual-language grounding.

\noindent\textbf{Filtering.} 
The collected data undergo a rigorous filtering process to ensure high quality and suitability for visual grounding research. 
We establish a comprehensive filtering pipeline covering three key aspects:
\begin{itemize}
    \item {Object visibility and scale:} 
We exclude excessively small targets that occupy only a few pixels, as they provide insufficient semantic information for reliable annotation and meaningful textual descriptions.  

\item {Cross-modality alignment accuracy:} 
The image pair will be removed if it exhibit significant spatial misalignment between the RGB and TIR modalities. 

\item {Category balance:} 
Long-tailed or underrepresented categories (\textit{e.g.}, \textit{dog} in FLIR and \textit{lamp} in M$^{3}$FD) are filtered out to alleviate class imbalance. 
\end{itemize}

For the retained samples, we further select the largest instance of each object category per image pair to enhance annotation clarity and preserve cross-modality consistency. 
This filtering process ensures that RGBT-GroundBench faithfully captures the complexity, diversity, and challenges inherent in real-world deployment scenarios.

\begin{wrapfigure}{r}{0.57\textwidth}
    \centering
    \vspace{-0.3cm}
    \includegraphics[width=\linewidth]{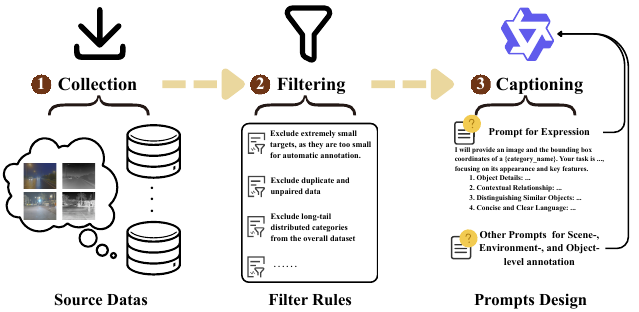}
    \vspace{-0.3cm}
    \caption{The pipeline of RGBT-GroundBench data preparation. Full pipeline details are provided in the supplementary material.}
    \vspace{-0.8cm}
    \label{fig:RGBT-GroundBench_MLLM_annotation_process}
\end{wrapfigure}

\noindent\textbf{Captioning.} 
As shown in Figure~\ref{fig:RGBT-GroundBench_MLLM_annotation_process}, we employ the Qwen3-VL to generate high-quality textual annotations in the captioning stage. 
Carefully designed prompts are used to produce two types of labels: 
(1)~Object referring expressions, (2)~Scene-level, environment-level, and object-level annotations. The specific prompt templates are provided in the \textit{supplementary material}.  
After automatic captioning, a hierarchical random sampling strategy is applied for human verification and refinement to ensure annotation accuracy and consistency.
As shown in Table~\ref{tab:RGBT-GroundBench_benchmark_distrbution_of_weather_light}, we further perform statistical validation to confirm the physical consistency of MLLM-generated labels, where lighting-weather distributions match real-world patterns.
This dual-level annotation scheme enriches RGBT-GroundBench samples with comprehensive semantic and contextual information, thereby supporting robust visual grounding across diverse and challenging environments.

\noindent\textbf{Annotation Quality Control.}
After automatic captioning, we apply hierarchical random sampling across scene, illumination, weather, object size, and occlusion for manual verification.
Annotators check hallucination, ambiguity, incorrect attributes, box-expression mismatch, and severe grammar issues, and correct them by rewrite/relabel or invalidation.
An annotation is accepted only when the expression uniquely identifies one target and is consistent with the final box and attributes across modalities.
Quality indicators are: 15\% instances reviewed, 5\% expressions edited, and 0.5\% removed as invalid; in the reviewed subset, 95\% of expressions correctly refer to the intended target.
As an additional double-check, Table~\ref{tab:RGBT-GroundBench_benchmark_distrbution_of_weather_light} shows plausible lighting-weather trends: sunny cases cluster in normal/strong light, while rainy/foggy cases cluster in weak/very-weak light, supporting fine-grained annotation consistency.


\subsection{Evaluation Protocol}
\label{subsec: evaluation protocol}
\noindent\textbf{Subset Split.}
As summarized in Table~\ref{tab:RGBT-GroundBench_benchmark_data_split}, all instances are divided into train, val, and test subsets following the source data split.
The test set is further partitioned into specialized subsets:
\textbf{TestA}, covering normal-size (NS) targets under normal-light (NL) and strong-light (SL) conditions;
\textbf{TestB}, focusing on nighttime scenarios with weak-light (WL) and very-weak-light (VL) environments; and
\textbf{TestC}, containing small-size (SS) targets.
Additional scene-, weather-, and occlusion-based splits and corresponding evaluation results will be provided in the supplementary material.
This strategy enables fine-grained evaluation of model performance across diverse difficulty levels and environmental conditions.

\noindent\textbf{Unified Evaluation Framework.}
As an integral part of RGBT-GroundBench, we build a unified implementation and evaluation infrastructure for benchmarking, rather than as a standalone model contribution.
The unified evaluation framework adapts representative grounding models (TransVG~\cite{deng2021transvg}, MMCA~\cite{yao2024visual}, D-MDETR~\cite{shi2023dynamic}, CLIP-VG~\cite{xiao2023clip}, FSVG~\cite{wang2025simple}, AttBalance~\cite{kang2025visual}, HiVG~\cite{xiao2024hivg}, and OneRef~\cite{xiao2024oneref}) to a single codebase supporting RGB-only, TIR-only, and RGB+TIR inputs.
All methods follow the same data preprocessing, training routines, and evaluation metrics, enabling fair comparison across modalities and architectures.
This benchmark infrastructure also serves as the code foundation for reproducible reporting and future benchmark-server integration.

\noindent\textbf{Release and Licensing.}
All benchmark metadata resources are publicly released for direct use under repository licenses. Original source datasets used by RGBT-GroundBench remain subject to their own licenses and usage terms.

\section{Reference Baseline Implementation: RGBT-VGNet}

\begin{figure*}[!t]
    \centering
    \includegraphics[width=0.95\linewidth]{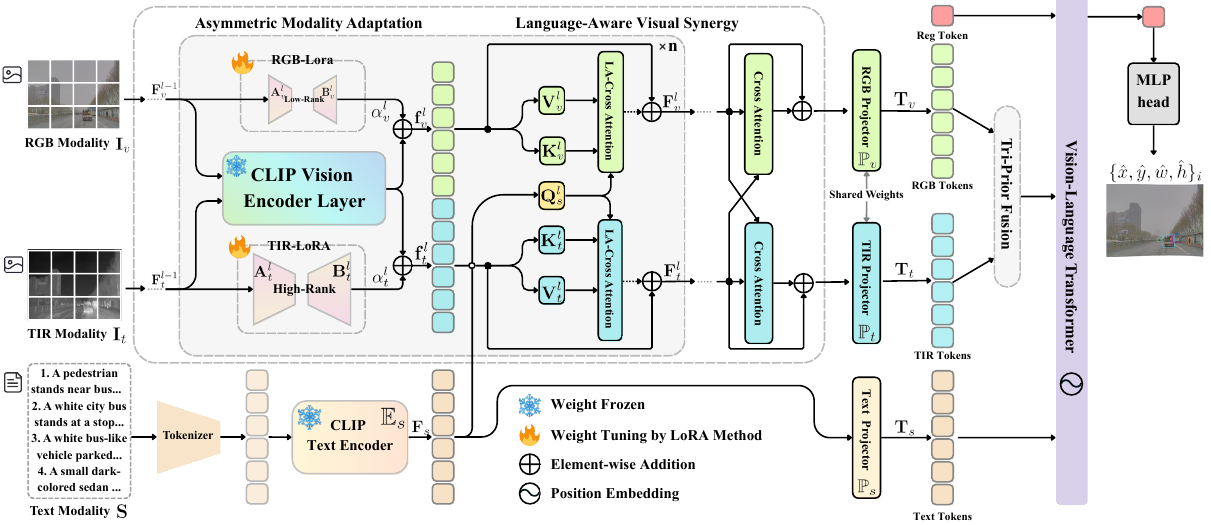}
    \vspace{-0.3cm}
    \caption{Architecture of the reference baseline method RGBT-VGNet.}
    \vspace{-0.6cm}
    \label{fig:RGBT-VGNet}
\end{figure*}

\noindent\textbf{Overall architecture.} This section presents implementation details of a reference baseline used for benchmark validation. Built on the unified evaluation framework in Sec.~\ref{subsec: evaluation protocol}, RGBT-VGNet provides a transparent and reproducible reference point under a unified protocol. Given an RGB--TIR pair and a referring expression, the model localizes the described object by integrating complementary cues from the two modalities. Specifically, RGBT-VGNet is built upon the CLIP model~\cite{radford2021learning} and uses three practical components: \ding{202}~Asymmetric Modality Adaptation~(AMA) for modality-adaptive visual feature learning, \ding{203}~Language-Aware Visual Synergy~(LAVS) for semantics-aware cross-modal interaction, and \ding{204}~Illumination-Local-Global Tri-Prior Fusion~(TPF) for reliability-aware cross-spectral fusion.

Formally, given an RGB image $I_{v}$, a TIR image $I_{t}$, and a referring expression 
$\mathbf{S} = \{\mathbf{s}_i\}_{i=1}^{T}$, the visual features at layer $l$ ($\mathbf{F}_{v}^{l}$ and $\mathbf{F}_{t}^{l}$) are extracted as:
\begin{equation}
\begin{aligned}
\mathbf{f}_{v}^{l} &=\begin{cases} 
\mathbb{E}_v^{l}(\mathbf{F}_{v}^{l-1}), & ~l > 1 \\ 
\mathbb{E}_v^{l}(\mathbf{I}_{v}), & ~l = 1
\end{cases},~ 
\mathbf{f}_{t}^{l} =\begin{cases} 
\mathbb{E}_t^{l}(\mathbf{F}_{t}^{l-1}), & ~l > 1 \\ 
\mathbb{E}_t^{l}(\mathbf{I}_{t}), & ~l = 1
\end{cases},
\end{aligned}
\end{equation}
where $\mathbb{E}_{v}$ and $\mathbb{E}_{t}$ denote the RGB and TIR CLIP vision encoders~(shared parameters) with AMA modules, and $l\in\{1,\dots,n\}$ is the encoder layer index.
The intermediate RGB and TIR features $\mathbf{f}_{v}^{l}$ and $\mathbf{f}_{t}^{l}$ are then refined by the LAVS module to produce modality-enhanced features $\mathbf{F}_{v}^{l}$ and $\mathbf{F}_{t}^{l}$.
The textual features are extracted as follows:
\begin{equation}
  \begin{aligned}
\mathbf{F}_{s}^n &= \mathbb{E}_s(\mathbf S), \quad \mathbf{T}_s = \mathbb {P}_s\left(\mathbf F_{s}^n\right),
\end{aligned}
\end{equation}
where $\mathbb{E}_{s}$ denotes the CLIP text encoder, and $\mathbb{P}_{s}$ is a linear projection that maps the last-layer textual embeddings into the grounding space.

All CLIP encoders remain frozen to preserve pretrained language-vision alignment. Extracted representations are aligned and refined by subsequent modules. The final input concatenates projected Fused Visual and Text tokens with a learnable regression token $[\mathrm{Reg}]$, which is then fed into the VL-Transformer~\cite{deng2021transvg}. A lightweight regression head predicts the bounding box from the $[\mathrm{Reg}]$ token:
\begin{equation}
\hat{\mathbf{B}} = {\hat{x},\hat{y},\hat{w},\hat{h}}_i = \mathrm{MLP}([\mathrm{Reg}]).
\end{equation}



 \noindent\ding{202}~\textbf{Asymmetric Modality Adaptation~(AMA).}
Vision-language pretrained models such as CLIP have strong performance in the RGB domain, but their feature representations are biased toward RGB data used during pretraining, resulting in a large modality gap when extended to TIR modality.
While LoRA-based~\cite{hu2022lora} fine-tuning has been widely adopted to adapt frozen backbones to downstream tasks, applying identical LoRA configurations to different modalities overlooks their differences and the VLP models' bias.
To address this limitation, we adopt an asymmetric LoRA configuration that assigns different adaptation capacities to the two visual modalities. Specifically, we perform low-rank decomposition on the attention projection weights of each vision encoder layer $l$:
\begin{equation}
\begin{aligned}
\mathbf{W}_{v}^l = \mathbf{W}_{\mathbb{E}_v}^l + \mathbf{\alpha}_v^l \mathbf{A}_v^l \mathbf{B}_v^l, \quad
\mathbf{W}_{t}^l = \mathbf{W}_{\mathbb{E}_t}^l + \mathbf{\alpha}_t^l \mathbf{A}_t^l \mathbf{B}_t^l,   
\end{aligned}
\end{equation}
where $\mathbf{W}_{\mathbb{E}_v}^l$ and $\mathbf{W}_{\mathbb{E}_t}^l$ are the frozen pre-trained CLIP weights (initialized from the same CLIP model), $\mathbf{A}_v^l \in \mathbb{R}^{d \times r_v}$, $\mathbf{A}_t^l \in \mathbb{R}^{d \times r_t}$, $\mathbf{B}_v^l \in \mathbb{R}^{r_v \times d}$, and $\mathbf{B}_t^l \in \mathbb{R}^{r_t \times d}$ are learnable low-rank matrices, and $\mathbf{\alpha}_v^l,\mathbf{\alpha}_t^l$ are scaling factors.
We set asymmetric ranks $r_v\leq r_t$ for RGB and TIR modalities, allowing the TIR branch to use higher adaptation capacity when needed.
We further adopt the hierarchical design~\cite{xiao2024hivg} as a practical setting for progressive alignment from low-level structural to high-level semantic representations.



 \noindent\ding{203}~\textbf{Language-Aware Visual Synergy (LAVS).}
In the RGBT visual grounding task, the target described by the referring expression usually appears in both RGB and TIR images.
Since grounding is inherently language-driven, we use a language-aware visual fusion module in which textual features act as semantic queries for cross-modal interaction.
Given the encoder outputs $\mathbf{f}_{v}^{l}$ and $\mathbf{f}_{t}^{l}$ from layer $l$ and the textual embedding $\mathbf{F}_{s}$, we first compute their Query-Key-Value projections and construct language-guided cross-modal attention:
\begin{equation}
\begin{aligned}
\mathbf{A}_{\text{{attn}}_{v}}^l = \mathrm{\sigma}\left( \frac{\mathbf{Q}_{s}^{l} (\mathbf K_{v}^{l})^T}{\sqrt{d}} \right), \quad  \mathbf{A}_{\text{{attn}}_{t}}^l = \mathrm{\sigma}\left( \frac{\mathbf Q_{s}^{l} (\mathbf K_{t}^{l})^T}{\sqrt{d}} \right),
\label{eq:5}
\end{aligned}
\end{equation}
where $\mathrm{\sigma}$ is $\mathrm{Softmax}$, and $\mathbf A_{\text{{attn}}_{v}}^l, \mathbf A_{\text{{attn}}_{t}}^l$ denote the attention matrices. 
\begin{equation}
\begin{aligned}
\mathbf F_{v}^{l} =\mathbf f_{v}^{l}+(\mathbf{A}_{\text{{attn}}_{v}}^l)^T (\mathbf{A}_{\text{{attn}}_{v}}^l \mathbf V_{v}^{l}), \quad
\mathbf F_{t}^{l} = \mathbf f_{t}^{l}+(\mathbf{A}_{\text{{attn}}_{t}}^l)^T (\mathbf{A}_{\text{{attn}}_{t}}^l \mathbf V_{t}^{l}).
\label{eq:6}
\end{aligned}
\end{equation}
Eq.~\ref{eq:5} and \ref{eq:6} enable textual semantics to guide the integration of visual cues from each modality, determining both which features should be fused and to what extent, while preserving feature dimensional consistency via left multiplication with the transpose of $\mathbf{A}_{\text{attn}}^l$. After obtaining the text-queried visual features $\mathbf F_{v}^{l}$ and $\mathbf F_{t}^{l}$, we apply cross-attention $\mathbb{CA}$ between the two modalities:
\begin{equation}
\begin{aligned}
\mathbf T_{v}^{l} = \mathbb P_v\left(\mathbf F_{v}^{l} + \mathbb{CA}(\mathbf F_{v}^{l}, \mathbf F_{t}^{l})\right),~~ 
\mathbf T_{t}^{l} = \mathbb P_t\left(\mathbf F_{t}^{l} + \mathbb{CA}(\mathbf F_{t}^{l}, \mathbf F_{v}^{l})\right)
\label{eq:7}
\end{aligned}
\end{equation}
where $\mathbb P_v$ and $\mathbb P_t$ denote RGB and TIR linear projections, respectively, mapping the output embeddings into a unified feature space for subsequent grounding.

\noindent\ding{204}~\textbf{Illumination-Local-Global Tri-Prior Fusion (TPF).}
TPF is a lightweight heuristic fusion module in our reference baseline, designed to instantiate Insight 2 (cross-spectral complementarity) under a controlled protocol rather than to introduce standalone illumination modeling. It takes the two modality-enhanced outputs from Eq.~\ref{eq:7}, $\mathbf T_{v}^{l}$ and $\mathbf T_{t}^{l}$, as inputs. We first split each stream into a CLS token and patch tokens, and then flatten patch maps into token sets $\{\mathbf t_{v,p}^{l}\}_{p=1}^{N}$ and $\{\mathbf t_{t,p}^{l}\}_{p=1}^{N}$.
The three priors are defined in two steps. We first compute the illumination prior:
$
i_p = \left[\mathcal{I}(\mathbf I_v)\right]_p, \quad
q_p = 1-\left|2i_p-1\right|,
$
where $\mathcal{I}(\cdot)$ denotes grayscale conversion, pooling to the patch grid, and min-max normalization to $[0,1]$; $i_p$ is the normalized illumination intensity of the RGB patch, and $q_p$ is the illumination quality that measures the proximity to the optimal illumination point (0.5).
Then we compute local-global semantic priors:
\begin{equation}
\begin{aligned}
g_p &= \sigma\!\left(\mathrm{MLP}_{\text{tok}}([\hat{\mathbf t}_{v,p}^{l},\hat{\mathbf t}_{t,p}^{l},|\hat{\mathbf t}_{v,p}^{l}-\hat{\mathbf t}_{t,p}^{l}|])\right),\\
\mathbf g^{\mathrm{ch}} &= \sigma\!\left(\mathrm{MLP}_{\text{ch}}([\bar{\mathbf t}_{v}^{l},\bar{\mathbf t}_{t}^{l},|\bar{\mathbf t}_{v}^{l}-\bar{\mathbf t}_{t}^{l}|])\right), \qquad
b = \mathrm{mean}_{c}(\mathbf g^{\mathrm{ch}}),
\end{aligned}
\label{eq:8b}
\end{equation}
where $\hat{\mathbf t}_{v,p}^{l},\hat{\mathbf t}_{t,p}^{l}$ are layer-normalized patch features from Eq.~\ref{eq:7}, and $\bar{\mathbf t}_{v}^{l},\bar{\mathbf t}_{t}^{l}$ are global pooled features. $\mathrm{mean}_{c}$ denotes channel-wise averaging. $b$ is a global prior shared across all patches (broadcast over $p$).
The RGB reliability weight is:
\begin{equation}
\begin{aligned}
w_p=\sigma\!\left((i_p-0.5)q_p + (g_p-0.5) + (b-0.5)\right),
\end{aligned}
\label{eq:9}
\end{equation}
where $0.5$ is the neutral reliability point for each normalized prior; we use equal weighting to avoid extra hyperparameters.
Finally, the fused patch token is:
\begin{equation}
\begin{aligned}
\mathbf z_p^{l} = w_p\mathbf t_{v,p}^{l} + (1-w_p)\mathbf t_{t,p}^{l}
+ \mathbf g^{\mathrm{ch}}\odot(\mathbf t_{v,p}^{l}-\mathbf t_{t,p}^{l}),
\end{aligned}
\label{eq:10}
\end{equation}
where $\mathbf g^{\mathrm{ch}}$ is the channel gate from Eq.~\ref{eq:8b}. This design provides adaptive RGB--TIR fusion with illumination-aware reliability control and local-global semantic compensation in a compact form.

\section{Experiments}
\label{sec:experiments}
\subsection{Experimental Setup}

\noindent\textbf{Implementation Details.} 
Our method is trained with a batch size of 8 using AdamW and a learning rate of $1\times10^{-3}$. Adapted baselines follow the default hyper-parameters in their original implementations where applicable. All experiments are trained for 120 epochs with the same input size ($224\times224$) and the same data augmentation policy on RTX 4090 GPUs. More detailed implementation settings can be found in the open-source unified evaluation framework.

\noindent\textbf{Evaluation Metrics.}
Following the previous visual grounding methods~\cite{chen2023advancing,xiao2024towards}, we adopt the \textbf{Acc@0.5} as the primary metric, measuring the localization accuracy when the intersection-over-union (IoU) between the predicted and ground-truth bounding boxes exceeds 0.5. In addition, we report results separately for val, test, testA, testB, and testC subsets (detailed in Section~\ref{subsec: evaluation protocol}) to evaluate model robustness under diverse conditions. \textit{Results for \textbf{Acc@0.7} are provided in the supplementary material for stricter performance validation.}

\subsection{Evaluation on the RGBT-GroundBench}
\noindent\textbf{Pretrained model zero-shot transfer setting.}
All models are evaluated by directly loading pretrained weights without fine-tuning on the RGBT-GroundBench training split. As shown in Table~\ref{tab:4}, all methods exhibit low localization accuracy under both RGB and TIR modalities. This is mainly because the pretrained weights are trained on RefCOCO, ReferIt, and Flickr30K datasets, which contain limited scene complexity and lack environmental diversity, making zero-shot transfer difficult. Moreover, since these datasets contain only RGB modality, performance further degrades on TIR, where zero-shot results drop most drastically. Compared with subsequently trained models, zero-shot Acc\texttt{@}0.5 shows an average degradation of 30\% on RGB and 50\% on TIR modality, demonstrating that effective cross-modal adaptation is necessary for reliable RGBT visual grounding.

\begin{table*}[t]
\centering
\caption{Performance evaluation of current representative uni-modal visual grounding works and the baseline multi-modal RGBT visual grounding method on the RGBT-GroundBench. The ``-'entries correspond to models that fail to produce stable bounding boxes. Complete evaluation results will be included in the supplementary materials.}

    \vspace{-0.3cm}
\label{tab:4}
\resizebox{\textwidth}{!}{
\begin{tabular}{c|c|c|c|ccccc|ccccc|ccccc}
\toprule
\multirow{2}{*}{\textbf{Methods}} & \multirow{2}{*}{\textbf{Venue}} & \multirow{2}{*}{\begin{tabular}[c]{@{}c@{}}\textbf{Visual / Language}\\\textbf{Backbone}\end{tabular}} &\multirow{2}{*}{\begin{tabular}[c]{@{}c@{}}\textbf{Visual}\\\textbf{Modality}\end{tabular}} &  \multicolumn{5}{c|}{\textbf{RefFLIR}} & \multicolumn{5}{c|}{\textbf{RefM$^{3}$FD}} & \multicolumn{5}{c}{\textbf{RefMFAD}} \\

 & & & &val&test&testA&testB&testC&val&test&testA&testB&testC&val&test&testA&testB&testC\\
\midrule
\multicolumn{8}{l}{\textbf{a. Zero-shot transfer from RGB-pretrained models:}} \\\midrule

CLIP-VG \textsuperscript{~\cite{xiao2023clip}} & TMM'23 & CLIP-B / CLIP-B & \multirow{5}{*}{{RGB}} & 5.59& 8.12& 15.51& 5.00& 1.11& 7.74& 8.55& 16.31& 9.05& 1.35& 7.48& 7.48& 16.58& 6.93& 1.22\\

HiVG-B \textsuperscript{~\cite{xiao2024hivg}} & ACMMM'24 & CLIP-B / CLIP-B && 23.03& 34.27& 61.69& 30.16& 2.23& 29.17& 31.22& 58.28& 34.37& 5.30& 18.71& 19.62& 41.14& 19.90& 1.92\\
HiVG-L \textsuperscript{~\cite{xiao2024hivg}} & ACMMM'24 & CLIP-L / CLIP-L && 34.05& 44.46& 78.88& 33.13& 7.81& 52.98& 53.10& 87.66& 56.86& 19.59& 40.13& 40.42& 82.15& 37.68& 12.86\\
OneRef-B\textsuperscript{~\cite{xiao2024oneref}}  & NeurIPS'24 & BEIT3-B / BEIT3-B &  & 32.89& 41.46& 77.68& 31.56& 3.55& 33.93& 39.65& 74.11& 48.63& 3.19& 29.70& 30.05& 70.63& 28.14& 2.86\\
OneRef-L\textsuperscript{~\cite{xiao2024oneref}} & NeurIPS'24 & BEIT3-L / BEIT3-L  & & 38.82& 49.20& 80.19& 36.09& 17.04& 46.63& 46.82& 79.55& 52.83& 14.72& 37.50& 36.89& 73.04& 35.15& 12.35\\
\cline{4-4}
CLIP-VG \textsuperscript{~\cite{xiao2023clip}} & TMM'23 & CLIP-B / CLIP-B &\multirow{5}{*}{{TIR}}  & 4.77& 8.03& 14.92& 6.41& 0.61& 2.38& 6.59& 9.33& 11.15& 0.76& 4.54& 4.26& 9.49& 4.04& 5.49\\
HiVG-B \textsuperscript{~\cite{xiao2024hivg}} & ACMMM'24 & CLIP-B / CLIP-B && 10.86& 12.10& 20.64& 15.94& 0.81& 7.74& 11.49& 18.34& 18.19& 0.97& 6.29& 7.36& 17.97& 6.44& 0.51\\
HiVG-L \textsuperscript{~\cite{xiao2024hivg}} & ACMMM'24 & CLIP-L / CLIP-L& & 18.26& 23.80& 43.56& 24.06& 0.00& 24.40& 28.89& 55.36& 36.47& 2.92& 23.73& 23.11& 56.33& 20.39& 2.82\\
OneRef-B\textsuperscript{~\cite{xiao2024oneref}}  & NeurIPS'24 & BEIT3-B / BEIT3-B &  & 21.38& 26.57& 49.28& 23.75& 0.20& 17.26& 22.83& 40.67& 32.91& 0.11& 15.13& 15.60& 41.90& 13.50& 0.12\\
OneRef-L\textsuperscript{~\cite{xiao2024oneref}} & NeurIPS'24 & BEIT3-L / BEIT3-L &  & 21.55& 27.00& 46.42& 26.56& 3.65& 23.21& 23.65& 38.15& 33.55& 4.17& 17.52& 17.81& 41.14& 16.39& 2.43\\
\midrule
\multicolumn{8}{l}{\textbf{b. In-domain training with uni-modal visual input (RGB or TIR):}} \\ \midrule

TransVG  \textsuperscript{~\cite{deng2021transvg}} & ICCV'21 & RN50+DETR / BERT-B &  \multirow{11}{*}{{RGB}}& 50.74& 42.16& 62.25& 34.89& 19.76& 39.29& 40.08& 63.42& 46.03& 15.10& 52.27& 51.94& 82.64& 49.35& 31.06\\

CLIP-VG \textsuperscript{~\cite{xiao2023clip}} & TMM'23 & CLIP-B / CLIP-B  & & 43.68& 42.77& 67.14& 33.80& 15.81& 32.14& 33.76& 58.80& 41.83& 5.39& 36.52& 37.47& 73.13& 34.92& 13.16\\
D-MDETR \textsuperscript{~\cite{shi2023dynamic}} & TPAMI'24 & CLIP-B / CLIP-B  & & 57.07& 49.25& 70.17& 39.22& 26.77& 52.38& 50.19& 71.67& 57.04& 26.08& 60.59& 59.39& 86.71& 55.79& 40.98\\
D-MDETR \textsuperscript{~\cite{shi2023dynamic}} & TPAMI'24 & RN50+DETR / BERT-B & & 49.67& 44.28& 67.14& 34.38& 20.95&-&-&-&-&-& 51.99& 51.18& 82.07& 48.65& 30.41\\
MMCA \textsuperscript{~\cite{yao2024visual}} & ACMMM'24 & RN50+DETR / BERT-B &  & 50.16& 46.57& 71.48& 35.63& 21.50&-&-&-&-&-& 47.85& 46.52& 76.71& 43.84& 26.24\\
FSVG \textsuperscript{~\cite{wang2025simple}} & ICME'25 & RN50+DETR / BERT-B & & 56.09 &45.25&67.14 &36.88&20.55&49.40&49.39&73.86&54.65&23.70&59.71&58.65&86.24&55.34&39.81\\
AttBalance \textsuperscript{~\cite{kang2025visual}} & ACMMM'25 & CLIP-B / CLIP-B & & 51.64&49.10&73.81&40.16&21.76&-&-&-&-&-&51.04&51.30&83.92&48.74&29.61 \\
HiVG-B \textsuperscript{~\cite{xiao2024hivg}} & ACMMM'24 & CLIP-B / CLIP-B  & & 69.08& 66.65& 88.81& 54.06& 43.52& 69.24& 68.15& 89.40& 56.09& 45.04& 65.45& 64.02& 90.66& 60.52& 45.57\\

HiVG-L \textsuperscript{~\cite{xiao2024hivg}} & ACMMM'24 & CLIP-L / CLIP-L&  & 68.75& 71.13& 90.69& 60.63& 50.00& 65.48& 67.80& 94.48& 70.11& 41.29& 64.25& 63.72& \underline{\textbf{91.52}}& 60.97& 44.98\\
OneRef-B\textsuperscript{~\cite{xiao2024oneref}} & NeurIPS'24 & BEIT3-B / BEIT3-B  &  & 63.82& 61.69& 84.73& 52.34& 35.90& 66.07& 66.66& \underline{\textbf{94.64}}& 70.93& 39.07& 62.34& 60.61& 90.76& 57.63& 40.86\\
OneRef-L\textsuperscript{~\cite{xiao2024oneref}} & NeurIPS'24 & BEIT3-L / BEIT3-L & & 66.61& 64.60& 88.31& 53.13& 39.76& 63.10& 64.33& 93.75& 73.02& 34.52& 64.49& 63.15& 90.00& 59.46& 45.18\\
\cline{4-4}

TransVG  \textsuperscript{~\cite{deng2021transvg}} & ICCV'21 & RN50+DETR / BERT-B &  \multirow{11}{*}{{TIR}}& 49.92& 42.63& 62.37& 35.67& 21.18& 45.83& 47.54& 69.72& 55.58& 22.51& 52.79& 52.33& 83.92& 49.84& 31.37\\

CLIP-VG \textsuperscript{~\cite{xiao2023clip}} & TMM'23 & CLIP-B / CLIP-B &  & 36.59& 37.18& 59.38& 34.11& 12.26& 23.81& 27.16& 43.55& 38.63& 4.92& 32.06& 30.79& 61.85& 28.81& 10.31\\

D-MDETR \textsuperscript{~\cite{shi2023dynamic}} & TPAMI'24 & CLIP-B / CLIP-B  &  & 48.36& 38.87& 55.83& 37.66& 18.38& 47.62& 50.34& 70.13& 57.68& 28.52& 56.37& 54.52& 79.72& 51.63& 37.73\\

D-MDETR \textsuperscript{~\cite{shi2023dynamic}} & TPAMI'24 & RN50+DETR / BERT-B &  & 50.57& 42.96& 62.01& 42.52& 20.36&-&-&-&-&-& 50.60& 50.01& 80.10& 47.88& 29.53\\

MMCA \textsuperscript{~\cite{yao2024visual}}  & ACMMM'24 & RN50+DETR / BERT-B &  & 48.52& 41.41& 60.98& 40.00& 18.46&-&-&-&-&-& 46.26& 45.22& 76.33& 42.86& 24.43\\
FSVG \textsuperscript{~\cite{wang2025simple}} & ICME'25 & RN50+DETR / BERT-B & & 47.86 &38.50&54.64&37.81&17.13 &52.98 &51.48&72.24&60.31 &27.00 & 54.54& 54.52&79.92 &52.81 &36.05\\
AttBalance \textsuperscript{~\cite{kang2025visual}} & ACMMM'25 & CLIP-B / CLIP-B & &43.75&43.39&66.83&39.22&15.92&33.93&37.63&64.44&45.80&8.77&45.30&45.45&78.61&43.35&22.43 \\


HiVG-B\textsuperscript{~\cite{xiao2024hivg}} & ACMMM'24 & CLIP-B / CLIP-B  &  & 68.75& 64.07& 81.55& 66.09& 42.41& 68.09& 61.77& 78.21& 68.31& 40.59& 63.22& 62.63& 87.50& 60.85& 45.69\\

HiVG-L \textsuperscript{~\cite{xiao2024hivg}} & ACMMM'24 & CLIP-L / CLIP-L& & 65.79& 66.71& 85.44& 68.44& 43.91& 61.31& 65.12& 87.34& 73.95& 40.69& 59.87& 58.87& 85.95& 57.18& 39.33\\

OneRef-B\textsuperscript{~\cite{xiao2024oneref}} & NeurIPS'24 & BEIT3-B / BEIT3-B &   & 62.01& 60.00& 80.55& 60.00& 35.80& 60.12& 65.10& 87.74& 73.03& 41.02& 58.60& 57.95& 86.46& 55.71& 38.67\\

OneRef-L\textsuperscript{~\cite{xiao2024oneref}} & NeurIPS'24 & BEIT3-L / BEIT3-L  & & 64.14& 61.17& 81.15& 60.47& 37.53& 65.48& 68.30& 90.26& 75.32& 45.29& 62.34& 60.09& 85.70& 57.54& 42.67\\ 

\midrule
\multicolumn{8}{l}{\textbf{c. In-domain training with multi-modal visual input (RGB+TIR):}} \\\midrule

MV-TransVG   &ICCV'21& RN50+DETR / BERT-B & \multirow{12}{*}{RGB+TIR}& 54.44& 46.06& 68.62& 40.31& 21.10& 45.83& 48.04& 70.29& 55.3& 22.89& 53.18& 53.84& 83.54& 51.14& 33.84\\
MV-CLIP-VG  &TMM'23& CLIP-B / CLIP-B & & 45.57& 46.01& 72.62& 41.56& 16.06& 34.52& 38.92& 63.42& 50.05& 10.01& 47.02& 47.52& 82.76& 45.07& 23.96\\
MV-D-MDETR  &TPAMI'24 & CLIP-B / CLIP-B &  & 55.26& 47.42& 68.33& 41.25& 23.38& 48.81& 46.50& 65.91& 53.82& 24.19& 58.76& 57.31& 83.84& 53.79& 39.30\\
MV-D-MDETR  &TPAMI'24 & RN50+DETR / BERT-B&  & 55.01& 47.84& 68.10& 43.77& 24.52& 44.64& 45.62& 67.23& 53.88& 21.81& 54.73& 54.13& 82.89& 52.00& 34.24\\

MV-MMCA   & ACMMM'24& RN50+DETR / BERT-B && 54.93& 48.97& 72.43& 42.34& 22.62& 46.43& 47.83& 72.16& 56.76& 20.45& 53.62& 54.41& 84.41& 51.47& 34.16\\
MV-FSVG &ICME'25& RN50+DETR / BERT-B & & 54.61 & 47.96 &68.57 &40.16&24.29&47.62&46.32&64.61&54.74&25.00&58.44&58.60&85.86&55.18&40.05\\
MV-AttBalance & ACMMM'25& CLIP-B / CLIP-B & & 55.26&54.04&76.07&48.59&27.63&46.43&50.50&79.30&58.39&20.40&52.71&52.69&82.58&50.94&31.74\\

MV-HiVG-B  & ACMMM'24& CLIP-B / CLIP-B  & & \underline{\textbf{75.33}}& 69.20& 85.80& 66.72& 50.20& 69.64& 72.35& 93.99& 79.43& 49.40& 67.07& 67.04& 91.01& 64.52& \underline{\textbf{51.53}}\\
MV-HiVG-L  & ACMMM'24& CLIP-L / CLIP-L  & & 71.05& \underline{\textbf{72.50}}& \underline{\textbf{91.07}}& \underline{\textbf{70.17}}& 51.16& 63.16& 68.67& 89.52& 63.28& 44.74& 61.78& 61.31& 88.38& 58.85& 42.36\\
MV-OneRef-B  &NeurIPS'24& BEIT3-B / BEIT3-B  & & 63.82& 62.05& 86.19& 57.34& 35.43& 62.50& 63.98& 89.20& 69.34& 37.93& 61.62& 59.97& 89.25& 56.85& 41.58\\
MV-OneRef-L  &NeurIPS'24& BEIT3-L / BEIT3-L && 66.61& 60.89& 84.01& 53.75& 35.09& 67.26& 69.70& 92.78& 75.14& 46.27& 64.57& 62.73& 88.99& 59.42& 45.25\\
RGBT-VGNet &Ours& CLIP-B / CLIP-B  & & \underline{\textbf{75.33}}& 72.43& 89.64& 69.22& \underline{\textbf{52.53}}& \underline{\textbf{73.21}}& \underline{\textbf{74.89}}& 94.16& \underline{\textbf{80.75}}& \underline{\textbf{55.52}}& \underline{\textbf{68.31}}& \underline{\textbf{67.58}}& 91.29& \underline{\textbf{64.80}}& 51.18\\

\bottomrule
\end{tabular}
}
    \vspace{-0.6cm}
\end{table*}

\noindent\textbf{Single source dataset training setting.}
All models are trained under our unified framework with identical settings. To combat the deterioration of RGB-modality visual grounding in low-light conditions and small objects conditions, we introduce the TIR modality to provide illumination-invariant cues and thereby enhance overall robustness. In this setting, our RGBT-VGNet achieves highest performance across all three sub-datasets, benefiting from the complementary information offered by RGB--TIR fusion. Uni-modality visual grounding baselines such as HiVG~\cite{xiao2024hivg} and OneRef~\cite{xiao2024oneref} also show competitive results within their respective settings. In addition, all models trained on the target dataset exhibit superior improvements over their zero-shot versions, confirming the inherent difficulty of the RGBT-GroundBench.

\noindent\textbf{Multi-modal RGBT visual grounding.}
As shown in Table~\ref{tab:4}, multi-modal visual input consistently improves performance across all models, with an average improvement of approximately 10\% in Acc\texttt{@}0.5 over uni-modal settings. All models exhibit noticeable improvements on testB and testC, demonstrating the advantages of multimodal visual input under small objects and low-light conditions. In particular, our RGBT-VGNet achieves the highest performance across all subsets, with Acc\texttt{@}0.5 exceeding 91\%, 64\%, and 49\% on testA, testB, and testC, respectively, further validating its robustness and generalization under diverse lighting and object-size conditions. \textit{More evaluation results are included in the supplementary materials.}

\noindent\textbf{MLLM Evaluation.}
We additionally evaluate several representative multi-modal large language models (MLLMs), including GLM-4.6V, Kimi-K2.5, and Qwen3.5-Plus, under both $224\times224$ and original-resolution settings. \textit{The detailed experimental results and analysis are reported in the supplementary material.}

\subsection{Ablation on RGBT-VGNet Components}
\begin{table*}[t]
    \centering
    \caption{\textbf{Ablation of AMA, LAVS, and TPF in RGBT-VGNet.} This table reports the effect of each practical component in the reference baseline under the same protocol. Best results are bold and underlined.}
    
    \vspace{-0.3cm}
    \resizebox{\textwidth}{!}{
    \begin{tabular}{c|c|c|ccccc|ccccc|ccccc}
        \toprule
        \multirow{2}{*}{AMA}&\multirow{2}{*}{LAVS}&\multirow{2}{*}{TPF}& \multicolumn{5}{c|}{\textbf{RefFLIR}} & \multicolumn{5}{c|}{\textbf{RefM$^{3}$FD}} & \multicolumn{5}{c}{\textbf{RefMFAD}} \\
&&& \textbf{val} & \textbf{test}& \textbf{testA} & \textbf{testB} & \textbf{testC}& \textbf{val} & \textbf{test}& \textbf{testA} & \textbf{testB} & \textbf{testC}&\textbf{val} & \textbf{test}& \textbf{testA} & \textbf{testB}& \textbf{testC} \\
 \midrule
& &&55.42&46.19&67.42&41.56&22.81&54.16&57.57&78.89&65.63&34.46&53.50&54.63&81.77&51.91&36.62 \\
\checkmark &&&74.01&71.17&90.57&66.25&49.69&70.83&72.53&93.99&80.16&50.43&68.07&65.27&90.63&62.19&47.96\\
\checkmark &\checkmark& &73.68&\underline{\textbf{72.65}}&\underline{\textbf{91.31}}&67.19&52.23&\underline{\textbf{73.21}}&74.34&\underline{\textbf{94.72}}&\underline{\textbf{81.93}}&53.63&67.83&66.62&91.16&64.07&49.76\\
\checkmark &\checkmark&\checkmark &\underline{\textbf{75.33}}&72.43&89.64&\underline{\textbf{69.22}}&\underline{\textbf{52.53}}&\underline{\textbf{73.21}}&\underline{\textbf{74.89}}&94.16&80.75&\underline{\textbf{55.52}}&\underline{\textbf{68.31}}&\underline{\textbf{67.58}}&\underline{\textbf{91.29}}&\underline{\textbf{64.80}}&\underline{\textbf{51.18}} \\
    \bottomrule
    \end{tabular}
    }
        \vspace{-0.3cm}
    \label{tab:ablation_components}
\end{table*}

\begin{table}[t]
\centering
\footnotesize
\caption{Comparison of different RGBT feature fusion method.}

    \vspace{-0.3cm}
\setlength{\tabcolsep}{1.0mm}
\renewcommand{\arraystretch}{1.1}
\resizebox{\textwidth}{!}{
\begin{tabular}{c|c|ccccc|ccccc|ccccc}
\toprule
\multirow{2}{*}{\begin{tabular}[c]{@{}c@{}}\textbf{Fusion}\\\textbf{Method}\end{tabular}}& \multirow{2}{*}{\textbf{Venue}} & \multicolumn{5}{c|}{\textbf{RefFLIR}} & \multicolumn{5}{c|}{\textbf{RefM$^{3}$FD}} & \multicolumn{5}{c}{\textbf{RefMFAD}} \\
& & \textbf{val} & \textbf{test} & \textbf{testA} & \textbf{testB} & \textbf{testC} & \textbf{val} & \textbf{test} & \textbf{testA} & \textbf{testB} & \textbf{testC} & \textbf{val} & \textbf{test} & \textbf{testA} & \textbf{testB} & \textbf{testC} \\
\midrule
Naive Add  & - & 73.68 & 69.39 & 87.98 & 66.72 & 46.86 & 70.24 & 73.04 & \underline{\textbf{94.40}} & 80.20 & 51.46 &67.91  &66.16  & 90.53 & 63.09 &49.41\\
CMX  & TITS'24 & 64.31 & 63.93 & 85.36 & 60.63 & 38.87 & 60.71&64.01&89.45&71.90&38.15 & 53.74 &53.77  & 85.48 & 50.45 &32.33  \\
LIF  & ICCV'25 & 74.34 & 71.39  & \underline{\textbf{90.00}} & 67.50 & 49.70 & 69.64 &72.96  & 93.99 & 80.47 & 51.30  & 67.59 & 66.33 & 90.28&63.66  &49.96  \\
\textbf{TPF} & \textbf{Ours} &\underline{\textbf{75.33}}&\underline{\textbf{72.43}}&89.64&\underline{\textbf{69.22}}&\underline{\textbf{52.53}}&\underline{\textbf{73.21}}&\underline{\textbf{74.89}}&94.16&\underline{\textbf{80.75}}&\underline{\textbf{55.52}}&\underline{\textbf{68.31}}&\underline{\textbf{67.58}}&\underline{\textbf{91.29}}&\underline{\textbf{64.80}}&\underline{\textbf{51.18}} \\
\bottomrule
\end{tabular}
}
    \vspace{-0.2cm}
\label{tab:ablation_fusion_strategies}
\end{table}

\begin{figure*}[!t]
    \centering
    \includegraphics[width=\linewidth]{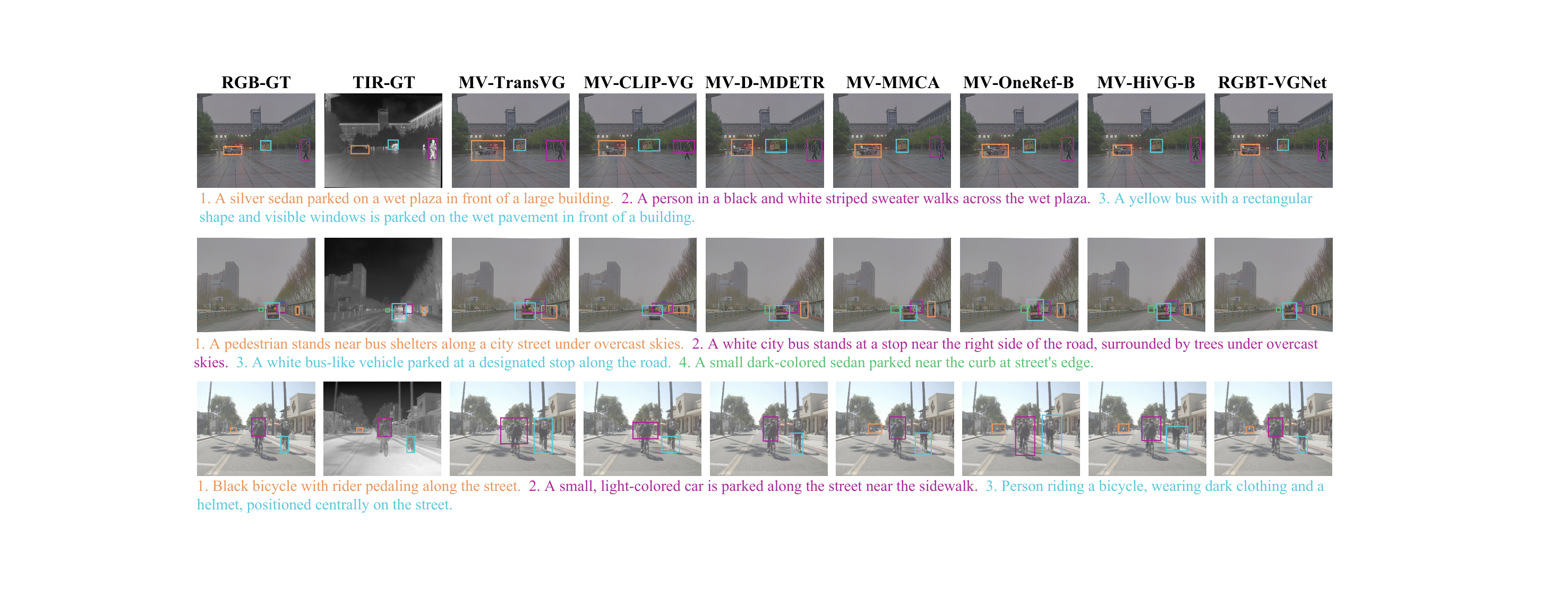}
    \caption{Qualitative comparison of RGBT visual grounding on RGBT-GroundBench. Bounding box colors follow the legend below.}
    \vspace{-0.6cm}
    \label{fig:Qualitative Analysis} 
\end{figure*}

Table~\ref{tab:ablation_components} examines the effect of each component in the reference baseline. AMA brings the largest single-step gain over the no-module setting across all datasets and splits, indicating that modality adaptation is a prerequisite for stable RGB--TIR fusion. Adding LAVS on top of AMA further improves most subsets, with clear benefits on difficult splits (testB/testC), suggesting that language-guided cross-modal interaction improves robustness under weak visibility and small-target conditions. Adding TPF on top of AMA+LAVS yields additional gains on most hard subsets, especially RefFLIR-testB/testC, RefM$^{3}$FD-testC, and RefMFAD-testC, which is consistent with its illumination-aware design.

Table~\ref{tab:ablation_fusion_strategies} compares different RGBT fusion strategies under the same evaluation protocol. The proposed TPF achieves the strongest overall performance, obtaining the highest average accuracy across the three sub-datasets. Although simpler visual fusion methods such as Naive Add and LIF outperform TPF in a few specific cases, their performance is less consistent across datasets. In addition, the representative RGBT fusion baseline CMX shows a noticeable performance gap compared with the other approaches. Overall, the results indicate that the language-aware fusion design in TPF provides more robust cross-modal alignment, leading to stronger and more stable grounding performance.

\subsection{Qualitative Comparisons and Analysis} 
The qualitative comparison of multi-modal RGBT visual grounding models under the RGBT-GroundBench protocol is shown in Figure~\ref{fig:Qualitative Analysis}. The objects are in various real-world conditions, including post-rain, foggy, and sunny weather, which presents challenges for models with different lighting, visibility, and object sizes. All results are evaluated using our framework adapted to the RGB--TIR modality. The results reveal that uni-modal visual grounding methods struggle with small or distant targets, often producing overly large or inaccurate boxes. In contrast, RGBT-VGNet consistently delivers precise and semantically consistent detections for humans and vehicles, effectively leveraging complementary cues from RGB and Thermal modalities under textual guidance. This demonstrates its robustness and generalization in various complex scenarios.

\section{Discussion and Conclusion}
\label{sec:discussion}

In this paper, we present RGBT-GroundBench for RGB-Thermal visual grounding in complex real-world environments, which contains over 40K images (21,535 RGB-TIR pairs) and 38,760 object instances with fine-grained annotations in various complex scenes. By conducting large-scale experiments to evaluate 11 VG models under the proposed RGBT-GroundBench protocol, we draw some important findings, which are summarized below:
\ding{202} \textbf{Scene Complexity Correlates with Grounding Accuracy.}
VG models are stable in simpler environments (see Table~\ref{tab:4}-testA) but degrade in cluttered scenes (\textit{Supplementary} Tables~\ref{tab:scene_refflir_all_methods},~\ref{tab:scene_refm3fd_all_methods},~\ref{tab:scene_refmfad_all_methods},~\ref{tab:size_all_methods} and Fig.~\ref{fig:Qualitative Analysis}), especially with distractors, occlusion, or small off-center targets. This suggests sensitivity to background complexity and object density, potentially amplified by dataset biases (e.g., salient/centered objects). Overall, realistic grounding demands stronger context modeling and more robust disambiguation under clutter and occlusion.

\ding{203} \textbf{LoRA-based Adaptation Enhances Robustness.} LoRA-based grounding models show more stable behavior across diverse scenes~(see Table~\ref{tab:4} and Table~\ref{tab:ablation_components}). This suggests that parameter-efficient adaptation helps align pretrained vision-language representations with grounding objectives under domain shifts and challenging environments. The results highlight that adaptation strategies can play an important role in improving robustness beyond architectural design.


\ding{204} \textbf{Low Illumination Exposes Robustness Gap.}
Low-light conditions consistently degrade performance (see Table~\ref{tab:4}-testB and \textit{Supplementary} Table~\ref{tab:illum_all_methods}), despite being underexplored in prior benchmarks. This suggests that illumination variation is a major failure mode for models trained on well-lit RGB imagery. Thermal cues help under low illumination, but grounding remains difficult, especially for small targets or adverse weather (see Table~\ref{tab:4}-testC and \textit{Supplementary} Tables~\ref{tab:weather_all_methods} and \ref{tab:size_all_methods}).

\ding{205} \textbf{MLLMs Struggle with Cross-Spectral Grounding.}
Results from the MLLM evaluation (see \textit{Supplementary} Tables~\ref{tab:mllm_224_supp}~and~\ref{tab:mllm_ori_supp}) indicate that naively adding thermal inputs does not yield consistent gains in localization accuracy, and performance varies markedly with spatial resolution. This suggests that existing MLLMs still struggle to reliably fuse cross-spectral cues for precise spatial grounding.

Based on these observations, we propose RGBT-VGNet as a baseline model to achieve satisfactory performance in complex scenarios covered by RGBT-GroundBench. Overall, we hope our comprehensive benchmark, insightful findings, and competitive baseline method can be helpful for evaluating the performance of VG models in various complex scenes and improving their grounding accuracy in the future.


\section*{Acknowledgement}

This work was supported in part by the Project of the National Natural Science Foundation of China under Grant 62576020, in part by the Fundamental Research Funds for the Central Universities and in part by the China Postdoctoral Science Foundation under Grant Number 2025M784289.

%
%
\bibliographystyle{splncs04}
\bibliography{main}

@String(CVPR= {IEEE Conf. Comput. Vis. Pattern Recog.})

@String(ICCV= {Int. Conf. Comput. Vis.})

@String(ECCV= {Eur. Conf. Comput. Vis.})

@String(TMM  = {IEEE Trans. Multimedia})

@String(ICME = {Int. Conf. Multimedia and Expo})

@String(ICIP = {IEEE Int. Conf. Image Process.})

@String(ICLR = {Int. Conf. Learn. Represent.})

@String(CVPR  = {CVPR})

@String(ICCV  = {ICCV})

@String(ECCV  = {ECCV})

@String(TCSVT = {IEEE TCSVT})

@String(TMM   =	{IEEE TMM})

@String(ICME  =	{ICME})

@String(ICIP  = {ICIP})

@String(ICLR  = {ICLR})

@article{zhao2023mitigating,
  title={Mitigating modality discrepancies for RGB-T semantic segmentation},
  author={Zhao, Shenlu and Liu, Yichen and Jiao, Qiang and Zhang, Qiang and Han, Jungong},
  journal={IEEE Transactions on Neural Networks and Learning Systems},
  volume={35},
  number={7},
  pages={9380--9394},
  year={2023},
  publisher={IEEE}
}

@inproceedings{kang2025your,
  title={Your large vision-language model only needs a few attention heads for visual grounding},
  author={Kang, Seil and Kim, Jinyeong and Kim, Junhyeok and Hwang, Seong Jae},
  booktitle={Proceedings of the Computer Vision and Pattern Recognition Conference},
  pages={9339--9350},
  year={2025}
}

@inproceedings{wu2025f,
  title={F-lmm: Grounding frozen large multimodal models},
  author={Wu, Size and Jin, Sheng and Zhang, Wenwei and Xu, Lumin and Liu, Wentao and Li, Wei and Loy, Chen Change},
  booktitle={Proceedings of the IEEE/CVF Conference on Computer Vision and Pattern Recognition},
  pages={24710--24721},
  year={2025}
}

@inproceedings{zhao2025rethinking,
  title={Rethinking multi-modal object detection from the perspective of mono-modality feature learning},
  author={Zhao, Tianyi and Liu, Boyang and Gao, Yanglei and Sun, Yiming and Yuan, Maoxun and Wei, Xingxing},
  booktitle={Proceedings of the IEEE/CVF International Conference on Computer Vision},
  pages={6364--6373},
  year={2025}
}

@inproceedings{kang2025visual,
  title={Visual grounding with attention-driven constraint balancing},
  author={Kang, Weitai and Zhou, Luowei and Wu, Junyi and Sun, Changchang and Yan, Yan},
  booktitle={Proceedings of the 33rd ACM International Conference on Multimedia},
  pages={1637--1645},
  year={2025}
}

@inproceedings{wang2025simple,
  title={A Simple and Better Baseline for Visual Grounding},
  author={Wang, Jingchao and Zhang, Wenlong and Huang, Dingjiang and Wang, Hong and Zheng, Yefeng},
  booktitle={2025 IEEE International Conference on Multimedia and Expo (ICME)},
  pages={1--6},
  year={2025},
  organization={IEEE}
}

@inproceedings{yuan2025unirgb,
  title={Unirgb-ir: A unified framework for visible-infrared semantic tasks via adapter tuning},
  author={Yuan, Maoxun and Cui, Bo and Zhao, Tianyi and Wang, Jiayi and Fu, Shan and Yang, Xue and Wei, Xingxing},
  booktitle={ACM MM},
  pages={2409--2418},
  year={2025}
}

@article{yuan2024c2former,
  title={C$^2$former: Calibrated and complementary transformer for rgb-infrared object detection},
  author={Yuan, Maoxun and Wei, Xingxing},
  journal={TGRS},
  volume={62},
  pages={1--12},
  year={2024},
  publisher={IEEE}
}

@article{yuan2024improving,
  title={Improving RGB-infrared object detection with cascade alignment-guided transformer},
  author={Yuan, Maoxun and Shi, Xiaorong and Wang, Nan and Wang, Yinyan and Wei, Xingxing},
  journal={Information Fusion},
  volume={105},
  pages={102246},
  year={2024},
  publisher={Elsevier}
}

@article{zhao2025removal,
  title={Removal then selection: A coarse-to-fine fusion perspective for RGB-infrared object detection},
  author={Zhao, Tianyi and Yuan, Maoxun and Jiang, Feng and Wang, Nan and Wei, Xingxing},
  journal={IEEE Transactions on Intelligent Transportation Systems},
  year={2025},
  publisher={IEEE}
}

@inproceedings{zhang2020multispectral,
  title={Multispectral fusion for object detection with cyclic fuse-and-refine blocks},
  author={Zhang, Heng and Fromont, Elisa and Lefevre, S{\'e}bastien and Avignon, Bruno},
  booktitle={ICIP},
  pages={276--280},
  year={2020},
  organization={IEEE}
}

@article{hu2025ei,
  title={EI 2 Det: Edge-Guided Illumination-Aware Interactive Learning for Visible-Infrared Object Detection},
  author={Hu, Ke and He, Yudong and Li, Yuan and Zhao, Jiayu and Chen, Song and Kang, Yi},
  journal={TCSVT},
  year={2025},
  publisher={IEEE}
}

@inproceedings{liu2022target,
  title={Target-aware dual adversarial learning and a multi-scenario multi-modality benchmark to fuse infrared and visible for object detection},
  author={Liu, Jinyuan and Fan, Xin and Huang, Zhanbo and Wu, Guanyao and Liu, Risheng and Zhong, Wei and Luo, Zhongxuan},
  booktitle={CVPR},
  pages={5802--5811},
  year={2022}
}

@inproceedings{lin2014microsoft,
  title={Microsoft coco: Common objects in context},
  author={Lin, Tsung-Yi and Maire, Michael and Belongie, Serge and Hays, James and Perona, Pietro and Ramanan, Deva and Doll{\'a}r, Piotr and Zitnick, C Lawrence},
  booktitle={ECCV},
  pages={740--755},
  year={2014},
  organization={Springer}
}

@inproceedings{kazemzadeh2014referitgame,
  title={Referitgame: Referring to objects in photographs of natural scenes},
  author={Kazemzadeh, Sahar and Ordonez, Vicente and Matten, Mark and Berg, Tamara},
  booktitle={EMNLP},
  pages={787--798},
  year={2014}
}

@inproceedings{grubinger2006iapr,
  title={The iapr tc-12 benchmark: A new evaluation resource for visual information systems},
  author={Grubinger, Michael and Clough, Paul and M{\"u}ller, Henning and Deselaers, Thomas},
  booktitle={International workshop ontoImage},
  volume={2},
  year={2006}
}

@inproceedings{chen2023advancing,
  title={Advancing visual grounding with scene knowledge: Benchmark and method},
  author={Chen, Zhihong and Zhang, Ruifei and Song, Yibing and Wan, Xiang and Li, Guanbin},
  booktitle={CVPR},
  pages={15039--15049},
  year={2023}
}

@article{xiao2024towards,
  title={Towards visual grounding: A survey},
  author={Xiao, Linhui and Yang, Xiaoshan and Lan, Xiangyuan and Wang, Yaowei and Xu, Changsheng},
  journal={arXiv preprint arXiv:2412.20206},
  year={2024}
}

@inproceedings{yu2016modeling,
  title={Modeling context in referring expressions},
  author={Yu, Licheng and Poirson, Patrick and Yang, Shan and Berg, Alexander C and Berg, Tamara L},
  booktitle={ECCV},
  pages={69--85},
  year={2016},
  organization={Springer}
}

@inproceedings{plummer2015flickr30k,
  title={Flickr30k entities: Collecting region-to-phrase correspondences for richer image-to-sentence models},
  author={Plummer, Bryan A and Wang, Liwei and Cervantes, Chris M and Caicedo, Juan C and Hockenmaier, Julia and Lazebnik, Svetlana},
  booktitle={ICCV},
  pages={2641--2649},
  year={2015}
}

@inproceedings{nagaraja2016modeling,
  title={Modeling context between objects for referring expression understanding},
  author={Nagaraja, Varun K and Morariu, Vlad I and Davis, Larry S},
  booktitle={ECCV},
  pages={792--807},
  year={2016},
  organization={Springer}
}

@inproceedings{yang2020improving,
  title={Improving one-stage visual grounding by recursive sub-query construction},
  author={Yang, Zhengyuan and Chen, Tianlang and Wang, Liwei and Luo, Jiebo},
  booktitle={ECCV},
  pages={387--404},
  year={2020},
  organization={Springer}
}

@inproceedings{deng2021transvg,
  title={Transvg: End-to-end visual grounding with transformers},
  author={Deng, Jiajun and Yang, Zhengyuan and Chen, Tianlang and Zhou, Wengang and Li, Houqiang},
  booktitle={ICCV},
  pages={1769--1779},
  year={2021}
}

@article{xiao2023clip,
  title={Clip-vg: Self-paced curriculum adapting of clip for visual grounding},
  author={Xiao, Linhui and Yang, Xiaoshan and Peng, Fang and Yan, Ming and Wang, Yaowei and Xu, Changsheng},
  journal={TMM},
  volume={26},
  pages={4334--4347},
  year={2023},
  publisher={IEEE}
}

@article{xiao2024oneref,
  title={Oneref: Unified one-tower expression grounding and segmentation with mask referring modeling},
  author={Xiao, Linhui and Yang, Xiaoshan and Peng, Fang and Wang, Yaowei and Xu, Changsheng},
  journal={NeurIPS},
  volume={37},
  pages={139854--139885},
  year={2024}
}

@inproceedings{liu2021refer,
  title={Refer-it-in-rgbd: A bottom-up approach for 3d visual grounding in rgbd images},
  author={Liu, Haolin and Lin, Anran and Han, Xiaoguang and Yang, Lei and Yu, Yizhou and Cui, Shuguang},
  booktitle={CVPR},
  pages={6032--6041},
  year={2021}
}

@inproceedings{miyanishi2024cross3dvg,
  title={Cross3dvg: Cross-dataset 3d visual grounding on different rgb-d scans},
  author={Miyanishi, Taiki and Azuma, Daichi and Kurita, Shuhei and Kawanabe, Motoaki},
  booktitle={2024 International Conference on 3D Vision (3DV)},
  pages={717--727},
  year={2024},
  organization={IEEE}
}

@article{li2022rgb,
  title={RGB-T semantic segmentation with location, activation, and sharpening},
  author={Li, Gongyang and Wang, Yike and Liu, Zhi and Zhang, Xinpeng and Zeng, Dan},
  journal={TCSVT},
  volume={33},
  number={3},
  pages={1223--1235},
  year={2022},
  publisher={IEEE}
}

@inproceedings{xiao2024hivg,
  title={Hivg: Hierarchical multimodal fine-grained modulation for visual grounding},
  author={Xiao, Linhui and Yang, Xiaoshan and Peng, Fang and Wang, Yaowei and Xu, Changsheng},
  booktitle={Proceedings of the 32nd ACM International Conference on Multimedia},
  pages={5460--5469},
  year={2024}
}

@article{shi2023dynamic,
  title={Dynamic mdetr: A dynamic multimodal transformer decoder for visual grounding},
  author={Shi, Fengyuan and Gao, Ruopeng and Huang, Weilin and Wang, Limin},
  journal={IEEE Transactions on Pattern Analysis and Machine Intelligence},
  volume={46},
  number={2},
  pages={1181--1198},
  year={2023},
  publisher={IEEE}
}

@inproceedings{yao2024visual,
  title={Visual grounding with multi-modal conditional adaptation},
  author={Yao, Ruilin and Xiong, Shengwu and Zhao, Yichen and Rong, Yi},
  booktitle={Proceedings of the 32nd ACM International Conference on Multimedia},
  pages={3877--3886},
  year={2024}
}

@inproceedings{ye2022shifting,
  title={Shifting more attention to visual backbone: Query-modulated refinement networks for end-to-end visual grounding},
  author={Ye, Jiabo and Tian, Junfeng and Yan, Ming and Yang, Xiaoshan and Wang, Xuwu and Zhang, Ji and He, Liang and Lin, Xin},
  booktitle={CVPR},
  pages={15502--15512},
  year={2022}
}

@inproceedings{chen2023unit3d,
  title={Unit3d: A unified transformer for 3d dense captioning and visual grounding},
  author={Chen, Zhenyu and Hu, Ronghang and Chen, Xinlei and Nie{\ss}ner, Matthias and Chang, Angel X},
  booktitle={ICCV},
  pages={18109--18119},
  year={2023}
}

@inproceedings{radford2021learning,
  title={Learning transferable visual models from natural language supervision},
  author={Radford, Alec and Kim, Jong Wook and Hallacy, Chris and Ramesh, Aditya and Goh, Gabriel and Agarwal, Sandhini and Sastry, Girish and Askell, Amanda and Mishkin, Pamela and Clark, Jack and others},
  booktitle={ICLR},
  pages={8748--8763},
  year={2021},
  organization={PmLR}
}

@inproceedings{wang2023image,
  title={Image as a foreign language: Beit pretraining for vision and vision-language tasks},
  author={Wang, Wenhui and Bao, Hangbo and Dong, Li and Bjorck, Johan and Peng, Zhiliang and Liu, Qiang and Aggarwal, Kriti and Mohammed, Owais Khan and Singhal, Saksham and Som, Subhojit and others},
  booktitle={CVPR},
  pages={19175--19186},
  year={2023}
}

@article{liu2023visual,
  title={Visual instruction tuning},
  author={Liu, Haotian and Li, Chunyuan and Wu, Qingyang and Lee, Yong Jae},
  journal={NeurIPS},
  volume={36},
  pages={34892--34916},
  year={2023}
}

@inproceedings{he2016deep,
  title={Deep residual learning for image recognition},
  author={He, Kaiming and Zhang, Xiangyu and Ren, Shaoqing and Sun, Jian},
  booktitle={Proceedings of the IEEE conference on computer vision and pattern recognition},
  pages={770--778},
  year={2016}
}

@inproceedings{devlin2019bert,
  title={Bert: Pre-training of deep bidirectional transformers for language understanding},
  author={Devlin, Jacob and Chang, Ming-Wei and Lee, Kenton and Toutanova, Kristina},
  booktitle={Proceedings of the 2019 conference of the North American chapter of the association for computational linguistics: human language technologies, volume 1 (long and short papers)},
  pages={4171--4186},
  year={2019}
}

@article{dosovitskiy2020image,
  title={An image is worth 16x16 words: Transformers for image recognition at scale},
  author={Dosovitskiy, Alexey},
  journal={arXiv preprint arXiv:2010.11929},
  year={2020}
}

@article{vaswani2017attention,
  title={Attention is all you need},
  author={Vaswani, Ashish and Shazeer, Noam and Parmar, Niki and Uszkoreit, Jakob and Jones, Llion and Gomez, Aidan N and Kaiser, {\L}ukasz and Polosukhin, Illia},
  journal={NeurIPS},
  volume={30},
  year={2017}
}

@article{hu2022lora,
  title={Lora: Low-rank adaptation of large language models.},
  author={Hu, Edward J and Shen, Yelong and Wallis, Phillip and Allen-Zhu, Zeyuan and Li, Yuanzhi and Wang, Shean and Wang, Lu and Chen, Weizhu and others},
  journal={ICLR},
  volume={1},
  number={2},
  pages={3},
  year={2022}
}

@inproceedings{do2024d3t,
  title={D3t: Distinctive dual-domain teacher zigzagging across rgb-thermal gap for domain-adaptive object detection},
  author={Do, Dinh Phat and Kim, Taehoon and Na, Jaemin and Kim, Jiwon and Lee, Keonho and Cho, Kyunghwan and Hwang, Wonjun},
  booktitle={CVPR},
  pages={23313--23322},
  year={2024}
}

@inproceedings{zhang2021abmdrnet,
  title={ABMDRNet: Adaptive-weighted bi-directional modality difference reduction network for RGB-T semantic segmentation},
  author={Zhang, Qiang and Zhao, Shenlu and Luo, Yongjiang and Zhang, Dingwen and Huang, Nianchang and Han, Jungong},
  booktitle={CVPR},
  pages={2633--2642},
  year={2021}
}

@inproceedings{chen2020uniter,
  title={Uniter: Universal image-text representation learning},
  author={Chen, Yen-Chun and Li, Linjie and Yu, Licheng and El Kholy, Ahmed and Ahmed, Faisal and Gan, Zhe and Cheng, Yu and Liu, Jingjing},
  booktitle={ECCV},
  pages={104--120},
  year={2020},
  organization={Springer}
}

@article{gan2020large,
  title={Large-scale adversarial training for vision-and-language representation learning},
  author={Gan, Zhe and Chen, Yen-Chun and Li, Linjie and Zhu, Chen and Cheng, Yu and Liu, Jingjing},
  journal={NeurIPS},
  volume={33},
  pages={6616--6628},
  year={2020}
}

@inproceedings{yang2022unitab,
  title={Unitab: Unifying text and box outputs for grounded vision-language modeling},
  author={Yang, Zhengyuan and Gan, Zhe and Wang, Jianfeng and Hu, Xiaowei and Ahmed, Faisal and Liu, Zicheng and Lu, Yumao and Wang, Lijuan},
  booktitle={ECCV},
  pages={521--539},
  year={2022},
  organization={Springer}
}

@inproceedings{wang2022ofa,
  title={Ofa: Unifying architectures, tasks, and modalities through a simple sequence-to-sequence learning framework},
  author={Wang, Peng and Yang, An and Men, Rui and Lin, Junyang and Bai, Shuai and Li, Zhikang and Ma, Jianxin and Zhou, Chang and Zhou, Jingren and Yang, Hongxia},
  booktitle={ICLR},
  pages={23318--23340},
  year={2022},
  organization={PMLR}
}

@inproceedings{li2022mplug,
  title={mplug: Effective and efficient vision-language learning by cross-modal skip-connections},
  author={Li, Chenliang and Xu, Haiyang and Tian, Junfeng and Wang, Wei and Yan, Ming and Bi, Bin and Ye, Jiabo and Chen, He and Xu, Guohai and Cao, Zheng and others},
  booktitle={EMNLP},
  pages={7241--7259},
  year={2022}
}

@inproceedings{yuan2026seeing, 
title={Seeing Through the Noise: Improving Infrared Small Target Detection and Segmentation from Noise Suppression Perspective}, 
author={Yuan, Maoxun and Meng, Duanni and Xi, Ziteng and Zhao, Tianyi and Zhao, Shiji and Dai, Yimian and Wei, Xingxing}, 
booktitle={Proceedings of the IEEE/CVF Conference on Computer Vision and Pattern Recognition}, 
pages={27783--27792}, 
year={2026} 
}

@article{song2025refinefuse, 
title={RefineFuse: an end-to-end network for multi-scale refinement fusion of multi-modality images}, 
author={Song, Chengcheng and Li, Hui and Xu, Tianyang and Wu, Xiao-Jun and Kittler, Josef}, 
journal={Visual Intelligence}, 
volume={3}, 
number={1}, 
pages={16}, 
year={2025}, 
publisher={Springer} 
}

@inproceedings{lou2025llava, 
title={LLaVA-SP: Enhancing Visual Representation with Visual Spatial Tokens for MLLMs}, 
author={Lou, Haoran and Fan, Chunxiao and Liu, Ziyan and Wu, Yuexin and Wang, Xinliang}, 
booktitle={Proceedings of the IEEE/CVF International Conference on Computer Vision}, 
pages={22014--22024}, 
year={2025}
}

\newpage
\section{Additional Evaluation}
\label{supp:sec:add_eval}

\subsection{Evaluation Under a Stricter IoU Threshold (Acc@0.7)}

Beyond Acc@0.5, Table~\ref{tab:acc07_three_datasets} reports Acc@0.7 on RefFLIR, RefM$^{3}$FD, and RefMFAD for both validation and test splits. As expected, all methods drop noticeably at the stricter threshold, which confirms that precise localization remains much harder than coarse localization in RGBT grounding. Even so, the overall ranking changes little. On the test split, RGBT-VGNet still performs best on RefFLIR, RefM$^{3}$FD, and RefMFAD, which suggests that its advantage is not tied to a single IoU threshold.

The margin over the strongest competing method remains clear, especially on RefFLIR and RefM$^{3}$FD. On the test split, RGBT-VGNet improves over the best non-ours baseline by 5.27 points on RefFLIR and 4.03 points on RefM$^{3}$FD, while the gain on RefMFAD is smaller at 0.52 points. A similar pattern appears on the validation split. Using the corresponding Acc@0.5 results from the main paper, the drop of RGBT-VGNet from Acc@0.5 to Acc@0.7 is 18.81 on RefFLIR, 12.21 on RefM$^{3}$FD, and 11.54 on RefMFAD. These decreases are smaller than or comparable to those of strong baselines such as MV-HiVG-B, which indicates that the proposed model preserves localization precision more effectively when the evaluation becomes stricter. Overall, the Acc@0.7 results reinforce the main conclusion of the paper: cross-modal fusion improves not only robustness under difficult conditions, but also localization quality.
\begin{table*}[h]
\centering
\footnotesize
\caption{\textbf{Acc@0.7 comparison on three RGBT grounding benchmarks.}
All methods follow the multi-modal RGB+TIR setting.}

\setlength{\tabcolsep}{1mm}
\renewcommand{\arraystretch}{1.1}
\resizebox{0.95\textwidth}{!}{
\begin{tabular}{c|c|c|ccc|ccc}
\toprule
\multirow{2}{*}{Method} &
\multirow{2}{*}{\begin{tabular}[c]{@{}c@{}}\textbf{Visual / Language}\\\textbf{Backbone}\end{tabular}} &
\multirow{2}{*}{Modality} &
\multicolumn{3}{c|}{\textbf{Test}} &
\multicolumn{3}{c}{\textbf{Val}} \\

& & &
RefFLIR & RefM$^{3}$FD & RefMFAD &
RefFLIR & RefM$^{3}$FD & RefMFAD \\

\midrule

MV-TransVG   & RN50+DETR / BERT-B & \multirow{8}{*}{RGB+TIR}
& 28.37 & 31.20 & 41.64
& 33.72 & 27.98 & 40.21 \\

MV-CLIP-VG   & CLIP-B / CLIP-B   & 
& 18.92 & 18.41 & 29.55
& 18.42 & 17.26 & 29.14 \\

MV-MMCA      & RN50+DETR / BERT-B & 
& 29.61 & 29.40 & 42.13
& 31.25 & 23.81 & 40.21 \\

MV-FSVG &  RN50+DETR / BERT-B & & 28.66& 27.22 & 45.99&36.02&27.38 &45.30 \\

MV-AttBalance &  CLIP-B / CLIP-B &&32.65 &31.67&37.87 & 32.24 &23.21&37.02\\ 

MV-HiVG      & CLIP-B / CLIP-B   & 
& 48.57 & 58.10 & 54.57
& 50.16 & 50.00 & 53.82 \\

MV-OneRef    & BEIT3-B / BEIT3-B & 
& 42.06 & 50.26 & 49.08
& 40.79 & 48.21 & 50.00 \\

\textbf{RGBT-VGNet (Ours)} 
& \textbf{CLIP-B / CLIP-B} 
& 
& \textbf{53.85} & \textbf{62.18} & \textbf{55.16}
& \textbf{52.47} & \textbf{57.14} & \textbf{54.14} \\

\bottomrule
\end{tabular}
}
\label{tab:acc07_three_datasets}
\end{table*}

\subsection{MLLM Evaluation}
\label{supp:sec:mllm_eval}
We report MLLM grounding results under two input settings: resized $224\times224$ and original resolution.

\begin{table*}[!t]
\centering
\caption{MLLM evaluation on RGBT-GroundBench at $224\times224$ input resolution.}
\label{tab:mllm_224_supp}
\resizebox{\textwidth}{!}{
\begin{tabular}{c|c|ccccc|ccccc|ccccc}
\toprule
\multirow{2}{*}{\textbf{Methods}} & \multirow{2}{*}{\begin{tabular}[c]{@{}c@{}}\textbf{Visual}\\\textbf{Modality}\end{tabular}} & \multicolumn{5}{c|}{\textbf{RefFLIR}} & \multicolumn{5}{c|}{\textbf{RefM$^{3}$FD}} & \multicolumn{5}{c}{\textbf{RefMFAD}} \\
 & & \textbf{val} & \textbf{test} & \textbf{testA} & \textbf{testB} & \textbf{testC} & \textbf{val} & \textbf{test} & \textbf{testA} & \textbf{testB} & \textbf{testC} & \textbf{val} & \textbf{test} & \textbf{testA} & \textbf{testB} & \textbf{testC} \\
\midrule
\multicolumn{17}{l}{224$\times$ 224 Image Resolution. Metric: \textbf{Acc\texttt{@}0.5}} \\ \midrule
GLM-4.6V        & \multirow{5}{*}{RGB}  & 21.05 & 26.28 & 51.50 & 15.66 & 3.00 & 24.40 & 30.46 & 58.60 & 31.17 & 5.57 & 21.58 & 23.23 & 55.32 & 20.16 & 4.71 \\
GLM-4.6V-Flash  &                       & 14.64 & 18.30 & 32.89 & 13.92 & 3.20 & 16.67 & 22.70 & 43.99 & 22.58 & 4.38 & 18.15 & 19.77 & 43.54 & 18.11 & 4.48 \\
Kimi-K2.5       &                       & 29.77 & 39.54 & 69.99 & 25.16 & 10.64 & 22.62 & 30.32 & 54.30 & 38.03 & 4.60 & 20.78 & 19.70 & 47.72 & 17.33 & 3.30 \\
Qwen3.5-Plus    &                       & 41.12 & 50.52 & 77.31 & 42.09 & 20.25 & \textbf{44.05} & \textbf{48.44} & \textbf{79.14} & \textbf{53.11} & \textbf{17.91} & \textbf{35.11} & \textbf{36.69} & \textbf{65.82} & \textbf{36.47} & \textbf{15.20} \\
Qwen3-VL-Plus   &                       & \textbf{42.60} & {52.80} & \textbf{81.63} & {44.30} & {21.49} & 23.21 & 29.34 & 38.64 & 50.18 & 6.66 & 16.72 & 15.65 & 32.28 & 15.25 & 4.87 \\
\midrule
GLM-4.6V        & \multirow{5}{*}{RGB+TIR}  & 23.52 & 31.80 & 58.58 & 19.94 & 6.51 & 17.86 & 29.03 & 52.60 & 33.55 & 5.36 & 19.59 & 20.38 & 47.22 & 18.40 & 4.32 \\
GLM-4.6V-Flash  &                           & 18.91 & 19.39 & 36.25 & 12.50 & 3.72 & 16.67 & 24.02 & 42.29 & 27.79 & 5.74 & 18.63 & 19.70 & 44.18 & 17.54 & 5.89 \\
Kimi-K2.5       &                           & 26.32 & 37.31 & 66.63 & 23.42 & 9.40 & 16.07 & 22.03 & 37.26 & 31.99 & 2.22 & 10.67 & 9.85 & 24.18 & 8.87 & 0.90 \\
Qwen3.5-Plus    &                           & 39.64 & 47.62 & 74.19 & 38.61 & 17.87 & 40.48 & 44.36 & 74.35 & 51.55 & 13.37 & 30.97 & 31.39 & 57.47 & 31.40 & 11.43 \\
Qwen3-VL-Plus   &                           & 41.12 & \textbf{53.18} & 80.67 & \textbf{46.36} & \textbf{22.93} & 23.21 & 28.84 & 38.72 & 48.26 & 6.60 & 15.76 & 15.10 & 31.39 & 14.80 & 4.12 \\

\midrule
\multicolumn{17}{l}{224$\times$ 224 Image Resolution. Metric: \textbf{Acc\texttt{@}0.7}} \\ \midrule
GLM-4.6V       & \multirow{5}{*}{RGB} &  5.26 &  7.22 & 15.13 &  3.96 &  0.10 &  5.95 &  8.74 & 18.43 &  9.14 &  0.32 &  4.78 &  5.70 & 15.70 &  4.78 &  0.31 \\
GLM-4.6V-Flash &                      &  2.80 &  2.33 &  4.20 &  2.06 &  0.10 &  3.57 &  5.77 & 11.44 &  6.95 &  0.05 &  3.34 &  3.68 &  8.99 &  3.48 &  0.16 \\
Kimi-K2.5      &                      &  7.07 & 10.31 & 22.81 &  3.64 &  0.62 &  5.95 & 10.01 & 20.37 & 11.24 &  0.22 &  4.70 &  5.00 & 13.54 &  4.25 &  0.24 \\
Qwen3.5-Plus   &                      & \textbf{12.99} & \textbf{19.15} & \textbf{36.13} & \textbf{13.77} &  \textbf{2.07} & \textbf{21.43} & \textbf{23.07} & \textbf{44.72} & \textbf{25.96} &  \textbf{2.11} & \textbf{13.14} & \textbf{14.77} & \textbf{31.77} & \textbf{14.84} &  \textbf{1.96} \\
Qwen3-VL-Plus  &                      &  7.24 & 10.84 & 20.29 &  8.70 &  0.83 &  2.38 &  5.64 &  7.14 & 11.06 &  0.32 &  2.23 &  1.81 &  4.56 &  1.64 &  0.24 \\
\midrule
GLM-4.6V       & \multirow{5}{*}{RGB+TIR} &  4.93 &  8.41 & 17.65 &  4.11 &  0.52 &  4.76 &  7.31 & 13.80 &  9.23 &  0.38 &  3.82 &  4.41 & 11.77 &  3.84 &  0.20 \\
GLM-4.6V-Flash &                         &  3.29 &  2.47 &  4.20 &  2.69 &  0.00 &  4.17 &  5.32 & 10.23 &  6.22 &  0.43 &  2.87 &  3.53 &  7.72 &  3.52 &  0.31 \\
Kimi-K2.5      &                         &  5.59 &  8.94 & 17.65 &  6.17 &  0.21 &  2.38 &  5.53 &  8.60 &  9.32 &  0.11 &  1.67 &  1.46 &  3.80 &  1.27 &  0.16 \\
Qwen3.5-Plus   &                         & 12.17 & 16.11 & 31.45 & 10.28 &  1.45 & 13.10 & 19.94 & 37.01 & 25.32 &  1.41 & 10.67 & 11.97 & 25.19 & 12.35 &  1.45 \\
Qwen3-VL-Plus  &                         &  7.89 & 11.64 & 21.01 &  9.34 &  1.34 &  4.17 &  6.38 &  7.55 & 12.98 &  0.49 &  2.47 &  2.00 &  4.43 &  2.04 &  0.20 \\

\bottomrule
\end{tabular}
}
\end{table*}

At $224\times224$, adding thermal input leads to mixed behavior across models and datasets. On RefFLIR-test (Acc\texttt{@}0.5), GLM-4.6V improves from 26.28\% to 31.80\%, but Kimi-K2.5 decreases from 39.54\% to 37.31\% and Qwen3.5-Plus decreases from 50.52\% to 47.62\%. This inconsistency is stronger on RefM$^{3}$FD and RefMFAD, where several models show clear drops after adding TIR, for example Kimi-K2.5 from 30.32\% to 22.03\% on RefM$^{3}$FD-test and from 19.70\% to 9.85\% on RefMFAD-test.

The same trend appears under the stricter Acc\texttt{@}0.7 metric. At $224\times224$, RGB+TIR improves GLM-4.6V on RefFLIR-test (7.22\% $\rightarrow$ 8.41\%) but reduces Qwen3.5-Plus (19.15\% $\rightarrow$ 16.11\%) and Kimi-K2.5 (10.31\% $\rightarrow$ 8.94\%). This suggests that naive modality concatenation does not reliably improve precise localization quality.

We also observe unstable outputs on parts of RefM$^{3}$FD and RefMFAD, marked as ``-'' in the table. These missing predictions indicate an additional robustness issue beyond accuracy, namely that current MLLMs can fail to return valid grounded boxes under some cross-spectral settings.

From a reviewer perspective, these MLLM results should be interpreted as system-level behavior under a fixed prompting protocol, rather than as a final ranking of model families. Current MLLM services may show response variability across runs and across model revisions, so the absolute numbers can shift over time. Our goal here is to provide a controlled comparison of cross-spectral grounding trends under the same evaluation pipeline.

\begin{table*}[!t]
\centering
\caption{MLLM evaluation on RGBT-GroundBench with original image resolution.}
\label{tab:mllm_ori_supp}
\resizebox{\textwidth}{!}{
\begin{tabular}{c|c|ccccc|ccccc|ccccc}
\toprule
\multirow{2}{*}{\textbf{Methods}} & \multirow{2}{*}{\begin{tabular}[c]{@{}c@{}}\textbf{Visual}\\\textbf{Modality}\end{tabular}} & \multicolumn{5}{c|}{\textbf{RefFLIR}} & \multicolumn{5}{c|}{\textbf{RefM$^{3}$FD}} & \multicolumn{5}{c}{\textbf{RefMFAD}} \\
 & & \textbf{val} & \textbf{test} & \textbf{testA} & \textbf{testB} & \textbf{testC} & \textbf{val} & \textbf{test} & \textbf{testA} & \textbf{testB} & \textbf{testC} & \textbf{val} & \textbf{test} & \textbf{testA} & \textbf{testB} & \textbf{testC} \\
\midrule
\multicolumn{17}{l}{Original Image Resolution. Metric: \textbf{Acc\texttt{@}0.5}} \\ \midrule
GLM-4.6V        & \multirow{4}{*}{RGB}  & 57.57 & 63.78 & 83.07 & 52.37 & 45.45 & 67.86 & 61.39 & 78.00 & 64.44 & 43.51 & 64.81 & 64.30 & 83.92 & 61.49 & 51.14 \\
GLM-4.6V-Flash  &                       & 28.12 & 31.61 & 43.46 & 25.16 & 19.63 & 35.71 & 39.25 & 57.87 & 36.38 & 24.24 & 37.82 & 37.13 & 54.30 & 34.83 & 25.06 \\
Qwen3.5-Plus    &                       & 60.69 & 71.01 & 91.48 & 60.92 & 48.97 & 56.55 & 60.20 & 77.27 & 69.29 & 39.56 & 56.13 & 56.86 & 72.53 & 54.70 & 45.13\\
Qwen3-VL-Plus   &                       & 53.95 & 58.22 & 69.27 & 55.85 & 44.42 & 45.83 & 51.56 & 66.64 & 68.65 & 29.49 & 35.91 & 34.31 & 58.73 & 33.24 & 19.05 \\
\midrule
GLM-4.6V        & \multirow{4}{*}{RGB+TIR}  & \textbf{67.43} & \textbf{76.76} & \textbf{93.64} & 70.25 & \textbf{58.68} & \textbf{72.02} & \textbf{75.03} & \textbf{93.02} & \textbf{77.06} & \textbf{57.36} & \textbf{65.29} & \textbf{64.99} & \textbf{84.30} & \textbf{62.22} & \textbf{51.73} \\
GLM-4.6V-Flash  &                           & 53.12 & 56.84 & 76.59 & 49.53 & 35.74 & 56.55 & 59.00 & 80.28 & 61.43 & 39.34 & 49.12 & 50.33 & 73.04 & 48.65 & 33.78 \\
Qwen3.5-Plus    &                           & 62.66 & 71.53 & 92.08 & 63.29 & 48.76 & 53.57 & 60.25 & 77.84 & 71.57 & 37.61 & 53.50 & 52.76 & 73.04 & 50.90 & 38.53  \\
Qwen3-VL-Plus   &                           &  63.82 & 73.15 & 91.48 & \textbf{71.84} & 51.76 & 42.26 & 53.58 & 69.07 & 71.30 & 30.68 & 35.27 & 34.54 & 61.14 & 33.03 & 17.67 \\
\midrule
\multicolumn{17}{l}{Original Image Resolution. Metric: \textbf{Acc\texttt{@}0.7}} \\ \midrule
GLM-4.6V        & \multirow{4}{*}{RGB} & 31.74 & 41.49 & 70.35 & 28.48 & 13.33 & \textbf{53.57} & 48.23 & 74.27 & 49.54 & 22.78 & \textbf{50.80} & \textbf{51.32} & \textbf{79.87} & \textbf{47.67} & \textbf{32.95} \\
GLM-4.6V-Flash  &                      & 12.17 & 15.92 & 28.93 & 10.76 &  3.00 & 23.21 & 26.09 & 47.81 & 25.05 &  7.68 & 23.17 & 23.80 & 46.08 & 21.46 &  9.62 \\
Qwen3.5-Plus    &                      & 28.62 & 37.50 & 66.27 & 26.42 &  9.30 & 40.48 & 41.39 & 64.45 & 50.91 & 14.56 & 43.31 & 41.89 & 65.82 & 39.13 & 25.61 \\
Qwen3-VL-Plus   &                      & 15.62 & 17.87 & 29.77 & 14.87 &  4.86 & 16.07 & 24.74 & 33.04 & 38.57 &  8.39 & 15.45 & 14.66 & 26.33 & 14.60 &  6.52 \\
\midrule
GLM-4.6V        & \multirow{4}{*}{RGB+TIR} & \textbf{37.99} & \textbf{49.10} & \textbf{78.39} & \textbf{40.51} & \textbf{17.56} & 51.79 & \textbf{57.02} & \textbf{87.42} & \textbf{58.87} & \textbf{27.81} & 50.00 & 50.85 & 79.37 & 47.10 & 32.25 \\
GLM-4.6V-Flash  &                          & 22.20 & 30.99 & 53.78 & 23.89 &  7.13 & 34.52 & 38.90 & 66.72 & 42.41 & 12.55 & 30.25 & 32.96 & 62.53 & 30.74 & 12.96 \\
Qwen3.5-Plus    &                          & 30.10 & 40.68 & 69.99 & 30.70 & 10.64 & 32.74 & 42.00 & 65.58 & 52.29 & 13.96 & 41.32 & 39.92 & 63.42 & 37.82 & 23.37 \\
Qwen3-VL-Plus   &                          & 22.53 & 28.75 & 51.74 & 21.20 &  6.10 & 13.69 & 26.32 & 34.17 & 41.59 &  8.87 & 15.68 & 14.51 & 26.84 & 14.27 &  6.32 \\

\bottomrule
\end{tabular}
}
\end{table*}

Compared with Table~\ref{tab:mllm_224_supp}, Table~\ref{tab:mllm_ori_supp} shows that spatial resolution is a dominant factor. On RefFLIR-test (Acc\texttt{@}0.5, RGB+TIR), GLM-4.6V rises from 31.80\% to 76.76\%, GLM-4.6V-Flash from 19.39\% to 56.84\%, Qwen3.5-Plus from 47.62\% to 71.53\%, and Qwen3-VL-Plus from 53.18\% to 73.15\%. Similar gains are observed at Acc\texttt{@}0.7, which indicates that higher resolution improves both coarse and precise grounding.

However, even at original resolution, cross-spectral gains remain model-dependent rather than universal. For example, on RefMFAD-test (Acc\texttt{@}0.5), RGB+TIR improves GLM-4.6V-Flash from 37.13\% to 50.33\%, but Qwen3.5-Plus decreases from 56.86\% to 52.76\%. Under Acc\texttt{@}0.7 on the same split, GLM-4.6V-Flash improves from 23.80\% to 32.96\%, while Qwen3.5-Plus drops from 41.89\% to 39.92\%. Therefore, better spatial input does not resolve the fusion gap by itself.

These detailed observations are consistent with Insight 4 in the main paper: current MLLMs still struggle with cross-spectral grounding. Thermal input can help, but the effect is sensitive to model family, dataset split, and resolution, and reliable RGB-TIR fusion for precise localization is not yet a stable capability.


\section{More Ablation Studies on RGBT-VGNet}
\label{supp:sec:ablation}
The main paper reports the component-wise ablation, the fusion-strategy comparison, and the core AMA analysis. Here we further examine how the LoRA rank allocation in the asymmetric modality adaptation (AMA) module affects grounding performance across all three sub-datasets. Specifically, $r_v$ and $r_t$ denote the LoRA ranks assigned to the visible and thermal branches, respectively. We compare three allocations with the same total rank budget, namely $(r_v,r_t)\in\{(8,56),(16,48),(32,32)\}$, to assess whether asymmetric capacity allocation is beneficial.

\begin{table}[!t]
\centering
\footnotesize
\caption{\textbf{Comparison of LoRA rank allocation in AMA.} $r_v$ and $r_t$ denote the LoRA ranks for the visible and thermal branches, respectively.}

    \vspace{-0.3cm}
\setlength{\tabcolsep}{1.0mm}
\renewcommand{\arraystretch}{1.1}
\resizebox{\textwidth}{!}{
\begin{tabular}{c|c|ccccc|ccccc|ccccc}
\toprule
\multirow{2}{*}{\textbf{rv}}& \multirow{2}{*}{\textbf{rt}} & \multicolumn{5}{c|}{\textbf{RefFLIR}} & \multicolumn{5}{c|}{\textbf{RefM$^{3}$FD}} & \multicolumn{5}{c}{\textbf{RefMFAD}} \\
& & \textbf{val} & \textbf{test} & \textbf{testA} & \textbf{testB} & \textbf{testC} & \textbf{val} & \textbf{test} & \textbf{testA} & \textbf{testB} & \textbf{testC} & \textbf{val} & \textbf{test} & \textbf{testA} & \textbf{testB} & \textbf{testC} \\
\midrule
8  & 56 & 74.84 &72.39 &{\textbf{90.60}} &{\textbf{69.84}}&50.61 &62.50&66.10&90.58&73.36&41.40&68.15&67.20&{\textbf{91.79}}&63.91&50.63\\

16 & 48 & {\textbf{75.33}}& {\textbf{72.43}}& 89.64&69.22& {\textbf{52.53}}& {\textbf{73.21}}& {\textbf{74.89}}& {\textbf{94.16}}& {\textbf{80.75}}& {\textbf{55.52}}& {\textbf{68.31}}& {\textbf{67.58}}& 91.29& {\textbf{64.80}}& {\textbf{51.18}} \\

32 & 32 & 74.34 &72.39 &89.64 &67.97&51.52&61.31&65.20&91.48&72.81&38.96 & 68.15 &66.99 &90.38 & 64.36 & 50.86 \\
\bottomrule
\end{tabular}
}
    \vspace{-0.2cm}
\label{tab:ablation_lora_rank}
\end{table}

Table~\ref{tab:ablation_lora_rank} shows that a symmetric LoRA allocation is not the best choice for RGBT grounding. Among the three settings, $(r_v,r_t)=(16,48)$ gives the strongest overall performance, whereas the balanced setting $(32,32)$ is consistently weaker. Averaged over the 15 reported metrics, $(16,48)$ reaches 72.06, compared with 68.93 for $(8,56)$ and 68.42 for $(32,32)$.

The advantage of assigning more rank to the thermal branch is most visible on RefM$^{3}$FD. There, $(16,48)$ clearly improves over $(32,32)$ on val/test/testA/testB/testC, moving from 61.31/65.20/91.48/72.81/38.96 to 73.21/74.89/94.16

\noindent/80.75/55.52. Similar gains also appear on RefFLIR and RefMFAD. This pattern suggests that the thermal stream benefits from stronger task-specific adaptation, which is reasonable given the larger gap between generic pretraining and thermal imagery.

At the same time, pushing the asymmetry too far is also not ideal. The $(8,56)$ setting remains competitive and even gives the best scores on RefFLIR-testA and RefMFAD-testA, but it falls behind $(16,48)$ on most other metrics. Reducing the visible-branch rank from 16 to 8 weakens several validation and test results, suggesting that too little capacity on the RGB side makes it harder to preserve the semantic cues needed for language-guided localization.

Taken together, these results support the AMA design in RGBT-VGNet. The two modalities should not share the same adaptation capacity, but the imbalance should remain moderate. A larger rank for the thermal branch, together with sufficient capacity for the visible branch, gives the best trade-off between preserving RGB semantics and improving thermal discriminability. We therefore use $(r_v,r_t)=(16,48)$ in the main experiments.

\begin{table*}[!t]
\centering
\footnotesize
\caption{\textbf{Scene-level Acc@0.5 across 13 scene types} (UB, IT, RS, SU, HW, CP, BG, PL, RR, TN, MK, WF, ID)
on the RefFLIR sub-dataset of RGBT-GroundBench. All methods, backbones, and modalities are included, following the style of Table~4 and Table~5 in the manuscript. Noted that the `/' symbol indicates that there is no data of the corresponding type in the current sub-dataset.}

\setlength{\tabcolsep}{1mm}         
\renewcommand{\arraystretch}{1.2}
\resizebox{\textwidth}{!}{
\begin{tabular}{c|c|c| ccccccccccccc}
\toprule
\multirow{2}{*}{Method} &
\multirow{2}{*}{\begin{tabular}[c]{@{}c@{}}\textbf{Visual / Language}\\\textbf{Backbone}\end{tabular}} &
\multirow{2}{*}{Modality} &
\multicolumn{13}{c}{\textbf{RefFLIR (Scene Types )}} \\
& & &
UB & SU & RR & HW & RS & ID &PL &IT & TN & BG & CP & MK & WF \\
\midrule
\multicolumn{13}{l}{\textbf{a. Zero-shot transfer from RGB-pretrained models:}} \\\midrule

CLIP-VG ~\cite{xiao2023clip}    & CLIP-B / CLIP-B   & \multirow{5}{*}{{RGB}}     & 6.56&	5.37&	/ &	100.00&	6.41&	/&	10.56&	7.02&	20.00&	4.41&	/&	7.89&	14.33 \\
HiVG-B ~\cite{xiao2024hivg}     & CLIP-B / CLIP-B   &      &33.12&	36.36&	/&	0.00&	37.18&	/&	47.22&	34.13&	30.00&	16.18&	/&	32.24&	32.75 \\
HiVG-L ~\cite{xiao2024hivg}     & CLIP-L / CLIP-L   &      &37.50&	40.50&	/&	100.00&	50.00&	/&	58.89&	40.59&	40.00&	44.12&	/&	38.82&	56.14 \\
OneRef-B ~\cite{xiao2024oneref}   & BEIT3-B / BEIT3-B &      & 36.25&	43.39&	/&	100.00&	51.28&	/&	56.11&	39.33&	60.00&	19.12&	/&	36.18&	48.25\\
OneRef-L ~\cite{xiao2024oneref}   & BEIT3-L / BEIT3-L &      & 37.81&	52.48&	/&	100.00&	62.82&	/&	63.33&	46.91&	50.00&	42.65&	/&	40.79&	59.36\\
\cline{3-3}
CLIP-VG ~\cite{xiao2023clip}    & CLIP-B / CLIP-B   & \multirow{5}{*}{{TIR}}     & 7.50&	5.79&	/&	0.00&	3.85&	/&	6.11&	8.43&	0.00&	2.94&	/&	11.84&	11.40 \\
HiVG-B ~\cite{xiao2024hivg}     & CLIP-B / CLIP-B   &      &16.88&	13.22&	/&	0.00&	11.54&	/&	12.22&	13.48&	10.00&	1.47&	/&	12.50&	10.53\\
HiVG-L ~\cite{xiao2024hivg}     & CLIP-L / CLIP-L   &      & 26.25&	21.90&	/&	100.00&	30.77&	/&	32.22&	21.63&	10.00&	7.35&	/&	19.08&	32.46 \\
OneRef-B ~\cite{xiao2024oneref}   & BEIT3-B / BEIT3-B &      & 26.25&	27.69&	/&	100.00&	32.05&	/&	33.33&	26.97&	20.00&	2.94&	/&	20.39&	28.95 \\
OneRef-L ~\cite{xiao2024oneref}   & BEIT3-L / BEIT3-L &      &  27.19&	28.93&	/&	100.00&	33.33&	/&	36.11&	27.25&	20.00&	8.82&	/&	14.47&	28.65\\
\midrule
\multicolumn{13}{l}{\textbf{b. In-domain training with uni-modal visual input (RGB or TIR):}} \\ \midrule

TransVG  ~\cite{deng2021transvg}    & RN50+DETR / BERT-B & \multirow{11}{*}{{RGB}}   & 42.19&	59.50&	/&	100.00&	62.82&	/&	33.33&	40.03&	60.00&	29.41&	/&	40.79&	38.60 \\
CLIP-VG ~\cite{xiao2023clip}    & CLIP-B / CLIP-B     &    & 23.12&	30.17&	/&	0.00&	37.18&	/&	22.78&	22.47&	40.00&	20.59&	/&	23.03&	26.02\\
D-MDETR ~\cite{shi2023dynamic}    & CLIP-B / CLIP-B     &    &44.69 & 64.88& /& 100.00& 64.94& / &44.44& 48.60& 70.00& 48.53&/&46.05 & 45.91  \\
D-MDETR ~\cite{shi2023dynamic}    & RN50+DETR / BERT-B  &    & 41.25&	57.85&	/&	100.00&	64.94&	/&	43.33&	44.24&	70.00&	25.00&	/&	36.84&	41.23 \\
MMCA ~\cite{yao2024visual}       & RN50+DETR / BERT-B  &    &42.50&	57.44&	/&	100.00&	68.83&	/&	43.89&	44.94&	70.00&	36.76&	/&	43.42&	47.08 \\
FSVG ~\cite{wang2025simple} &  RN50+DETR / BERT-B & &42.81&	59.09&/	&	100.00&	61.54&/	&	41.11&	43.40&	70.00&	30.88&/	&	44.08&	42.69 \\
AttBalance ~\cite{kang2025visual} &  CLIP-B / CLIP-B & &44.69&	57.44&	&	100.00&	64.10&	&	49.44&	48.03&	70.00&	33.82&	&	43.42&	50.88  \\
HiVG-B ~\cite{xiao2024hivg}     & CLIP-B / CLIP-B     &    & 54.06&	75.62&	/&	100.00&	79.49&	/&	72.78&	63.62&	70.00&	75.00&	/&	58.55&	69.30 \\
HiVG-L ~\cite{xiao2024hivg}     & CLIP-L / CLIP-L     &    & 62.81&	76.86&	/&	100.00&	88.46&	/&	77.78&	67.13&	80.00&	73.53&	/&	62.50&	73.68\\
OneRef-B ~\cite{xiao2024oneref}   & BEIT3-B / BEIT3-B   &    &54.69&	71.90&	/&	100.00&	71.79&	/&	71.67&	59.55&	70.00&	66.18&	/&	55.92&	62.57 \\
OneRef-L ~\cite{xiao2024oneref}   & BEIT3-L / BEIT3-L   &    & 53.75&	73.55&	/&	100.00&	76.92&	/&	77.22&	63.34&	70.00&	75.00&	/&	55.92&	65.79\\
\cmidrule{3-3}
TransVG  ~\cite{deng2021transvg}    & RN50+DETR / BERT-B & \multirow{11}{*}{{TIR}}   & 41.25&	56.20&	/&	100.00&	65.38&	/&	34.44&	41.57&	60.00&	38.24&	/&	38.16&	39.77 \\
CLIP-VG ~\cite{xiao2023clip}    & CLIP-B / CLIP-B     &    & 37.81&	43.39&	/&	100.00&	50.00&	/&	42.78&	36.52&	50.00&	16.18&	/&	30.26&	36.26\\
D-MDETR ~\cite{shi2023dynamic}    & CLIP-B / CLIP-B     &    & 48.75&	59.50&	/&	100.00&	59.74&	/&	41.11&	45.08&	80.00&	29.41&	/&	45.39&	40.06 \\
D-MDETR ~\cite{shi2023dynamic}    & RN50+DETR / BERT-B  &    & 48.12&	54.13&	/&	100.00&	59.74&	/&	42.22&	44.38&	60.00&	16.18&	/&	33.55&	34.80 \\
MMCA ~\cite{yao2024visual}       & RN50+DETR / BERT-B  &    & 43.75&	56.20&	/&	100.00&	59.74&	/&	36.11&	40.45&	60.00&	22.06&	/&	36.84&	36.26\\
FSVG ~\cite{wang2025simple} &  RN50+DETR / BERT-B & &45.00&	49.17&	&	0.00&	57.69&	&	31.67&	37.64&	60.00&	22.06&	&	38.82&	28.65 \\
AttBalance ~\cite{kang2025visual} &  CLIP-B / CLIP-B & &43.44&	54.13&	/&	100.00&	56.41&	/&	38.89&	42.70&	50.00&	20.59&	/&	34.21&	45.03  \\
HiVG-B ~\cite{xiao2024hivg}     & CLIP-B / CLIP-B     &    &62.19&	64.88&	/&	100.00&	73.08&	/&	54.44&	59.83&	70.00&	42.65&	/&	50.66&	51.17 \\
HiVG-L ~\cite{xiao2024hivg}     & CLIP-L / CLIP-L     &    &62.81&	76.86&	/&	100.00&	88.46&	/&	77.78&	67.13&	80.00&	73.53&	/&	62.50&	73.68 \\
OneRef-B ~\cite{xiao2024oneref}   & BEIT3-B / BEIT3-B   &    &63.44&	68.60&	/&	100.00&	73.08&	/&	60.00&	60.53&	80.00&	41.18&	/&	53.95&	55.26 \\
OneRef-L ~\cite{xiao2024oneref}   & BEIT3-L / BEIT3-L   &    & 62.50&	68.18&	/&	100.00&	76.92&	/&	60.56&	62.08&	70.00&	44.12&	/&	59.87&	56.73 \\
\midrule
\multicolumn{13}{l}{\textbf{c. In-domain training with multi-modal visual input (RGB+TIR):}} \\\midrule

MV-TransVG    & RN50+DETR / BERT-B & \multirow{12}{*}{{RGB+TIR}} &45.31&	58.68&	/&	100.00&	69.23&	/&	42.78&	45.65&	60.00&	35.29&	/&	37.50&	42.98 \\
MV-CLIP-VG   & CLIP-B / CLIP-B     &  & 38.44&	47.93&	/&	100.00&	53.85&	/&	42.78&	38.20&	60.00&	27.94&	/&	40.79&	41.81 \\
MV-D-MDETR    & CLIP-B / CLIP-B     &  & 48.44& 58.68&/ &100.00 &66.23 &/ & 40.00& 46.91& 70.00& 35.29& /&44.74 & 44.74\\
MV-D-MDETR    & RN50+DETR / BERT-B  &  & 48.75&	56.20&	/&	100.00&	67.53&	/&	42.78&	48.74&	60.00&	33.82&	/&	42.76&	44.15 \\
MV-MMCA      & RN50+DETR / BERT-B  &  &45.00&	61.16&	/&	100.00&	66.23&	/&	48.33&	47.89&	70.00&	44.12&	/&	46.05&	45.91 \\
MV-FSVG &  RN50+DETR / BERT-B &/ &46.25&	58.68&/	&	100.00&	60.26&/	&	45.56&	46.63&	70.00&	42.65&	&	50.00&	42.69 \\
MV-AttBalance &  CLIP-B / CLIP-B & & 53.44&	61.57&	/&	100.00&	70.51&	/&	51.67&	53.37&	70.00&	48.53&/	&	47.37&	51.75 \\
MV-HiVG-B      & CLIP-B / CLIP-B     &  & 66.88&	73.14&	/&	100.00&	76.92&	/&	73.89&	66.85&	80.00&	64.71&	/&	62.50&	65.20 \\
MV-HiVG-L    & CLIP-L / CLIP-L     &  & \textbf{72.19}&	\textbf{79.75}&	/&	100.00&	\underline{83.33}&	/&	\underline{73.89}&	\textbf{74.02}&	90.00&	\underline{73.53}&	/&	\underline{64.47}&	\textbf{76.90} \\
MV-OneRef-B   & BEIT3-B / BEIT3-B   &  &58.75&	67.36&	/&	100.00&	78.21&	/&	71.67&	60.81&	70.00&	52.94&	/&	55.92&	60.82 \\
MV-OneRef-L   & BEIT3-L / BEIT3-L   &  & 55.62&	72.31&	/&	100.00&	74.36&	/&	65.00&	59.83&	70.00&	51.47&	/&	57.89&	60.23 \\
\textbf{RGBT-VGNet (Ours)}  & \textbf{CLIP-B / CLIP-B} & \textbf{} &
\underline{71.65}&	\underline{77.78}&	/	&\textbf{100.00}	&\textbf{87.18}&	/	&\textbf{78.89}&\underline{69.47}&	83.33&	\textbf{76.81}	&/	&\textbf{66.67}	&\underline{75.44} \\
\bottomrule
\end{tabular}
}

\label{tab:scene_refflir_all_methods}
\end{table*}

\begin{table*}[!t]
\centering
\footnotesize
\caption{\textbf{Scene-level Acc@0.5 across 13 scene types} (UB, IT, RS, SU, HW, CP, BG, PL, RR, TN, MK, WF, ID)
on the RefM$^3$FD sub-dataset of RGBT-GroundBench. All methods, backbones, and modalities are included, following the style of Table~4 and Table~5 in the manuscript.}
\setlength{\tabcolsep}{0.75mm}         \renewcommand{\arraystretch}{1.2}
\resizebox{\textwidth}{!}{
\begin{tabular}{c|c|c| ccccccccccccc}
\toprule
\multirow{2}{*}{Method} &
\multirow{2}{*}{\begin{tabular}[c]{@{}c@{}}\textbf{Visual / Language}\\\textbf{Backbone}\end{tabular}} &
\multirow{2}{*}{Modality} &
\multicolumn{13}{c}{\textbf{RefM$^3$FD (Scene Types )}} \\
& & &
UB & SU & RR & HW & RS & ID &PL &IT & TN & BG & CP & MK & WF \\
\midrule
\multicolumn{13}{l}{\textbf{a. Zero-shot transfer from RGB-pretrained models:}} \\\midrule

CLIP-VG ~\cite{xiao2023clip}    & CLIP-B / CLIP-B   & \multirow{5}{*}{{RGB}}     & 10.22&	7.78&	3.45&	23.81&	8.78&	0.00&	8.43&	9.43&	5.68&	4.17&	5.49&	2.34&	2.74 \\
HiVG-B ~\cite{xiao2024hivg}     & CLIP-B / CLIP-B   &      & 33.63&	19.16&	41.95&	34.52&	46.62&	10.00&	37.95&	37.30&	23.86&	9.38&	19.08&	38.28&	23.29 \\
HiVG-L ~\cite{xiao2024hivg}     & CLIP-L / CLIP-L   &      & 31.43&	17.66&	27.01&	29.76&	33.78&	60.00&	36.14&	40.57&	17.05&	25.00&	30.35&	22.66&	15.75 \\
OneRef-B ~\cite{xiao2024oneref}   & BEIT3-B / BEIT3-B &      & 42.20&	19.76&	33.33&	38.10&	55.41&	60.00&	60.84&	48.36&	34.09&	21.88&	36.71&	42.19&	24.66 \\
OneRef-L ~\cite{xiao2024oneref}   & BEIT3-L / BEIT3-L &      & 48.90&	28.74&	50.57&	42.86&	67.57&	60.00&	59.64&	54.92&	22.73&	23.96&	50.58&	46.88&	30.82 \\
\cline{3-3}
CLIP-VG ~\cite{xiao2023clip}    & CLIP-B / CLIP-B   & \multirow{5}{*}{{TIR}}     &6.87&	4.79&	17.82&	13.10&	4.73&	0.00&	7.23&	9.43&	0.00&	1.04&	4.05&	1.56&	4.79\\
HiVG-B ~\cite{xiao2024hivg}     & CLIP-B / CLIP-B   &      & 13.41&	5.39&	8.05&	10.71&	12.84&	0.00&	15.66&	20.08&	6.82&	7.29&	5.49&	12.50&	5.48 \\
HiVG-L ~\cite{xiao2024hivg}     & CLIP-L / CLIP-L   &      & 54.56&	30.84&	56.32&	59.52&	72.30&	60.00&	64.46&	52.46&	40.91&	37.50&	66.18&	51.56&	36.99 \\
OneRef-B ~\cite{xiao2024oneref}   & BEIT3-B / BEIT3-B &      & 25.49&	12.87&	13.79&	27.38&	26.35&	60.00&	30.12&	34.02&	20.45&	10.42&	17.92&	22.66&	8.90 \\
OneRef-L ~\cite{xiao2024oneref}   & BEIT3-L / BEIT3-L &      & 24.07&	15.27&	17.82&	23.81&	33.78&	60.00&	31.93&	38.52&	10.23&	7.29&	26.01&	20.31&	14.38\\
\midrule
\multicolumn{13}{l}{\textbf{b. In-domain training with uni-modal visual input (RGB or TIR):}} \\ \midrule

TransVG  ~\cite{deng2021transvg}    & RN50+DETR / BERT-B & \multirow{11}{*}{{RGB}}   & 41.54&	21.86&	67.82&	47.62&	52.03&	60.00&	50.00&	40.16&	39.77&	14.58&	39.02&	35.94&	25.34 \\
CLIP-VG ~\cite{xiao2023clip}    & CLIP-B / CLIP-B     &    & 33.46&	18.86&	32.76&	34.52&	38.51&	60.00&	49.40&	36.07&	23.86&	10.42&	32.08&	21.88&	12.33 \\
D-MDETR ~\cite{shi2023dynamic}    & CLIP-B / CLIP-B     &    & 50.74& 37.72& 73.41&60.71 & 64.63&66.67 &65.06 &40.98 & 48.28& 22.92& 58.84& 35.43& 32.41 \\
D-MDETR ~\cite{shi2023dynamic}    & RN50+DETR / BERT-B  &    & -&-&-&-&-&-&-&-&-&-&-&-&-\\
MMCA ~\cite{yao2024visual}       & RN50+DETR / BERT-B  &    & -&-&-&-&-&-&-&-&-&-&-&-&- \\
FSVG ~\cite{wang2025simple} &  RN50+DETR / BERT-B & & 50.99&	34.43&	64.37&	61.90&	62.16&	60.00&	63.86&	43.03&	43.18&	27.08&	53.47&	42.97&	34.25\\
AttBalance ~\cite{kang2025visual} &  CLIP-B / CLIP-B & &-&-&-&-&-&-&-&-&-&-&-&-&-  \\
HiVG-B ~\cite{xiao2024hivg}     & CLIP-B / CLIP-B     &    & 66.54&	49.70&	79.89&	71.43&	75.00&	60.00&	81.33&	65.57&	53.41&	50.00&	73.41&	60.94&	44.52 \\
HiVG-L ~\cite{xiao2024hivg}     & CLIP-L / CLIP-L     &    & 67.97&	48.50&	80.46&	72.62&	80.41&	60.00&	80.72&	63.93&	53.41&	58.33&	73.41&	60.94&	43.15\\
OneRef-B ~\cite{xiao2024oneref}   & BEIT3-B / BEIT3-B   &    & 68.90&	50.30&	82.18&	72.62&	82.43&	60.00&	83.13&	61.07&	54.55&	58.33&	73.12&	64.06&	50.00 \\
OneRef-L ~\cite{xiao2024oneref}   & BEIT3-L / BEIT3-L   &    & 66.59&	43.11&	72.99&	69.05&	75.00&	60.00&	84.34&	62.30&	53.41&	52.08&	68.50&	61.72&	47.95 \\
\cline{3-3}
TransVG  ~\cite{deng2021transvg}    & RN50+DETR / BERT-B & \multirow{11}{*}{{TIR}}   & 48.46&	32.04&	75.86&	57.14&	60.81&	60.00&	63.25&	45.90&	43.18&	15.62&	49.71&	42.97&	25.34 \\
CLIP-VG ~\cite{xiao2023clip}    & CLIP-B / CLIP-B     &    & 29.12&	12.57&	35.06&	27.38&	31.08&	60.00&	47.59&	29.92&	18.18&	7.29&	28.32&	21.09&	13.70\\
D-MDETR ~\cite{shi2023dynamic}    & CLIP-B / CLIP-B     &    & 51.18&37.13 & 79.77&60.71 &61.90 &\textbf{100.00} &65.06 & 38.52& 52.87& 19.79 &56.81 & 39.37& 30.34\\
D-MDETR ~\cite{shi2023dynamic}    & RN50+DETR / BERT-B  &    & -&- & -&- & -& -& -& -& -&- & -&-& \\
MMCA ~\cite{yao2024visual}       & RN50+DETR / BERT-B  &    & -&-&-&-&-&-&-&-&-&-&-&-&- \\
FSVG ~\cite{wang2025simple} &  RN50+DETR / BERT-B & & 53.13&	36.83&	71.84&	66.67&	65.54&	60.00&	68.67&	47.54&	42.05&	25.00&	55.20&	42.97&	24.66\\
AttBalance ~\cite{kang2025visual} &  CLIP-B / CLIP-B & &  40.82&	20.96&	47.70&	54.76&	47.30&	60.00&	54.22&39.34	&29.55	&	20.83&	32.66&	33.59&	13.70\\
HiVG-B ~\cite{xiao2024hivg}     & CLIP-B / CLIP-B     &    & 64.12&	48.20&	81.61&	67.86&	72.97&	80.00&	83.73&	61.07&	59.09&	42.71&	70.23&	56.25&	35.62 \\
HiVG-L ~\cite{xiao2024hivg}     & CLIP-L / CLIP-L     &    &63.74&	50.60&	85.63&	71.43&	75.68&	80.00&	75.30&	65.16&	54.55&	46.88&	74.57&	58.59&	39.04 \\
OneRef-B ~\cite{xiao2024oneref}   & BEIT3-B / BEIT3-B   &    & 66.59&	49.10&	82.76&	71.43&	70.27&	80.00&	86.14&	61.48&	57.95&	50.00&	72.25&	60.94&	35.62 \\
OneRef-L ~\cite{xiao2024oneref}   & BEIT3-L / BEIT3-L   &    & 68.96&	52.69&	85.63&	73.81&	81.08&	\textbf{100.00}&	85.54&	62.70&	59.09&	56.25&	74.57&	66.41&	47.95\\
\midrule
\multicolumn{13}{l}{\textbf{c. In-domain training with multi-modal visual input (RGB+TIR):}} \\\midrule

MV-TransVG    & RN50+DETR / BERT-B & \multirow{12}{*}{{RGB+TIR}} & 48.41&	33.53&	80.46&	57.14&	58.78&	60.00&	61.45&	47.54&	43.18&	18.75&	52.02&	39.06&	27.40\\
MV-CLIP-VG   & CLIP-B / CLIP-B     &  &40.82&	20.06&	45.98&	45.24&	50.00&	60.00&	59.64&	45.49&	34.09&	15.62&	40.46&	28.91&	21.23\\
MV-D-MDETR    & CLIP-B / CLIP-B     &  & 47.72& 32.34&72.23 &55.95 & 57.82& 77.78& 59.64&38.93 & 48.28& 19.79& 51.88&35.43 & 25.52\\
MV-D-MDETR    & RN50+DETR / BERT-B  &  & 45.08&	30.84&	79.77&	57.14&	61.90&	66.67&	56.02&	37.30&	44.83&	16.67&	54.49&	38.58&	27.59\\
MV-MMCA      & RN50+DETR / BERT-B  &  & 49.53&	31.44&	69.36&	51.19&	61.22&	66.67&	63.86&	45.90&	37.93&	15.62&	52.46&	43.31&	26.90 \\
MV-FSVG &  RN50+DETR / BERT-B & &46.92&	36.23&	70.69&	57.14&	56.76&	70.00&	60.84&	36.48&	38.64&	23.96&	52.31&	36.72&	28.77 \\
MV-AttBalance &  CLIP-B / CLIP-B & & 52.20&	31.14&	74.71&	60.71&	63.51&	60.00&	64.46&	49.59&	43.18&	21.88&	56.65&	43.75&	26.03 \\
MV-HiVG-B      & CLIP-B / CLIP-B     &  & 71.98&	\underline{53.89}&	87.93&	\underline{77.38}&	79.05&	\textbf{100.00}&	87.35&	66.39&	56.82&	54.17&	\underline{78.61}&	64.84&	49.32 \\
MV-HiVG-L    & CLIP-L / CLIP-L     &  & \underline{73.74}&	53.59&	89.08&	76.19&	\underline{81.76}&	70.00&	\underline{89.16}&	\textbf{73.36}&	\underline{60.23}&	\underline{60.42}&	\underline{78.61}&	64.06&	43.84 \\
MV-OneRef-B   & BEIT3-B / BEIT3-B   &  & 65.82&	46.71&	82.18&	73.81&	77.70&	80.00&	81.33&	59.02&	52.27&	53.12&	69.08&	60.94&	43.15\\
MV-OneRef-L   &BEIT3-L / BEIT3-L   &  & 70.22&	53.89&	\underline{90.23}&	73.81&	80.41&	\textbf{100.00}&	87.35&	62.70&	55.68&	53.12&	77.17&	\underline{69.53}&	\underline{54.11} \\
\textbf{RGBT-VGNet (Ours)}  & \textbf{CLIP-B / CLIP-B} &  &
\textbf{74.90}&	\textbf{57.44}	&\textbf{91.95}	&\textbf{78.57}	&\textbf{83.67}	&88.89&	\textbf{92.26}&	\underline{68.29}&	\textbf{62.07}&	\textbf{63.54}&	\textbf{81.45}&	\textbf{73.64}	&\textbf{58.50}
\\
\bottomrule
\end{tabular}
}
\label{tab:scene_refm3fd_all_methods}
\end{table*}

\begin{table*}[!t]
\centering
\footnotesize
\caption{\textbf{Scene-level Acc@0.5 across 13 scene types} (UB, IT, RS, SU, HW, CP, BG, PL, RR, TN, MK, WF, ID)
on the RefMFAD sub-dataset of RGBT-GroundBench. All methods, backbones, and modalities are included, following the style of Table~4 and 5 in the manuscript.}
    \setlength{\tabcolsep}{0.75mm}         \renewcommand{\arraystretch}{1.2}
\resizebox{\textwidth}{!}{
\begin{tabular}{c|c|c| ccccccccccccc}
\toprule
\multirow{2}{*}{Method} &
\multirow{2}{*}{\begin{tabular}[c]{@{}c@{}}\textbf{Visual / Language}\\\textbf{Backbone}\end{tabular}} &
\multirow{2}{*}{Modality} &
\multicolumn{13}{c}{\textbf{RefMFAD (Scene Types )}} \\
& & &
UB & SU & RR & HW & RS & ID &PL &IT & TN & BG & CP & MK & WF\\
\midrule
\multicolumn{13}{l}{\textbf{a. Zero-shot transfer from RGB-pretrained models:}} \\\midrule

CLIP-VG ~\cite{xiao2023clip}    & CLIP-B / CLIP-B   & \multirow{5}{*}{{RGB}}     & 7.94&	9.53&	7.69&	7.41&	4.92&	16.67&	4.35&	8.16&	4.12&	6.93&	3.81&	0.00&	0.00 \\
HiVG-B ~\cite{xiao2024hivg}     & CLIP-B / CLIP-B   &      & 7.21&	6.51&	0.00&	9.09&	9.02&	11.11&	15.22&	5.93&	9.28&	8.91&	8.57&	0.00&	0.00 \\
HiVG-L ~\cite{xiao2024hivg}     & CLIP-L / CLIP-L   &      & 42.06&	34.88&	32.69&	42.76&	42.62&	27.78&	63.04&	35.16&	32.99&	48.02&	40.95&	50.00&	66.67 \\
OneRef-B ~\cite{xiao2024oneref}   & BEIT3-B / BEIT3-B &      &31.33&	26.98&	23.08&	27.95&	31.97&	22.22&	54.35&	28.19&	25.26&	29.70&	34.76&	33.33&	66.67 \\
OneRef-L ~\cite{xiao2024oneref}   & BEIT3-L / BEIT3-L &      &  38.42&	37.21&	23.08&	38.05&	33.61&	27.78&	45.65&	35.16&	28.35&	36.63&	38.10&	33.33&	83.33\\
\cline{3-3}
CLIP-VG ~\cite{xiao2023clip}    & CLIP-B / CLIP-B   & \multirow{5}{*}{{TIR}}     & 6.87&	4.79&	17.82&	13.10&	4.73&	0.00&	7.23&	9.43&	0.00&	1.04&	4.05&	1.56&	4.79\\
HiVG-B ~\cite{xiao2024hivg}     & CLIP-B / CLIP-B   &      & 20.61&	16.05&	13.46&	21.55&	18.03&	16.67&	22.83&	18.55&	22.68&	19.80&	19.05&	33.33&	50.00\\
HiVG-L ~\cite{xiao2024hivg}     & CLIP-L / CLIP-L   &      & 23.33&	20.47&	17.31&	20.88&	22.95&	11.11&	42.39&	21.22&	29.90&	26.24&	24.76&	33.33&	0.00 \\
OneRef-B ~\cite{xiao2024oneref}   & BEIT3-B / BEIT3-B &      & 16.67&	12.33&	13.46&	13.30&	18.03&	16.67&	33.70&	12.76&	17.53&	15.84&	19.52&	0.00&	66.67\\
OneRef-L ~\cite{xiao2024oneref}   & BEIT3-L / BEIT3-L &      & 18.30&	15.35&	9.62&	18.35&	17.21&	16.67&	35.87&	15.73&	15.46&	18.32&	21.90&	0.00&	50.00\\
\midrule
\multicolumn{13}{l}{\textbf{b. In-domain training with uni-modal visual input (RGB or TIR):}} \\ \midrule

TransVG  ~\cite{deng2021transvg}    & RN50+DETR / BERT-B & \multirow{11}{*}{{RGB}}   & 47.21&	49.30&	46.15&	56.73&	36.89&	50.00&	61.96&	41.99&	64.95&	53.96&	41.90&	33.33&	83.33 \\
CLIP-VG ~\cite{xiao2023clip}    & CLIP-B / CLIP-B     &    &38.97&	34.88&	32.69&	39.39&	25.41&	22.22&	50.00&	34.27&	47.94&	42.08&	26.19&	33.33&	66.67 \\
D-MDETR ~\cite{shi2023dynamic}    & CLIP-B / CLIP-B     &    &57.91 & 64.88& 73.08&65.60&49.18&44.44&68.45&52.30&75.77&57.43&53.11&60.00&60.00\\
D-MDETR ~\cite{shi2023dynamic}    & RN50+DETR / BERT-B  &    & 49.30&	52.79&	50.00&	60.37&	44.26&	38.89&	57.61&	44.13&	67.53&	56.44&	41.63&	60.00&	80.00 \\
MMCA ~\cite{yao2024visual}       & RN50+DETR / BERT-B  &    & 44.15&	48.14&	42.31&	56.32&	34.43&	44.44&	58.70&	38.63&	65.98&	51.98&	40.19&	20.00&	60.00\\
FSVG ~\cite{wang2025simple} &  RN50+DETR / BERT-B & & 56.85&	63.26&	67.31&	64.14&	45.08&	38.89&	72.83&	54.15&	73.71&	60.40&	48.57&	66.67&	66.67\\
AttBalance ~\cite{kang2025visual} &  CLIP-B / CLIP-B & & 49.82&	51.63&	42.31&	60.27&	44.26&	38.89&	60.87&	45.85&	64.95&	53.47&	43.81&	50.00&	66.67 \\
HiVG-B ~\cite{xiao2024hivg}     & CLIP-B / CLIP-B     &    & 60.91&	66.28&	63.46&	68.69&	62.30&	50.00&	80.43&	56.23&	74.23&	65.84&	58.57&	66.67&	83.33 \\
HiVG-L ~\cite{xiao2024hivg}     & CLIP-L / CLIP-L     &    & 61.70&	64.42&	65.38&	68.18&	56.56&	\textbf{55.56}&	81.52&	57.12&	73.71&	64.36&	59.05&	66.67&	83.33 \\
OneRef-B ~\cite{xiao2024oneref}   & BEIT3-B / BEIT3-B   &    & 59.33&	62.56&	61.54&	65.82&	57.38&	\textbf{55.56}&	79.35&	53.71&	72.68&	63.86&	55.24&	50.00&	83.33\\
OneRef-L ~\cite{xiao2024oneref}   & BEIT3-L / BEIT3-L   &    & 61.76&	66.98&	65.38&	67.85&	54.10&	44.44&	78.26&	56.97&	76.29&	64.85&	61.43&	33.33&	83.33 \\
\cline{3-3}
TransVG  ~\cite{deng2021transvg}    & RN50+DETR / BERT-B & \multirow{11}{*}{{TIR}}   & 50.42&	53.95&	51.92&	60.94&	43.44&	38.89&	56.52&	45.40&	69.59&	56.44&	47.62&	50.00&	83.33 \\
CLIP-VG ~\cite{xiao2023clip}    & CLIP-B / CLIP-B     &    & 32.55&	26.28&	19.23&	36.87&	23.77&	22.22&	42.39&	24.18&	40.72&	35.64&	20.95&	0.00&	16.67\\
D-MDETR ~\cite{shi2023dynamic}    & CLIP-B / CLIP-B     &    & 52.88& 59.77& 67.31& 61.05& 39.34&50.00 & 60.87&47.99 & 75.26&55.45 & 43.06& 80.00&60.00 \\
D-MDETR ~\cite{shi2023dynamic}    & RN50+DETR / BERT-B  &    & 47.91&	51.16&	57.69&	60.54&	40.16&	38.89&	58.70&	41.46&	71.13&	55.94&	38.28&	20.00&	80.00 \\
MMCA ~\cite{yao2024visual}       & RN50+DETR / BERT-B  &    &43.54&	46.28&	48.08&	54.30&	33.61&	44.44&	48.91&	37.44&	67.01&	50.50&	35.89&	40.00&	60.00\\
FSVG ~\cite{wang2025simple} &  RN50+DETR / BERT-B & &54.30&	58.37&	53.85&	61.95&	44.26&	27.78&	58.70&	46.44&	71.65&	56.44&	42.38&	50.00&	83.33 \\
AttBalance ~\cite{kang2025visual} &  CLIP-B / CLIP-B & & 45.76&	45.12&	42.31&	53.03&	29.51&	50.00&	57.61&	37.69&	59.28&	48.02&	37.14&	33.33&	66.67 \\
HiVG-B ~\cite{xiao2024hivg}     & CLIP-B / CLIP-B     &    & 58.24&	63.95&	65.38&	65.99&	51.64&	38.89&	72.83&	50.89&	73.71&	62.87&	51.43&	50.00&	83.33 \\
HiVG-L ~\cite{xiao2024hivg}     & CLIP-L / CLIP-L     &    &58.79&	61.16&	67.31&	63.64&	52.46&	55.56&	67.39&	50.89&	71.13&	63.37&	51.43&	50.00&	66.67\\
OneRef-B ~\cite{xiao2024oneref}   & BEIT3-B / BEIT3-B   &    & 52.79&	50.93&	44.23&	60.10&	47.54&	44.44&	58.70&	44.51&	69.59&	54.95&	46.67&	50.00&	83.33\\
OneRef-L ~\cite{xiao2024oneref}   & BEIT3-L / BEIT3-L   &    & 59.09&	63.72&	65.38&	65.15&	51.64&	50.00&	77.17&	53.56&	77.32&	58.42&	50.95&	\textbf{83.33}&	66.67 \\
\midrule
\multicolumn{13}{l}{\textbf{c. In-domain training with multi-modal visual input (RGB+TIR):}} \\\midrule

MV-TransVG    & RN50+DETR / BERT-B & \multirow{12}{*}{{RGB+TIR}} & 51.52&	54.19&	53.85&	64.48&	45.90&	50.00&	61.96&	45.40&	71.65&	59.41&	49.52&	66.67&	66.67\\
MV-CLIP-VG   & CLIP-B / CLIP-B     &  &48.24&	47.67&	40.38&	52.69&	36.07&	22.22&	56.52&	42.88&	59.79&	50.99&	35.24&	33.33&	66.67\\
MV-D-MDETR    & CLIP-B / CLIP-B     &  & 55.55& 62.09&69.23& 63.74& 43.44& 50.00&64.13 &50.67&75.77 & 57.92& 50.72& 40.00&80.00 \\
MV-D-MDETR    & RN50+DETR / BERT-B  &  & 51.79&	55.58&	59.62&	63.41&	41.80&	50.00&	59.78&	48.44&	73.71&	60.40&	42.58&	40.00&	60.00 \\
MV-MMCA      & RN50+DETR / BERT-B  &  & 53.12&	56.51&	55.77&	63.07&	45.08&	50.00&	58.70&	46.36&	70.10&	59.41&	44.98&	60.00&	80.00\\
MV-FSVG &  RN50+DETR / BERT-B & &57.27&	61.16&	71.15&	65.82&	49.18&	55.56&	67.39&	50.59&	75.26&	59.90&	51.90&	33.33&	83.33 \\
MV-AttBalance &  CLIP-B / CLIP-B & & 51.09&	51.40&	46.15&	61.28&	48.36&	44.44&	60.87&	46.59&	70.62&	56.44&	44.76&	50.00&	83.33 \\
MV-HiVG-B      & CLIP-B / CLIP-B     &  & \underline{64.42}&	\textbf{71.40}&	\textbf{75.00}&	\underline{70.20}&	\textbf{63.11}&	50.00&	\textbf{85.87}&	\underline{58.75}&	\underline{78.35}&	\underline{68.32}&	\underline{61.43}&	66.67&	83.33\\
MV-HiVG-L    & CLIP-L / CLIP-L     &  & \underline{62.42}&	65.58&	67.31&	66.50&	61.48&	50.00&	72.83&	57.42&	74.23&	67.33&	58.10&	66.67&	83.33 \\
MV-OneRef-B   & BEIT3-B / BEIT3-B   &  & 58.73&	64.88&	69.23&	66.84&	57.38&	44.44&	75.00&	53.86&	73.71&	60.89&	54.76&	66.67&	66.67 \\
MV-OneRef-L   & BEIT3-L / BEIT3-L   &  &61.88&	65.81&	67.31&	68.86&	54.10&	50.00&	78.26&	55.49&	76.29&	62.87&	55.71&	\textbf{83.33}&	83.33 \\
\textbf{RGBT-VGNet (Ours)}  & \textbf{CLIP-B / CLIP-B} &\textbf{ } &
\textbf{65.03}&	\underline{68.75}	&\underline{70.37}&	\textbf{70.37}	&\textbf{63.41}	&44.44	&\underline{78.49}	&\textbf{62.52}	&\underline{79.49}&	\textbf{70.10}&	\textbf{62.38}&	66.67&	83.33 \\
\bottomrule
\end{tabular}
}
\label{tab:scene_refmfad_all_methods}
\end{table*}

\begin{table*}[!t]
\centering
\footnotesize
\caption{\textbf{Acc@0.5 across four weather conditions} (Cloudy, Foggy, Rainy, Sunny) 
for RefFLIR, RefM$^3$FD, and RefMFAD. All methods, backbones, and modalities are included, 
following the style of Table~4 and Table~5 in the manuscript.}
    \setlength{\tabcolsep}{1mm}         \renewcommand{\arraystretch}{1.2}
\resizebox{\textwidth}{!}{
\begin{tabular}{c|c|c| cccc | cccc | cccc}
\toprule
\multirow{2}{*}{Method} &
\multirow{2}{*}{\begin{tabular}[c]{@{}c@{}}\textbf{Visual / Language}\\\textbf{Backbone}\end{tabular}} &
\multirow{2}{*}{Modality} &
\multicolumn{4}{c|}{\textbf{RefFLIR (Weather )}} &
\multicolumn{4}{c|}{\textbf{RefM$^3$FD (Weather )}} &
\multicolumn{4}{c}{\textbf{RefMFAD (Weather )}} \\
& & &
FY & RY & SY & CY &
FY & RY & SY & CY &
FY & RY & SY & CY  \\
\midrule
\multicolumn{13}{l}{\textbf{a. Zero-shot transfer from RGB-pretrained models:}} \\\midrule

CLIP-VG ~\cite{xiao2023clip}    & CLIP-B / CLIP-B   & \multirow{5}{*}{{RGB}}     & 3.48&	5.26&	9.65&	6.28&
4.55&	6.90&	9.79&	8.86&
5.84&	4.35&	9.27&	8.19 \\
HiVG-B ~\cite{xiao2024hivg}     & CLIP-B / CLIP-B   &     & 26.96&	44.74&	36.14&	31.15&
36.36&	33.07&	29.79&	30.96&
17.05&	18.94&	25.17&	20.14\\
HiVG-L ~\cite{xiao2024hivg}     & CLIP-L / CLIP-L   &      &30.43&	52.63&	49.39&	32.79&
52.45&	56.27&	52.54&	52.72&
35.35&	28.26&	44.04&	43.08 \\
OneRef-B ~\cite{xiao2024oneref}   & BEIT3-B / BEIT3-B &      &29.57&	44.74&	46.13&	32.24&
39.16&	42.79&	37.96&	40.10&
25.40&	23.29&	31.13&	32.27 \\
OneRef-L ~\cite{xiao2024oneref}   & BEIT3-L / BEIT3-L &      & 33.04&	44.74&	55.23&	37.70&
43.36&	50.47&	45.92&	46.86&
33.18&	28.57&	40.40&	38.75\\
\cline{3-3}
CLIP-VG ~\cite{xiao2023clip}    & CLIP-B / CLIP-B   &\multirow{5}{*}{{TIR}}     & 4.35&	13.16&	8.83&	7.10&
8.39&	7.99&	3.31&	8.93&
3.09&	2.80&	7.62&	4.48 \\
HiVG-B ~\cite{xiao2024hivg}     & CLIP-B / CLIP-B   &   & 13.91&	36.84&	11.75&	13.93&
18.53&	16.93&	10.21&	9.14&
5.95&	4.04&	11.92&	8.01\\
HiVG-L ~\cite{xiao2024hivg}     & CLIP-L / CLIP-L   &    &20.43&	44.74&	24.86&	24.59&
30.77&	41.07&	27.75&	25.73&
20.48&	16.77&	33.11&	23.53 \\
OneRef-B ~\cite{xiao2024oneref}   & BEIT3-B / BEIT3-B &     & 20.87&	42.11&	27.92&	24.04&
20.63&	27.59&	21.20&	22.80&
12.13&	10.25&	23.18&	16.53 \\
OneRef-L ~\cite{xiao2024oneref}   & BEIT3-L / BEIT3-L &      &25.65&	57.89&	27.31&	24.32&
30.77&	29.47&	20.28&	23.08&
16.93&	15.22&	21.19&	18.06  \\
\midrule
\multicolumn{13}{l}{\textbf{b. In-domain training with uni-modal visual input (RGB or TIR):}} \\ \midrule

TransVG  ~\cite{deng2021transvg}    & RN50+DETR / BERT-B & \multirow{11}{*}{{RGB}}   & 34.35&	47.37&	45.45&	34.70&
52.10&	44.51&	36.27&	39.54&
47.03&	44.41&	63.58&	48.25\\
CLIP-VG ~\cite{xiao2023clip}    & CLIP-B / CLIP-B     &    & 18.26&	36.84&	27.31&	16.94&
31.12&	35.89&	29.58&	30.61&
34.67&	30.12&	51.66&	37.65\\
D-MDETR ~\cite{shi2023dynamic}    & CLIP-B / CLIP-B     &    &40.00&54.05 & 53.91&37.98 & 56.49&54.55 &46.79 &50.35& 56.64& 51.24&77.15 & 59.40\\
D-MDETR ~\cite{shi2023dynamic}    & RN50+DETR / BERT-B  &    & 38.26&	48.65&	48.81&	30.87& -&-&-&-&50.34&	41.93&	64.57&	51.13 \\
MMCA ~\cite{yao2024visual}       & RN50+DETR / BERT-B  &    & 35.65&	51.35&	51.67&	34.43&
-&	-&	-&	-&
45.88&	36.65&	63.25&	46.10 \\
FSVG ~\cite{wang2025simple} &  RN50+DETR / BERT-B & &40.43&	50.00&	48.64&	34.15&	55.94&	52.51&	47.89&	48.26&	55.95&	49.69&	73.18&	58.96 \\
AttBalance ~\cite{kang2025visual} &  CLIP-B / CLIP-B & & 39.13&	52.63&	52.92&	39.62&	-&-&-&-&	48.63&	44.72&	66.89&	51.20 \\
HiVG-B ~\cite{xiao2024hivg}     & CLIP-B / CLIP-B     &    & 49.13&	63.16&	71.74&	53.01&
70.28&	67.71&	63.59&	65.55&
58.58&	54.66&	78.48&	63.66 \\
HiVG-L ~\cite{xiao2024hivg}     & CLIP-L / CLIP-L     &    & 59.13&	68.42&	74.80&	59.29&
67.83&	67.08&	65.70&	66.53&
58.70&	54.97&	75.83&	63.87\\
OneRef-B ~\cite{xiao2024oneref}   & BEIT3-B / BEIT3-B   &    & 50.43&	55.26&	66.92&	51.37&
70.28&	69.28&	65.21&	68.34&
56.64&	52.80&	74.50&	61.36 \\
OneRef-L ~\cite{xiao2024oneref}   & BEIT3-L / BEIT3-L   &    & 53.91&	60.53&	70.04&	52.19&
67.13&	68.34&	62.68&	63.67&
59.15&	54.97&	77.48&	63.91\\
\cline{3-3}
TransVG  ~\cite{deng2021transvg}    & RN50+DETR / BERT-B & \multirow{11}{*}{{TIR}}   & 36.96&	44.74&	45.92&	34.15&
58.39&	54.23&	42.39&	47.56&
50.34&	47.20&	68.54&	51.86\\
CLIP-VG ~\cite{xiao2023clip}    & CLIP-B / CLIP-B     &    & 35.22&	47.37&	38.72&	32.79&
33.92&	34.64&	23.31&	26.22&
29.75&	24.53&	40.07&	30.84 \\
D-MDETR ~\cite{shi2023dynamic}    & CLIP-B / CLIP-B     &    & 41.74&	54.05&	48.40&	40.44&64.56 & 53.29& 45.45& 51.05&54.69 &45.65 &69.87 & 53.93\\
D-MDETR ~\cite{shi2023dynamic}    & RN50+DETR / BERT-B  &    & 47.83&	51.35&	43.44&	39.07&-&-&-&-&50.00&	42.55&	66.23&	49.16 \\
MMCA ~\cite{yao2024visual}       & RN50+DETR / BERT-B  &    & 46.52&	45.95&	42.28&	35.79&
-&	-&	-&	-&
45.42&	38.82&	60.93&	44.28 \\
FSVG ~\cite{wang2025simple} &  RN50+DETR / BERT-B & & 42.61&	42.11&	38.52&	35.25&63.64&	59.09&	47.89&	49.30&55.38&	48.76&	67.88&	53.46\\
AttBalance ~\cite{kang2025visual} &  CLIP-B / CLIP-B & &  40.43&	55.26&	45.04&	37.16&	42.31&	42.32&	35.07&	37.31&	43.48&	40.68&	58.28&	45.23\\
HiVG-B ~\cite{xiao2024hivg}     & CLIP-B / CLIP-B     &    & 63.04&	73.68&	56.79&	59.29&
77.97&	69.28&	57.82&	62.90&
58.70&	49.07&	73.84&	59.32\\
HiVG-L ~\cite{xiao2024hivg}     & CLIP-L / CLIP-L     &    &66.09&	76.32&	64.67&	69.13&
81.12&	72.88&	56.97&	63.74&
58.58&	53.11&	72.52&	58.23\\
OneRef-B ~\cite{xiao2024oneref}   & BEIT3-B / BEIT3-B   &    &58.26&	68.42&	60.46&	60.93&
77.62&	71.94&	59.65&	64.99&
51.49&	47.20&	67.22&	52.15\\
OneRef-L ~\cite{xiao2024oneref}   & BEIT3-L / BEIT3-L   &    & 60.43&	65.79&	61.96&	60.93&
81.12&	73.20&	61.97&	69.87&
59.15&	53.73&	71.19&	60.01 \\
\midrule
\multicolumn{13}{l}{\textbf{c. In-domain training with multi-modal visual input (RGB+TIR):}}\\
\midrule
MV-TransVG    & RN50+DETR / BERT-B & \multirow{12}{*}{{RGB+TIR}} &40.87&	55.26&	48.98&	38.80&
59.44&	53.92&	43.59&	47.63&
52.17&	45.96&	71.85&	53.42\\
MV-CLIP-VG   & CLIP-B / CLIP-B     &  &33.04&	47.37&	43.34&	35.25&
45.80&	46.24&	34.44&	38.77&
44.28&	38.82&	60.60&	48.18\\
MV-D-MDETR    & CLIP-B / CLIP-B     &  &41.74& 56.76& 50.44& 40.44 &58.25 &50.16& 42.35& 46.65&55.15 &49.07 & 74.83& 57.14\\
MV-D-MDETR    & RN50+DETR / BERT-B  &  & 43.91&	59.46&	49.97&	42.62&61.05&	51.57&	39.04&	46.37&54.23&	46.89&	68.54&	53.46 \\
MV-MMCA      & RN50+DETR / BERT-B  &  & 43.91&	62.16&	52.07&	39.89&
54.04&	54.55&	43.20&	48.19&
53.20&	46.58&	69.21&	54.08 \\
MV-FSVG &  RN50+DETR / BERT-B & & 42.61&	55.26&	51.09&	37.98&	59.09&	53.13&	40.92&	46.16&	56.52&	49.38&	73.51&	58.70\\
MV-AttBalance &  CLIP-B / CLIP-B & & 48.26&	57.89&	56.25&	48.36&	59.44&	55.96&	46.34&	50.49&	51.26&	47.83&	69.21&	51.93 \\
MV-HiVG-B      & CLIP-B / CLIP-B     &  &63.04&	\underline{73.68}&	69.23&	65.57&
\underline{82.17}&	\underline{75.71}&	65.70&	71.48&
\underline{64.07}&	\underline{60.56}&	 \underline{77.81}&	\underline{66.24} \\
MV-HiVG-L    & CLIP-L / CLIP-L     &  &\textbf{70.00}&	\underline{73.68}&	\underline{75.48}&	74.04&
81.82&	\underline{75.71}&	\underline{68.38}&	\underline{72.04}&
60.76&	55.90&	73.84&	63.77 \\
MV-OneRef-B   & BEIT3-B / BEIT3-B   &  &56.96&	\underline{73.68}&	64.88&	54.10&
74.83&	66.30&	60.42&	65.48&
58.35&	50.31&	74.17&	61.14 \\
MV-OneRef-L   & BEIT3-L / BEIT3-L   &  & 52.17&	65.79&	64.47&	53.83&
80.42&	72.73&	65.00&	70.92&
59.95&	53.11&	 \underline{77.81}&	63.22\\
\textbf{RGBT-VGNet (Ours)}  & \textbf{CLIP-B / CLIP-B} & \textbf{} &
 \underline{65.80}&	\textbf{79.49}&\textbf{	75.63}& \underline{	67.49}&
\textbf{84.91}&	\textbf{79.50}&	\textbf{68.71}&\textbf{	75.52}&
\textbf{65.07}&	\textbf{61.42}&	\textbf{77.89}	& \textbf{66.48}\\
\bottomrule
\end{tabular}
}

\label{tab:weather_all_methods}
\end{table*}

\begin{table*}[!t]
\centering
\footnotesize
    \setlength{\tabcolsep}{1mm}         \renewcommand{\arraystretch}{1.2}
    
\caption{\textbf{Acc@0.5 across four illumination conditions} (Strong, Normal, Weak, Very Weak) 
on RefFLIR, RefM$^3$FD, and RefMFAD for all methods, backbones, and modalities.}
\resizebox{\textwidth}{!}{
\begin{tabular}{c|c|c| cccc | cccc | cccc}
\toprule
\multirow{2}{*}{Method} &
\multirow{2}{*}{\begin{tabular}[c]{@{}c@{}}\textbf{Visual / Language}\\\textbf{Backbone}\end{tabular}} &
\multirow{2}{*}{Modality} &
\multicolumn{4}{c|}{\textbf{RefFLIR (Illumination )}} &
\multicolumn{4}{c|}{\textbf{RefM$^3$FD (Illumination )}} &
\multicolumn{4}{c}{\textbf{RefMFAD (Illumination )}} \\
& & &
VL & WL & NL & SL &
VL & WL & NL & SL &
VL & WL & NL & SL  \\
\midrule
\multicolumn{13}{l}{\textbf{a. Zero-shot transfer from RGB-pretrained models:}} \\\midrule

CLIP-VG ~\cite{xiao2023clip}    & CLIP-B / CLIP-B   & \multirow{5}{*}{{RGB}}     & 12.50&	5.13&	0.00&	9.69&
12.86&	8.77&	8.48&	6.70&
8.33&	6.92&	8.20&	16.67 \\
HiVG-B ~\cite{xiao2024hivg}     & CLIP-B / CLIP-B   &      &37.50&	30.29&	50.00&	36.15&
48.57&	34.02&	30.31&	22.68&
22.22&	19.90&	19.59&	0.00 \\
HiVG-L ~\cite{xiao2024hivg}     & CLIP-L / CLIP-L   &      &50.00&	32.85&	50.00&	49.45&
65.71&	56.73&	51.65&	50.00&
27.78&	38.02&	43.97&	16.67 \\
OneRef-B ~\cite{xiao2024oneref}   & BEIT3-B / BEIT3-B &      & 37.50&	31.89&	66.67&	46.04&
54.29&	48.25&	36.37&	30.93&
25.00&	28.28&	32.76&	0.00 \\
OneRef-L ~\cite{xiao2024oneref}   & BEIT3-L / BEIT3-L &      &50.00&	36.22&	66.67&	55.25&
52.86&	52.92&	44.90&	37.63&
36.11&	35.20&	39.40&	0.00\\
\cline{3-3}
CLIP-VG ~\cite{xiao2023clip}    & CLIP-B / CLIP-B   & \multirow{5}{*}{{TIR}}     & 12.50&	6.41&	0.00&	8.87&
12.86&	11.01&	4.98&	1.55&
5.56&	4.02&	4.58&	16.67 \\
HiVG-B ~\cite{xiao2024hivg}     & CLIP-B / CLIP-B   &      &37.50&	15.06&	16.67&	11.73&
18.57&	18.13&	9.00&	6.19&
2.78&	6.88&	8.59&	0.00\\
HiVG-L ~\cite{xiao2024hivg}     & CLIP-L / CLIP-L   &      &37.50&	24.20&	16.67&	24.90&
52.86&	36.35&	26.25&	25.77&
25.00&	20.19&	26.95&	0.00 \\
OneRef-B ~\cite{xiao2024oneref}   & BEIT3-B / BEIT3-B &      & 25.00&	23.88&	33.33&	27.97&
35.71&	32.75&	18.81&	17.53&
16.67&	13.47&	18.53&	0.00 \\
OneRef-L ~\cite{xiao2024oneref}   & BEIT3-L / BEIT3-L &      &50.00&	26.28&	33.33&	27.35&
24.29&	34.11&	19.98&	14.95&
19.44&	16.38&	19.81&	0.00\\
\midrule
\multicolumn{13}{l}{\textbf{b. In-domain training with uni-modal visual input (RGB or TIR):}} \\ \midrule

TransVG  ~\cite{deng2021transvg}    & RN50+DETR / BERT-B & \multirow{11}{*}{{RGB}}   & 87.50&	34.62&	66.67&	45.43&
38.57&	46.59&	38.38&	28.35&
44.44&	46.52&	51.95&	33.33 \\
CLIP-VG ~\cite{xiao2023clip}    & CLIP-B / CLIP-B     &    & 25.00&	18.59&	33.33&	27.22&
31.43&	40.25&	27.89&	25.26&
33.33&	34.41&	41.69&	16.67 \\
D-MDETR ~\cite{shi2023dynamic}    & CLIP-B / CLIP-B     &    &62.50&39.26 &66.67 &53.89 & 50.72& 57.46& 47.91& 40.72& 60.00 & 55.87& 64.17&\textbf{100.00} \\
D-MDETR ~\cite{shi2023dynamic}    & RN50+DETR / BERT-B  &    & 87.50&	33.81&	66.67&	48.77 & - &	-	&-&	-&51.43	&48.73&	54.58&	50.00\\
MMCA ~\cite{yao2024visual}       & RN50+DETR / BERT-B  &    & 75.00&	35.26&	66.67&	51.64&
-&	-&	-&	-&
45.71&	43.92&	50.11&	50.00 \\
FSVG ~\cite{wang2025simple} &  RN50+DETR / BERT-B & & 75.00&	36.86&	66.67&	48.57&	48.57&	55.17&	47.87&	39.18&	61.11&	55.31&	63.00&	66.67\\
AttBalance ~\cite{kang2025visual} &  CLIP-B / CLIP-B & & 75.00&	39.74&	66.67&	52.86&	-&-&-&-&	50.00&	48.84&	54.63&	50.00 \\
HiVG-B ~\cite{xiao2024hivg}     & CLIP-B / CLIP-B     &    & 87.50&	51.76&	66.67&	71.76&
71.43&	69.40&	64.23&	59.79&
55.56&	59.83&	67.24&	83.33 \\
HiVG-L ~\cite{xiao2024hivg}     & CLIP-L / CLIP-L     &    &87.50&	59.46&	66.67&	74.83&
77.14&	69.20&	65.55&	58.76&
66.67&	60.28&	66.46&	83.33\\
OneRef-B ~\cite{xiao2024oneref}   & BEIT3-B / BEIT3-B   &    &87.50&	50.80&	66.67&	66.92&
84.29&	71.44&	66.08&	58.25&
66.67&	57.55&	64.73&	50.00\\
OneRef-L ~\cite{xiao2024oneref}   & BEIT3-L / BEIT3-L   &    & 87.50&	52.88&	66.67&	70.05&
78.57&	69.49&	62.42&	56.70&
61.11&	59.49&	68.14&	83.33 \\
\cline{3-3}
TransVG  ~\cite{deng2021transvg}    & RN50+DETR / BERT-B & \multirow{11}{*}{{TIR}}   & 75.00&	35.26&	66.67&	45.91&
54.29&	55.75&	45.14&	32.99&
52.78&	49.83&	55.64&	83.33 \\
CLIP-VG ~\cite{xiao2023clip}    & CLIP-B / CLIP-B     &    & 62.50&	34.13&	66.67&	38.61&
35.71&	38.89&	22.79&	18.56&
33.33&	28.73&	33.48&	33.33 \\
D-MDETR ~\cite{shi2023dynamic}    & CLIP-B / CLIP-B     &    &50.00&	41.51&	66.67&	48.36 &  55.07& 57.85& 48.15& 37.11&48.57 & 51.80& 58.37& 83.33\\
D-MDETR ~\cite{shi2023dynamic}    & RN50+DETR / BERT-B  &    &87.50 &42.31 & 66.67& 43.34& -& -& -& -& 42.86&	48.03&	52.90&	50.00 \\
MMCA ~\cite{yao2024visual}       & RN50+DETR / BERT-B  &    & 87.50&	39.58&	66.67&	42.22&
-&	-&	-&	-&
45.71&	42.93&	48.44&	50.00 \\
FSVG ~\cite{wang2025simple} &  RN50+DETR / BERT-B & & 87.50&	37.66&	66.67&	38.47&64.29&	60.14&	48.55&	39.18&50.00&	52.94&	56.58&	83.33\\
AttBalance ~\cite{kang2025visual} &  CLIP-B / CLIP-B & & 75.00&	38.94&	66.67&	45.02&	35.71&	46.59&	34.97&	26.29&	41.67&	43.45&	48.21&	33.33 \\
HiVG-B ~\cite{xiao2024hivg}     & CLIP-B / CLIP-B     &    &87.50&	61.22&	66.67&	56.75&
80.00&	70.66&	60.57&	51.55&
52.78&	56.84&	63.00&	83.33 \\
HiVG-L ~\cite{xiao2024hivg}     & CLIP-L / CLIP-L     &    & \textbf{100.00}&	\underline{67.95}&	\textbf{83.33}&	64.60&
68.57&	73.29&	61.17&	50.52&
52.78&	57.09&	61.55&	50.00 \\
OneRef-B ~\cite{xiao2024oneref}   & BEIT3-B / BEIT3-B   &    & 87.50&	60.10&	\textbf{83.33}&	60.37&
88.57&	72.03&	62.50&	53.61&
47.22&	50.83&	55.25&	66.67 \\
OneRef-L ~\cite{xiao2024oneref}   & BEIT3-L / BEIT3-L   &    &87.50&	60.74&	66.67&	61.94&
80.00&	75.05&	66.32&	54.12&
55.56&	57.63&	63.50&	83.33 \\

\midrule
\multicolumn{13}{l}{\textbf{c. In-domain training with multi-modal visual input (RGB+TIR):}} \\\midrule

MV-TransVG    & RN50+DETR / BERT-B & \multirow{12}{*}{{RGB+TIR}} & 87.50&	39.90&	66.67&	48.91&
54.29&	55.46&	45.66&	37.63&
44.44&	51.37&	57.42&	83.33 \\
MV-CLIP-VG   & CLIP-B / CLIP-B     &  & 50.00&	34.94&	66.67&	43.25&
42.86&	50.58&	34.61&	30.93&
50.00&	45.02&	50.84&	50.00 \\
MV-D-MDETR    & CLIP-B / CLIP-B     &  & 62.50& 41.51&66.67 &50.41 & 50.72& 54.05& 44.17& 35.05& 60.00&53.84&61.94&\textbf{100.00} \\
MV-D-MDETR    & RN50+DETR / BERT-B  &  & 75.00&	43.59&	66.67&	49.93 & 46.38&	54.34&	43.01	&32.47 &60.00&	52.01&	57.03	&50.00 \\
MV-MMCA      & RN50+DETR / BERT-B  &  & 87.50&	41.99&	66.67&	52.05&
49.28&	57.27&	44.94&	34.54&
54.29&	51.51&	58.31&	50.00 \\
MV-FSVG &  RN50+DETR / BERT-B & & 62.50&	40.38&	66.67&	51.02&	41.43&	55.75&	43.73&	31.96&	52.78&	55.31&	63.06&	66.67\\
MV-AttBalance &  CLIP-B / CLIP-B & & 87.50&	48.40&	66.67&	56.21&	64.29&	58.09&	47.83&	40.21&	52.78&	51.04&	54.85&	83.33\\
MV-HiVG-B      & CLIP-B / CLIP-B     &  & 87.50&	64.74&	66.67&	69.24&
\textbf{92.86}&	77.19&	68.45&	59.79&
63.89&	\underline{63.10}&	\textbf{70.37}&	66.67 \\
MV-HiVG-L    & CLIP-L / CLIP-L     &  & 87.50&	\textbf{72.28}&	\textbf{83.33}&	\underline{75.44}&
87.14&	\underline{77.68}&	\underline{69.98}&	\underline{62.89}&
\underline{63.89}&	60.78&	66.52&	83.33\\
MV-OneRef-B   & BEIT3-B / BEIT3-B   &  &87.50&	55.77&	66.67&	64.87&
78.57&	68.91&	63.06&	53.09&
58.33&	57.71&	64.62&	66.67 \\
MV-OneRef-L   & BEIT3-L / BEIT3-L   &  & 75.00&	53.69&	66.67&	64.46&
81.43&	74.76&	68.17&	58.76&
61.11&	59.49&	67.19&	83.33 \\
RGBT-VGNet (Ours)  & CLIP-B / CLIP-B &  &
 \underline{88.89}&	 \underline{67.63}&	66.67&	\textbf{75.73}&
 \underline{89.86}&	\textbf{81.48}&\textbf{	71.81}&\textbf{	65.13}&
\textbf{66.67}&	\textbf{64.05}&	 \underline{70.07}&	 \underline{83.33} \\
\bottomrule
\end{tabular}
}
\label{tab:illum_all_methods}
\end{table*}

\begin{table*}[!t]
\centering
\footnotesize
    \setlength{\tabcolsep}{1mm}         \renewcommand{\arraystretch}{1.2}
    
\caption{\textbf{Acc@0.5 across object sizes } (Normal Size vs Small Size) and \textbf{different occlusion levels} (No-or-Partial vs Heavy) for RefFLIR, RefM$^3$FD, and RefMFAD.}
\resizebox{\textwidth}{!}{
\begin{tabular}{c|c|c|cccc|cccc|cccc}
\toprule
\multirow{2}{*}{Method} &
\multirow{2}{*}{\begin{tabular}[c]{@{}c@{}}\textbf{Visual / Language}\\\textbf{Backbone}\end{tabular}} &
\multirow{2}{*}{Modality} &
\multicolumn{4}{c|}{\textbf{RefFLIR (Size and Occ)}} &
\multicolumn{4}{c|}{\textbf{RefM$^3$FD (Size and Occ)}} &
\multicolumn{4}{c}{\textbf{RefMFAD (Size and Occ)}} \\
& & &
NS & SS & PO& HO&
NS & SS &PO& HO&
NS & SS&PO& HO \\
\midrule
\multicolumn{13}{l}{\textbf{a. Zero-shot transfer from RGB-pretrained models:}} \\\midrule

CLIP-VG ~\cite{xiao2023clip}    & CLIP-B / CLIP-B   & \multirow{5}{*}{{RGB}}     &14.44&	1.11&	11.36&	4.12&
15.46&	1.35&	12.09&	4.03&
16.88&	1.22&	12.41&	4.52  \\
HiVG-B ~\cite{xiao2024hivg}     & CLIP-B / CLIP-B   &      & 62.06&	2.23&	48.48&	25.66&
56.12&	5.30&	42.01&	23.08&
46.53&	1.92&	42.20&	17.87 \\
HiVG-L ~\cite{xiao2024hivg}     & CLIP-L / CLIP-L   &      & 75.97&	7.81&	60.61&	29.92&
85.53&	19.59&	76.43&	40.35&
81.76&	12.86&	69.86&	32.20  \\
OneRef-B ~\cite{xiao2024oneref}   & BEIT3-B / BEIT3-B &      & 74.47&	3.55&	55.30&	28.86&
74.59&	3.19&	59.63&	29.22&
70.82&	2.86&	60.28&	25.66\\
OneRef-L ~\cite{xiao2024oneref}   & BEIT3-L / BEIT3-L &      & 77.20&	17.04&	59.85&	38.70&
77.59&	14.72&	69.26&	34.53&
73.71&	12.35&	67.73&	28.09\\
\cline{3-3}
CLIP-VG ~\cite{xiao2023clip}    & CLIP-B / CLIP-B   & \multirow{5}{*}{{TIR}}     & 14.52&	0.61&	13.64&	4.79&
12.19&	0.76&	10.45&	4.60&
9.88&	5.49&	8.51&	2.71\\
HiVG-B ~\cite{xiao2024hivg}     & CLIP-B / CLIP-B   &      &23.15&	0.81&	18.94&	10.51&
21.68&	0.97&	15.57&	10.81&
18.12&	0.51&	21.99&	7.02 \\
HiVG-L ~\cite{xiao2024hivg}     & CLIP-L / CLIP-L   &      & 45.77&	0.00&	37.88&	19.68&
54.77&	2.92&	43.85&	24.36&
53.18&	2.82&	45.74&	18.22\\
OneRef-B ~\cite{xiao2024oneref}   & BEIT3-B / BEIT3-B &      & 49.38&	0.20&	37.88&	20.74&
44.61&	0.11&	33.20&	19.69&
38.82&	0.12&	35.46&	13.91\\
OneRef-L ~\cite{xiao2024oneref}   & BEIT3-L / BEIT3-L &      &47.10&	3.65&	36.36&	22.87&
42.32&	4.17&	35.04&	19.12&
40.88&	2.43&	37.23&	15.09\\
\midrule
\multicolumn{13}{l}{\textbf{b. In-domain training with uni-modal visual input (RGB or TIR):}} \\ \midrule
TransVG  ~\cite{deng2021transvg}    & RN50+DETR / BERT-B & \multirow{11}{*}{{RGB}}   & 61.53&	19.76&	49.24&	31.12&
64.00&	15.10&	51.02&	31.91&
82.47&	31.06&	74.82&	36.65 \\
CLIP-VG ~\cite{xiao2023clip}    & CLIP-B / CLIP-B     &    &40.85&	15.81&	27.27&	16.22&
55.86&	5.39&	42.42&	22.51&
72.71&	13.16&	63.48&	27.89\\
D-MDETR ~\cite{shi2023dynamic}    & CLIP-B / CLIP-B     &    & 68.75& 26.77 &54.55 &37.10 
& 73.29&26.08 & 65.78&38.87
& 87.00& 40.98 &78.72 & 45.97 \\
D-MDETR ~\cite{shi2023dynamic}    & RN50+DETR / BERT-B  &    & 64.52&	20.95&	50.00&	34.04
& -& -&- &- 
& 82.24	&30.41	&74.47	&37.20  \\
MMCA ~\cite{yao2024visual}       & RN50+DETR / BERT-B  &    & 68.49&	21.50&	54.55&	33.64&
-&	-&	-&	-&
76.94&	26.24&	71.28&	33.31  \\
FSVG ~\cite{wang2025simple} &  RN50+DETR / BERT-B & &66.02&	20.87&	54.55&	32.45&	74.01&	23.70&	64.55&	37.79&	86.71&	39.90&	79.43&	45.27 \\
AttBalance ~\cite{kang2025visual} &  CLIP-B / CLIP-B & & 72.36&	21.80&	59.09&	36.84&	-&-&-&-&	83.65&	29.68&	76.95&	38.46 \\
HiVG-B ~\cite{xiao2024hivg}     & CLIP-B / CLIP-B     &    &85.39&	43.52&	75.76&	52.66&
90.51&	45.04&	84.84&	54.41&
90.41&	45.57&	86.17&	50.14\\
HiVG-L ~\cite{xiao2024hivg}     & CLIP-L / CLIP-L     &    &89.08&	50.00&	81.06&	59.31&
92.84&	41.29&	87.70&	56.65&
91.82&	44.98&	87.94&	51.60  \\
OneRef-B ~\cite{xiao2024oneref}   & BEIT3-B / BEIT3-B   &    & 85.21&	35.90&	75.76&	49.07&
92.79&	39.07&	87.30&	58.44&
90.24&	40.86&	86.17&	48.47 \\
OneRef-L ~\cite{xiao2024oneref}   & BEIT3-L / BEIT3-L   &    & 86.27&	39.76&	72.73&	52.26&
92.89&	34.52&	84.63&	54.92&
90.12&	45.18&	85.46&	49.72  \\
\cline{3-3}
TransVG  ~\cite{deng2021transvg}    & RN50+DETR / BERT-B & \multirow{11}{*}{{TIR}}   & 61.27&	21.18&	50.76&	32.98&
71.52&	22.51&	64.14&	37.98&
83.76&	31.37&	78.72&	38.46  \\
CLIP-VG ~\cite{xiao2023clip}    & CLIP-B / CLIP-B     &    &58.80&	12.26&	49.24&	26.99&
48.44&	4.92&	37.70&	22.83&
61.47&	10.31&	54.96&	23.23  \\
D-MDETR ~\cite{shi2023dynamic}    & CLIP-B / CLIP-B     &    &65.23&	18.38&	56.82&	35.24&
71.27&28.52 & 64.55&41.11 &
79.71 &37.73 &74.11 & 40.26 \\
D-MDETR ~\cite{shi2023dynamic}    & RN50+DETR / BERT-B  &    &62.59&	20.36&	56.06&	37.10
& -& -&- &- & 
80.71&	29.53&	73.76&	37.27  \\
MMCA ~\cite{yao2024visual}       & RN50+DETR / BERT-B  &    &61.44&	18.46&	52.27&	31.78&
-&	-&	-&	-&
76.41&	24.43&	70.57&	32.55  \\
FSVG ~\cite{wang2025simple} &  RN50+DETR / BERT-B & &55.90&	18.08&	44.70&	30.98& 74.95&	27.00&	64.96&	42.52&82.00&	36.16&	75.53&	40.26\\
AttBalance ~\cite{kang2025visual} &  CLIP-B / CLIP-B & &  66.55&	16.22&	55.30&	32.85&	65.30&	8.82&	50.61&	29.28&	79.82&	22.48&	71.28&	34.35\\
HiVG-B ~\cite{xiao2024hivg}     & CLIP-B / CLIP-B     &    & 77.11&	42.41&	68.94&	51.99&
87.50&	40.59&	80.53&	55.18&
86.59&	45.69&	79.08&	46.80 \\
HiVG-L ~\cite{xiao2024hivg}     & CLIP-L / CLIP-L     &    &84.95&	43.91&	84.85&	57.31&
87.60&	40.69&	82.79&	56.14&
87.41&	39.33&	80.85&	46.45  \\
OneRef-B ~\cite{xiao2024oneref}   & BEIT3-B / BEIT3-B   &    & 80.99&	35.80&	70.45&	52.39&
88.17&	41.02&	85.66&	58.76&
85.47&	38.67&	80.50&	40.47  \\
OneRef-L ~\cite{xiao2024oneref}   & BEIT3-L / BEIT3-L   &    &81.78&	37.53&	75.00&	53.99&
90.35&	45.29&	85.45&	62.02&
86.24&	42.67&	80.14&	47.36\\
\midrule
\multicolumn{13}{l}{\textbf{c. In-domain training with multi-modal visual input (RGB+TIR):}} \\\midrule

MV-TransVG    & RN50+DETR / BERT-B & \multirow{12}{*}{{RGB+TIR}} & 67.78&	21.10&	53.79&	35.51&
72.15&	22.89&	62.91&	38.87&
83.82&	33.84&	76.60&	40.40 \\
MV-CLIP-VG   & CLIP-B / CLIP-B     &  & 64.44&	16.06&	50.00&	29.79&
66.60&	10.01&	54.71&	30.43&
82.82&	23.96&	74.82&	36.93 \\
MV-D-MDETR    & CLIP-B / CLIP-B     &  &68.40& 23.38& 56.06& 36.70& 
67.89& 24.19 & 60.66& 38.04& 
84.24&39.30 &75.89&43.32 \\
MV-D-MDETR    & RN50+DETR / BERT-B  &  & 68.22&	24.52&	62.88&	37.90
&68.41&	21.81&	60.25&	35.49
&83.94&	34.24&	75.18&	40.75  \\
MV-MMCA      & RN50+DETR / BERT-B  &   & 71.92&	22.62&	59.85&	37.10&
74.07&	20.45&	64.14&	37.72&
84.76&	34.16&	77.66&	41.31  \\
MV-FSVG &  RN50+DETR / BERT-B & & 67.78&	24.69&	56.82&	36.57&	66.75&	25.00&	60.25&	37.98&	86.18&	40.17&	80.85&	44.65\\
MV-AttBalance &  CLIP-B / CLIP-B & & 76.32&	27.89&	68.18&	42.42&	79.36&	20.40&	71.52&	40.15&	83.94&	31.84&	76.60&	39.15 \\
MV-HiVG-B      & CLIP-B / CLIP-B     &  & 86.36&	50.20&	78.03&	56.91&
92.74&	\underline{49.40}&	89.55&	63.55&
\underline{91.24}&	\textbf{51.53}&	\textbf{89.72}&	\underline{54.03}\\
MV-HiVG-L    & CLIP-L / CLIP-L     &  & \underline{90.23}&	\underline{51.16}&	\textbf{83.33}&	\textbf{66.22}&
\underline{94.09}&	44.74&	\textbf{90.37}&	\underline{65.03}&
90.71&	42.36&	86.88&	51.32\\
MV-OneRef-B   & BEIT3-B / BEIT3-B   &  & 80.99&	35.43&	70.45&	52.39&
88.17&	37.93&	85.66&	58.76&
85.47&	41.58&	80.50&	40.47  \\
MV-OneRef-L   & BEIT3-L / BEIT3-L   &  & 83.19&	35.09&	71.97&	51.20&
92.17&	46.27&	87.50&	62.34&
88.94&	45.25&	81.21&	49.72 \\
RGBT-VGNet (Ours)  & CLIP-B / CLIP-B &  &
\textbf{90.59}	&\textbf{53.04}	& \underline{80.3}	& \underline{65.21}&
\textbf{94.25}&	\textbf{53.35}&	 \underline{90.18}&\textbf{	69.09}&
\textbf{91.83}&	 \underline{49.92}&	 \underline{89.36}&	\textbf{55.00}\\

\bottomrule
\end{tabular}
}
\label{tab:size_all_methods}
\end{table*}

\section{Fine-grained Benchmark Results on RGBT-GroundBench}
\label{supp:sec:more_eval}
\subsection{Evaluation Protocol}
To complement the results in the main paper, we report additional evaluations on all fine-grained subsets in RGBT-GroundBench. These subsets follow the multi-level annotations defined in the dataset and cover 13 scene types, 4 weather conditions, 4 illumination levels, 2 occlusion levels, and 2 object-size groups. Most experiments use the unified evaluation framework from the main paper at an image resolution of $224\times224$. We evaluate the same representative set of RGB-only, TIR-only, and RGB-TIR models listed in Table 4 of the main paper, including TransVG, CLIP-VG, D-MDETR, MMCA, HiVG, OneRef, their multimodal extensions, and RGBT-VGNet. All results are reported in terms of Acc@0.5, consistent with the manuscript.

\subsection{Results on Scene-Type Subsets (13 categories)}
\label{sec:scene_results_supp}

To study the effect of scene semantics on grounding performance, we evaluate all methods on the 13 scene categories in RGBT-GroundBench. The scene-level Acc@0.5 results on RefFLIR, RefM$^{3}$FD, and RefMFAD are reported in Tables~\ref{tab:scene_refflir_all_methods}, \ref{tab:scene_refm3fd_all_methods}, and \ref{tab:scene_refmfad_all_methods}. The scene labels cover a wide range of realistic environments and induce large performance variation across methods, which confirms that scene semantics is a meaningful source of difficulty for visual grounding.

Across all three sub-datasets, RGB+TIR models perform particularly well in the more challenging scenes. In rural and tunnel scenes, where illumination is uneven and texture is limited, RGB-only models tend to degrade more sharply, whereas multimodal models remain more stable. In parking-lot and market scenes, where object density and occlusion are high, thermal cues help separate the correct target from visually similar distractors. In highway and bridge scenes, RGB-only models are already relatively strong, but RGB+TIR still provides modest gains, especially for distant or partially visible objects.

The comparison across training settings shows a similar pattern. In the zero-shot setting, RGB models based on CLIP or BEiT3 provide a reasonable starting point, while zero-shot TIR models usually lag behind because of the larger gap between pretraining data and thermal imagery. After in-domain training, both RGB-only and TIR-only models improve substantially, but they remain sensitive to scene changes. Multimodal extensions of existing RGB models reduce this sensitivity across most scene types, and the gains are most visible in difficult scenes such as RR, TN, PL, and MK. RGBT-VGNet achieves the best or second-best performance in most cases, with its largest improvements appearing in scenes that combine difficult illumination with complex layouts.

These results also highlight two broader properties of the benchmark. First, the benefit of additional sensing modalities depends strongly on the scene, which makes RGBT-GroundBench useful for studying when thermal information matters most. Second, the relative ranking of methods is broadly consistent across RefFLIR, RefM$^{3}$FD, and RefMFAD for the same scene type, suggesting that the benchmark captures model behavior rather than dataset-specific artifacts.

\subsection{Results on Weather Subsets (4 categories)}
\label{sec:weather_results_supp}

We further evaluate the robustness of all methods under four weather conditions. The detailed Acc@0.5 results are summarized in Table~\ref{tab:weather_all_methods}, which follows the same organization as the scene-wise tables.

\noindent\textbf{Impact of adverse weather.}
Adverse weather significantly affects the visibility and appearance of objects in RGB images (as shown in Figures~\ref{suppfig:1}, \ref{suppfig:2}, \ref{suppfig:3}, and \ref{suppfig:4}), and this trend is clearly reflected in the table:
Foggy (FY) conditions are consistently the most challenging. In both zero-shot and single-source settings, RGB-only models show a clear drop in accuracy from SY to FY across all three sub-datasets. The reduced contrast, veiling glare, and scattering effects in FY/RY degrade high-level RGB semantics and make it difficult to precisely localize targets, especially at medium and long ranges. Cloudy (CY) scenes are less extreme: RGB-only models exhibit moderate drops compared with SY but remain relatively stable, as illumination is still sufficient and texture is largely preserved.

In contrast, the thermal modality is much less sensitive to these appearance changes, since it captures emitted or reflected infrared radiation rather than visible light. TIR-only and RGB+TIR models therefore maintain more stable performance under FY and RY, particularly in the single-source and multi-modal rows of Table~\ref{tab:weather_all_methods}.

\noindent\textbf{Advantages of multi-modal fusion.}
Across all three sub-datasets, multi-modal models (MV-* variants and RGBT-VGNet) show clear advantages over their single-modality counterparts:
The performance gap between RGB-only and RGB+TIR models is most pronounced in FY columns, where visible contrast is heavily reduced while thermal responses remain informative. Gains are also evident in RY columns, suggesting that RGB+TIR fusion compensates for rain streaks, reflections, and motion blur that frequently degrade RGB quality. Under CY and SY, RGB-only models are already strong, but RGB+TIR still brings incremental improvements, indicating that thermal information can provide complementary cues even in relatively benign weather.

\noindent\textbf{Performance of RGBT-VGNet.}
Our RGBT-VGNet attains the highest or second-highest accuracy in almost all weather conditions and datasets. In particular, its improvements over other RGB+TIR models are most visible in FY and RY, where the combination of AMA and LAVS yields stronger robustness to 
visibility degradation and background clutter. These results support the conclusion that carefully designed cross-modal interaction and language-aware fusion are especially beneficial when weather conditions significantly alter RGB appearance.

\noindent\textbf{Benchmark-level observations.}
From a benchmark perspective, the weather-wise results reveal several useful characteristics of RGBT-GroundBench:
For almost all models and training settings, accuracy follows SY $>$ CY $>$ (FY, RY), which indicates that the four subsets form a meaningful and stable difficulty spectrum for evaluating weather robustness. The benefit of TIR is amplified in adverse weather, particularly when combined with challenging scenes (e.g., rural roads or crowded markets), making the benchmark suitable for studying how environmental factors and sensing modalities jointly affect grounding performance. In addition, the relative ranking of methods under each weather condition is largely consistent across RefFLIR, RefM$^{3}$FD, and RefMFAD, suggesting that the weather annotations provide reliable signals rather than dataset-specific artifacts.

\subsection{Results on Illumination Subsets (4 categories)}
\label{sec:illum_results_supp}

The illumination subsets complement the TestB (low-light) evaluation in the main manuscript by providing four explicit illumination levels. The Acc@0.5 results on RefFLIR, RefM$^{3}$FD, and RefMFAD are reported in Table~\ref{tab:illum_all_methods}, using the same organization as the scene-wise and weather-wise tables.

\noindent\textbf{Illumination sensitivity of RGB-only models.}
In SL and NL conditions, RGB-only methods achieve relatively high performance on all three sub-datasets. In these regimes, appearance cues, color contrast, and contextual details are clearly visible, and high-level RGB semantics remain reliable. As illumination decreases from NL to WL and further to VL, all RGB-only baselines exhibit a sharp accuracy drop. This trend is consistent in both the zero-shot and single-source settings in Table~\ref{tab:illum_all_methods}. Low-light images contain stronger noise, reduced dynamic range, and missing color information, which substantially impairs the ability of RGB-only models to localize the referred objects, especially at medium and long distances.

\noindent\textbf{Characteristics of TIR-only models.}
Thermal-only models show a different pattern:
Their performance is generally lower than RGB-only methods in SL and NL, where visible textures and colors provide rich discriminative cues that are not fully captured by TIR. However, as illumination weakens, TIR-only models degrade much more slowly than RGB-only ones, and in several VL cases they become comparable or superior to RGB-only models, reflecting the illumination invariance of thermal sensing.

\noindent\textbf{Advantages of RGB+TIR fusion.}
Across all three sub-datasets, multi-modal RGB+TIR models (MV-* variants and RGBT-VGNet) provide the best balance between semantic richness and robustness to illumination changes:
In SL/NL, RGB+TIR models perform at least on par with the best RGB-only models, indicating that fusion does not harm performance in well-lit conditions. In WL/VL, RGB+TIR models clearly outperform both RGB-only and TIR-only baselines, confirming that the two modalities provide complementary information: RGB offers high-level semantics when usable, whereas TIR supplies stable structural cues when RGB deteriorates.

\noindent\textbf{Performance of RGBT-VGNet.}
The proposed RGBT-VGNet achieves the highest or second-highest accuracy in almost all illumination conditions and datasets. Its advantage is particularly pronounced in the \textbf{VL} columns of Table~\ref{tab:illum_all_methods}, where most RGB-only models collapse. The combination of AMA and LAVS allows RGBT-VGNet to rely more on TIR features when RGB becomes unreliable, while still exploiting RGB details under SL and NL. This leads to consistently strong performance across the entire illumination spectrum.

\noindent\textbf{Benchmark-level observations.}
From a benchmark perspective, the illumination-wise results highlight several useful properties of RGBT-GroundBench:
For nearly all methods, accuracy follows SL $\approx$ NL $>$ WL $>$ VL, forming a stable difficulty ladder for evaluating robustness to illumination degradation. Modality roles are also illumination-dependent: RGB dominates in SL/NL, TIR becomes increasingly important in WL/VL, and RGB+TIR fusion yields the best overall trade-off. The relative ranking of models at each illumination level is largely consistent across RefFLIR, RefM$^{3}$FD, and RefMFAD, which suggests that the illumination labels provide reliable, dataset-agnostic signals for future comparisons.

\subsection{Results on Object Size Subsets (2 categories)}
\label{sec:size_results_supp}

Although TestC in the main manuscript already focuses on small objects, here we report the full evaluation on Normal-Size (NS) and Small-Size (SS) subsets for RefFLIR, RefM$^{3}$FD, and RefMFAD in Table~\ref{tab:size_all_methods}.

\noindent\textbf{Challenges of small objects.}
Small-size (SS) targets remain the most challenging across all datasets and methods. Their limited pixel footprint and frequent partial occlusions make it difficult to extract discriminative visual features, and precise grounding requires accurate localization in very small regions. RGB-only models exhibit substantial performance degradation when moving from NS to SS, indicating that simply scaling model capacity or backbone strength is not sufficient to solve the small-object problem.

\noindent\textbf{Effect of RGB-TIR Fusion and RGBT-VGNet.}
RGB-TIR fusion generally yields consistent improvements on both NS and SS subsets. The gains are especially notable on SS, where thermal cues help highlight pedestrians and vehicles that are only weakly visible in RGB, particularly in cluttered or low-light environments. Multimodal MV-* baselines reduce the NS$\rightarrow$SS performance drop compared with single-modality models, demonstrating that complementary modalities are beneficial for tiny targets. RGBT-VGNet achieves the best or second-best performance on NS and SS across the three sub-datasets, with the largest margins typically observed on the SS subset. This confirms that AMA and LAVS effectively capture cross-modal cues and focus attention on small, linguistically referred instances.

\noindent\textbf{Benchmark-level observations.}
The size-wise results further characterize RGBT-GroundBench as a benchmark:
For nearly all methods, NS $>$ SS in accuracy, forming a clear and stable difficulty hierarchy with respect to object scale. The benefit of RGB+TIR fusion is more pronounced on SS than on NS, showing that multi-modal information is particularly valuable when visual evidence in a single modality is extremely limited. Consistent trends across RefFLIR, RefM$^{3}$FD, and RefMFAD indicate that the size annotations capture a general challenge rather than dataset-specific bias, providing a solid basis for evaluating future methods targeting small-object grounding.

\subsection{Results on Occlusion Subsets (2 categories)}
\label{sec:occlusion_results_supp}

To explicitly examine the influence of object occlusion, we report Acc@0.5 on the No-or-Partial Occlusion (PO) and Heavy Occlusion (HO) subsets for each sub-dataset in Table~\ref{tab:size_all_methods}.

\noindent\textbf{Effect of occlusion on single-modality models.}
Under the PO subset, most methods obtain relatively high and stable performance, since the referred objects are fully visible or only lightly occluded. In this regime, a strong semantic alignment between text and RGB appearance cues is usually sufficient for accurate grounding. Under the HO subset, all single-modality models suffer a noticeable drop, with RGB-only models being affected the most. When large portions of the target are occluded by other objects or scene structures, RGB appearance becomes highly ambiguous and models often confuse the target with nearby distractors. TIR-only models are slightly more robust than RGB-only models in HO for some cases, as thermal signals can still capture coarse shape or heat signatures. However, the lack of fine-grained semantics limits their overall accuracy.

\noindent\textbf{Benefits of thermal cues and multi-modal fusion.}
Thermal cues provide complementary information that remains visible even when RGB texture is severely occluded or visually similar to the background. This is reflected by the fact that RGB+TIR models exhibit significantly smaller performance drops from PO to HO compared with RGB-only baselines on all three sub-datasets. Multimodal extensions (MV-* variants) consistently improve over their single-modality counterparts in both PO and HO subsets, indicating that simple fusion strategies already help the model disambiguate occluded targets. RGBT-VGNet further reduces the PO$\rightarrow$HO gap and achieves the highest or second-highest accuracy in most occlusion settings. Its language-aware fusion (LAVS) allows the model to exploit relational phrases (e.g., ``behind the truck'', ``next to the bike'') to select the correct RGB-TIR regions, which is particularly important under heavy occlusion.

\noindent\textbf{Benchmark-level observations.}
From a benchmark perspective, the occlusion-wise results show that:
HO is consistently more difficult than PO for all methods and training settings, providing a clear and stable difficulty separation. The relative robustness of different modalities (RGB, TIR, RGB+TIR) is highly sensitive to occlusion, making these subsets suitable for studying how multi-modal fusion mitigates missing visual evidence. Method rankings under PO and HO are largely consistent across RefFLIR, RefM$^{3}$FD, and RefMFAD, suggesting that the occlusion annotations provide reliable, dataset-agnostic evaluation signals.

\subsection{Summary}
\label{sec:summary_multi_level_supp}

Across all extended evaluations on scene semantics, weather, illumination, object size, and occlusion, we obtain the following consistent observations:
The five types of annotations together form meaningful and stable difficulty spectra. Challenging conditions such as rural and tunnel scenes, foggy and rainy weather, weak or very weak illumination, small objects, and heavy occlusion all lead to significant performance drops for most methods, whereas structurally simple or well-lit settings are comparatively easier. This confirms that the multi-level factors in RGBT-GroundBench effectively stress different aspects of visual grounding robustness.

Multimodal RGB-TIR models consistently outperform RGB-only and TIR-only counterparts across almost all environmental and contextual subsets. RGB-only methods are strong in favorable conditions but degrade sharply when visibility is compromised; TIR-only methods are more stable in low-light or adverse weather but lack fine-grained semantics in easy regimes. RGB-TIR fusion achieves the best trade-off, demonstrating that thermal information is not only useful in extreme cases (e.g., very low light), but also provides complementary cues in normal settings.

The complementary nature of RGB and TIR is particularly crucial under low-light, foggy, heavy-occlusion, and small-object conditions, where appearance cues from a single modality become unreliable or ambiguous. In these challenging subsets, RGB-TIR fusion not only improves absolute performance but also reduces the performance gap between easy and hard cases, making RGBT-GroundBench a suitable testbed for studying when and how additional sensing modalities should be exploited.

The proposed RGBT-VGNet consistently ranks among the top performers across more than 25 fine-grained subsets (13 scene types, 4 weather conditions, 4 illumination levels, 2 occlusion levels, and 2 object sizes) on all three RGBT-GroundBench sub-datasets. Its gains are especially pronounced in the most challenging regimes, such as rural and tunnel scenes, foggy and rainy weather, weak or very weak illumination, small objects, and heavy occlusion. These results indicate that the combination of AMA and LAVS is effective not only for improving average accuracy, but also for stabilizing performance under severe domain shifts, where either RGB or TIR alone becomes unreliable. In this sense, RGBT-VGNet provides a strong and reproducible baseline for future work on RGB-TIR visual grounding.

The relative ranking of competing methods remains largely consistent across different subsets and across the three sub-datasets (RefFLIR, RefM$^{3}$FD, RefMFAD), despite their differences in acquisition conditions. This suggests that RGBT-GroundBench provides stable, dataset-agnostic measurements of model capability, rather than being dominated by idiosyncrasies of a single domain. Moreover, the multi-level annotations along scene, weather, illumination, size, and occlusion dimensions enable systematic analysis of failure modes, modality contributions, and design choices (e.g., single-modality vs.\ multi-modal fusion, zero-shot vs.\ fine-tuned training). Taken together, these properties make RGBT-GroundBench a comprehensive and reliable benchmark for fairly comparing future RGB-TIR visual grounding approaches under diverse and controllable real-world conditions.

\begin{figure*}[!t]
    \centering
    \includegraphics[width=\textwidth]{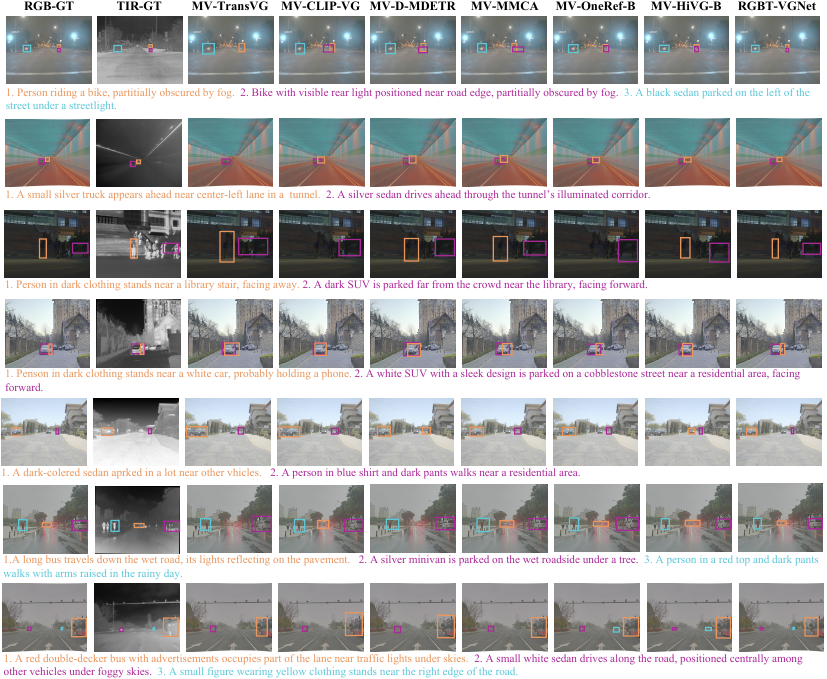}
    \caption{Additional qualitative comparison of multi-modal RGBT visual grounding results on RGBT-GroundBench across diverse real-world scenarios. Bounding box colors match the description below. Each column shows a different visual grounding method, stressing differences across methods under diverse scenes, illumination, weather, object size, and occlusion.}
    \label{suppfig:1}
\end{figure*}

\begin{figure*}
    \centering
    \includegraphics[width=\textwidth]{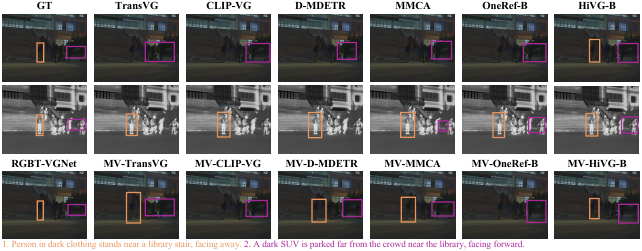}
    \caption{Qualitative comparison of RGB-only, TIR-only and RGB-TIR grounding results~(Figure~\ref{suppfig:1} row 3) on RGBT-GroundBench across diverse real-world scenarios. Bounding box colors match the description below. Each column shows a different visual grounding method, stressing differences across methods under diverse scenes, illumination, weather, object size, and occlusion.}
    \label{suppfig:2}
\end{figure*}

\begin{figure*}
    \centering
    \includegraphics[width=\textwidth]{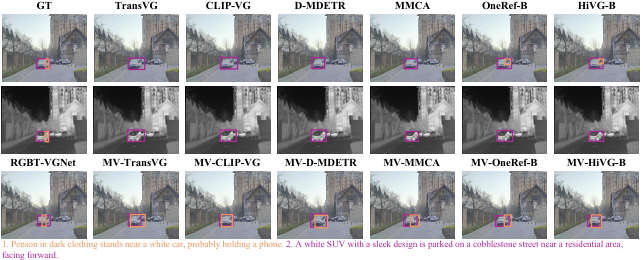}
    \caption{Qualitative comparison of RGB-only, TIR-only and RGB-TIR grounding results~(Figure~\ref{suppfig:1} row 4) on RGBT-GroundBench across diverse real-world scenarios. Bounding box colors match the description below. Each column shows a different visual grounding method, stressing differences across methods under diverse scenes, illumination, weather, object size, and occlusion.}
    \label{suppfig:3}
\end{figure*}

\begin{figure*}
    \centering
    \includegraphics[width=\textwidth]{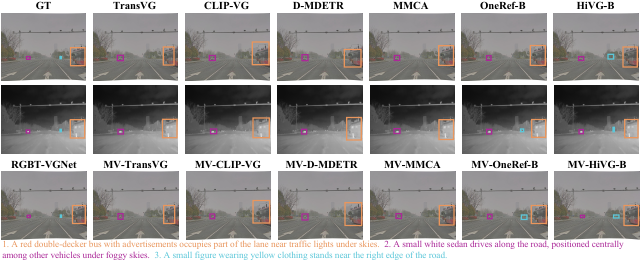}
    \caption{Qualitative comparison of RGB-only, TIR-only and RGB-TIR grounding results~(Figure~\ref{suppfig:1} row 7) on RGBT-GroundBench across diverse real-world scenarios. Bounding box colors match the description below. Each column shows a different visual grounding method, stressing differences across methods under diverse scenes, illumination, weather, object size, and occlusion.}
    \label{suppfig:4}
\end{figure*}

\section{More Qualitative Comparison Experiments}
\label{supp:sec:qual}
Figure~\ref{suppfig:1} presents additional qualitative results from multi-modal RGBT grounding methods under the unified evaluation framework. The examples cover a broad range of scenarios in RGBT-GroundBench and show how different methods behave under changes in illumination, weather, scene layout, object scale, and occlusion.

Nighttime and extremely low-light examples are shown in Figure~\ref{suppfig:1} (rows 1 and 3). In these cases, RGB images lose texture and contrast, while thermal images still preserve clear object contours. Most multi-modal methods remain more stable than their single-modality counterparts under these conditions. RGBT-VGNet is particularly reliable in these examples, suggesting that its fusion mechanism can draw useful information from the thermal modality without depending too heavily on degraded RGB inputs.

Tunnel scenes in Figure~\ref{suppfig:1} (row 2) illustrate a different type of difficulty. Around tunnel entrances and exits, RGB appearance changes abruptly because the illumination shifts between artificial light and daylight. This affects color, contrast, and shading, making visual features less stable over the scene. Thermal observations vary much less, and the predictions suggest that RGBT-VGNet is able to rely on these stable cues while still using RGB details when they remain informative.

Closed-road scenes in Figure~\ref{suppfig:1} (rows 3, 4, and 5) and Figure~\ref{suppfig:2} contain complex structures, strong perspective effects, and multiple plausible targets along the same road. In this setting, accurate grounding depends not only on appearance but also on the spatial relations described in the expression, such as proximity to landmarks or position along the lane. RGBT-VGNet often produces tighter and more precise predictions in these cases, which suggests that its language-aware design helps separate the referred target from nearby distractors.

The examples in Figure~\ref{suppfig:1} (row 4) and Figure~\ref{suppfig:3} focus on heavy occlusion. Here, dense objects and partial blockage make RGB evidence incomplete or ambiguous. The multi-modal results show that adding thermal input helps preserve the target's visibility when RGB texture is unreliable. RGBT-VGNet is more spatially consistent in these cases, which is in line with the quantitative results under occlusion.

Figure~\ref{suppfig:1} (rows 1, 6, and 7) further illustrates foggy, rainy, and otherwise challenging weather conditions. Fog and mist reduce contrast and blur object boundaries in RGB images, while thermal responses remain comparatively stable. The visualizations show that multi-modal input improves localization under these conditions, and RGBT-VGNet often returns more compact predictions, especially in foggy scenes.

At the same time, the qualitative results also reveal remaining failure modes. When targets are very small, heavily occluded, or surrounded by semantically similar distractors, even strong multi-modal models can still produce shifted or oversized boxes. These cases are consistent with the hardest quantitative subsets (testB/testC and HO/SS), and they indicate that robust fine-grained disambiguation remains an open problem.

\section{Complete Data Annotation Process}
\label{supp:sec:annotation}
\subsection{Scene and Weather Annotation Prompt}
To characterize the global context of each image, we designed a scene-level annotation prompt that captures the main environmental attributes. The prompt template is given below.

\lstset{
    columns=fullflexible,
    keepspaces=true
}
\begin{lstlisting}

Comprehensively analyze the global scene context of the provided image and return exactly four numerical codes representing the following environmental attributes:
(1)Scene:
    0-(Urban),
    1-(Suburban),
    2-(Rural),
    3-(Highway),
    4-(Residential),
    5-(Industrial),
    6-(Parking),
    7-(Intersection),
    8-(Tunnel),
    9-(Bridge),
    10-(Campus),
    11-(Market),
    12-(Waterfront)
(2)Weather:   
    0-(Foggy),
    1-(Rainy),
    2-(Sunny),
    3-(Cloudy)
Return only four numbers without additional text (space-separated).
\end{lstlisting}

This stage provides global context for later analysis of multi-sensor visual grounding.

\subsection{Lighting Annotation Prompt}
\noindent The lighting annotation prompt captures the overall illumination level of each scene. By assigning each image to one of four brightness levels, it provides metadata that is useful for analyzing the robustness of RGB, TIR, and RGB-TIR models under different lighting conditions.
\begin{lstlisting}[autogobble]
You are an expert at analyzing lighting conditions in images. I will provide an image, and your task is to classify the overall lighting intensity of the image into one of the following categories:
0. Very Weak Light: The image is mostly dark, but some features can be faintly seen.
1. Weak Light: The image has low visibility, typical for dawn or dusk.
2. Normal Light: The image has normal daylight brightness.
3. Strong Light: The image is brightly lit, typically from midday sunlight.
Please return only one number corresponding to the lighting condition: 0 (very_weak_light), 1 (weak_light), 2 (normal_light), or 3 (strong_light).
Now, please output the number.
\end{lstlisting}

\subsection{Object-Level Annotation Prompts and Rules}

Object-level annotations describe individual instances at a finer level of detail. We divide this process into three parts: referring-expression generation, occlusion annotation, and size classification.

The referring-expression prompt generates short but informative descriptions of each instance, covering appearance, spatial relations, and other distinguishing cues. The template is given below.

\begin{lstlisting}
I will provide an image and the bounding box coordinates (bbox) of a {category_name}. Your task is to describe the object within the bounding box in one concise sentence, focusing on its appearance and key features.
1. Object Details: Please describe the object in detail, including but not limited to its appearance, color, shape, texture, size, and posture.  
2. Contextual Relationship: If there is any relationship between the objects within the image or between the object and the background (e.g., relative positioning, interaction), please reflect this in your description.  
3. Distinguishing Similar Objects: If there are multiple similar objects in the image, differentiate them by comparing their details such as color intensity, position, state, etc.  
4. Concise and Clear Language: Provide a single, concise sentence that captures the key features without breaking down into different aspects. The description should be rich in information yet simple for later data processing.
Please generate only a few words of description for the {category_name} with the following bounding box coordinates: [{bbox}]
Now, please begin generating the description:
\end{lstlisting}

The occlusion annotation prompt estimates how much of the target remains visible. Its template is shown below.

\begin{lstlisting}[autogobble]
Analyze the occlusion level of the specified object and return exactly one integer value:
0 - No occlusion: The object is fully visible
1 - Partial occlusion: Some parts are obscured but key features remain visible  
2 - Heavy occlusion: More than 50% of the object is obscured
Target object bounding box: [x1, y1, x2, y2]
Return only 0, 1, or 2 without additional text.
\end{lstlisting}

Object size is determined automatically from the bounding-box area ratio, which divides instances into two classes. Table~\ref{tab:size-classification} summarizes the threshold definition used in this paper, and the same rule is applied consistently across all three sub-datasets.

\begin{table}[ht]
\centering
\caption{Object Size Classification Criteria}
\label{tab:size-classification}
\begin{tabular}{p{2cm}c} 
\toprule
\textbf{Category} & \textbf{Area Percentage} \\
\midrule
Small & $<$1\% \\
Normal & $\geq$1\% \\
\bottomrule
\end{tabular}
\end{table}

\begin{lstlisting}[autogobble, language=python]
def classify_object_size(bbox, img_width, img_height):
    x, y, w, h = bbox  # bbox format: [x, y, width, height]
    bbox_area = w * h
    image_area = img_width * img_height
    size_ratio = bbox_area / image_area
    
    return "small" if size_ratio < 0.01 else "normal"
\end{lstlisting}

Together, these annotations provide both scene-level context and object-level detail while keeping the overall pipeline simple to implement. The resulting labels support analysis of failure cases, modality sensitivity, and grounding behavior under diverse conditions.

\section{Potential Applications of the RGBT-GroundBench}
\label{supp:sec:applications}

Although RGBT-GroundBench is constructed for referring expression grounding, it can also support a compact evaluation suite for multimodal large models. Each sample provides paired RGB-TIR inputs, grounded language descriptions, and condition labels (weather, illumination, occlusion, and object size), which makes it suitable for evaluating perception under realistic degradation.

\subsection{Referring Expression Comprehension and Segmentation (REC/RES)}

REC evaluates whether a model can localize the referred target from language, while RES further evaluates whether it can recover the target mask. The existing boxes and expressions directly support REC, and they also provide practical supervision seeds for weakly supervised RES. This pair of tasks covers both coarse localization and fine-grained boundary perception, especially in difficult low-light and occluded subsets where RGB-only evidence is often insufficient.

\subsection{Counting}

The annotations can be converted into counting queries with limited additional processing, for example by asking for the number of objects that satisfy a description or relation constraint. This task stresses instance discrimination under clutter, scale variation, and partial occlusion. Compared with pure localization, counting better reveals whether a model separates nearby similar objects or collapses multiple instances into one.

\subsection{Visual Question Answering (VQA)}

The language annotations include attributes, positions, and inter-object relations, so they can be transformed into grounded VQA pairs with limited additional labeling. Typical questions include presence verification, relation reasoning, and condition-aware queries (e.g., under fog, low illumination, or occlusion). Because answers can be traced to explicit regions and paired modalities, this task helps evaluate whether a model reasons from visual evidence rather than from language priors.

\subsection{Captioning}

Captioning evaluates global scene understanding and language faithfulness. With aligned RGB-TIR observations and fine-grained condition labels, the benchmark can be used to test whether generated descriptions remain accurate under weak visibility and complex environments. A useful protocol is to score not only fluency, but also grounding consistency, namely whether mentioned objects, attributes, and relations are supported by the visual evidence.

Overall, these four tasks (REC/RES, Counting, VQA, and Captioning) provide a practical and complementary benchmark for measuring MLLM perception ability in RGB-TIR settings, from instance localization to semantic reasoning and holistic scene understanding. In this supplementary material, we present them as high-value extension directions supported by the current annotations; building standardized training and evaluation protocols for each task is a natural next step.

\end{document}